\documentclass[journal]{IEEEtran}
\usepackage{amsmath,amsfonts,amsthm}
\usepackage{algorithmic}
\usepackage{algorithm}
\usepackage{array}
\usepackage{textcomp}
\usepackage{stfloats}
\usepackage{url}
\usepackage{verbatim}
\usepackage{graphicx}
\usepackage{cite}
\usepackage{enumitem}
\usepackage[dvipsnames]{xcolor}
\usepackage[caption=false]{subfig}
\usepackage{booktabs}
\usepackage{multicol}
\usepackage{tabularx,ragged2e,siunitx}
\usepackage{makecell}
\usepackage[hidelinks]{hyperref}
\hypersetup{
    colorlinks,
    linkcolor={blue!50!black},
    citecolor={blue!50!black},
    urlcolor={blue!80!black}
}
\usepackage[capitalise,noabbrev]{cleveref}
\usepackage{multirow}

\theoremstyle{definition}
\newtheorem{definition}{Definition}[section]

\usepackage{tikz}
\usetikzlibrary{shapes,arrows}
\usetikzlibrary{positioning}
\tikzstyle{level0} = [rectangle,rounded corners, minimum width=18cm,minimum height=2cm,draw=black, thick, fill=orange!8]
\tikzstyle{level1} = [rectangle,rounded corners, minimum width=2cm,minimum height=1cm,draw=black, thick, fill=gray!8]

\tikzstyle{ca1} = [rectangle,minimum width=3.43cm,minimum height=0.8cm]
\tikzstyle{ca2} = [rectangle,minimum width=3.5cm,minimum height=0.8cm]
\tikzstyle{ca3} = [rectangle,minimum width=3.35cm,minimum height=0.8cm]
\tikzstyle{ca4} = [rectangle,minimum width=3.4cm,minimum height=0.8cm]
\tikzstyle{ca5} = [rectangle,minimum width=3.3cm,minimum height=0.8cm]

\tikzstyle{bg} = [rectangle,minimum width=3cm,minimum height=1cm,thick, draw=black]
\tikzstyle{cab1} = [rectangle,minimum width=3cm,minimum height=1cm,thick, draw=blue]
\tikzstyle{cab2} = [rectangle,minimum width=3.5cm,minimum height=1cm,thick, draw=gray]
\tikzstyle{cab3} = [rectangle,minimum width=3.3cm,minimum height=1cm,thick, draw=green]
\tikzstyle{cab4} = [rectangle,minimum width=3.3cm,minimum height=1cm,thick, draw=yellow]
\tikzstyle{cab5} = [rectangle,minimum width=3.3cm,minimum height=1cm,thick, draw=purple]
\tikzstyle{arrow} = [thick,->,>=stealth]

\begin{document}

\title{Explainable AI for Safe and Trustworthy Autonomous Driving: A Systematic Review}

\author{Anton Kuznietsov, Balint Gyevnar, Cheng Wang, Steven Peters, Stefano V. Albrecht
\thanks{The research by A. Kuznietsov is accomplished within the project “AUTOtech.agil” (FKZ 01IS22088S). We
acknowledge the financial support for the project by the Federal Ministry of Education
and Research of Germany (BMBF); B. Gyevnar was supported in part by the UKRI Centre for Doctoral Training in Natural Language Processing (grant EP/S022481/1). \textit{(A. Kuznietsov and B. Gyevnar contributed equally to this work.)} \textit{(Corresponding author: Cheng Wang.)}}%
\thanks{
        B. Gyevnar, and S.V. Albrecht are with the School of Informatics, University of Edinburgh, EH8 9AB Edinburgh, U.K. (e-mail: \{balint.gyevnar, s.albrecht\}@ed.ac.uk)

        A. Kuznietsov and S. Peters are with the Institute of Automotive Engineering,
        Technical University (TU) of Darmstadt, 64287 Darmstadt, Germany
        (e-mail: anton.kuznietsov@tu-darmstadt.de; steven.peters@tu-darmstadt.de)

        Cheng Wang is with the School of Engineering and Physical Sciences, Heriot-Watt University, EH14 4AS Edinburgh, U.K. (e-mail: Cheng.Wang@hw.ac.uk).

}}

\maketitle

\begin{abstract}
Artificial Intelligence (AI) shows promising applications for the perception and planning tasks in autonomous driving (AD) due to its superior performance compared to conventional methods. 
However, highly complex AI systems exacerbate the existing challenge of safety assurance of AD. 
One way to mitigate this challenge is to utilize explainable AI (XAI) techniques. 
To this end, we present the first comprehensive systematic literature review of explainable methods for safe and trustworthy AD.
We begin by analyzing the requirements for AI in the context of AD, focusing on three key aspects: data, model, and agency.
We find that XAI is fundamental to meeting these requirements. 
Based on this, we explain the sources of explanations in AI and describe a taxonomy of XAI. 
We then identify five key contributions of XAI for safe and trustworthy AI in AD, which are interpretable design, interpretable surrogate models, interpretable monitoring, auxiliary explanations, and interpretable validation. 
Finally, we propose a conceptual modular framework called SafeX to integrate the reviewed methods, enabling explanation delivery to users while simultaneously ensuring the safety of AI models. 
\end{abstract}


\section{Introduction}\label{sec:intro}
\IEEEPARstart{A}{rtificial} intelligence (AI) has gained a lot of attention in various technical fields in the last decades.
Particularly, deep learning (DL) based on deep neural networks (DNNs) provides human-comparable or potentially even better performance for some tasks due to its data-driven high-dimensional learning ability~\cite{Mathew2020DL,Jammal2023HumanMachine}, so it has naturally emerged as a vital component in the field of autonomous driving (AD). 

Nevertheless, deep learning suffers from a lack of transparency. 
It exhibits black-box behaviour, obscuring insights into its internal workings. 
This opacity makes it harder to identify issues and to determine which applications of AI are admissible in the real world.
However, in safety-relevant domains such as AD, it is crucial to develop safe and trustworthy AI. 
Although there are several mitigation processes to handle safety concerns in AI, such as well-justified data acquisition~\cite{willers2020safety}, the adequacy of these measures in ensuring sufficient safety remains an open question, highlighting the need for further approaches.
 
Moreover, no standards currently explicitly address the use of data-driven AI in AD. 
The existing safety standard ISO 26262 - \textit{Road Vehicles - Functional safety}~\cite{iso_iso_2018} was not explicitly developed for data-driven AI systems and their unique characteristics~\cite{salay2018using}. 
The standard ISO 21448 - \textit{Safety of the Intended Functionality} (SOTIF)~\cite{iso_iso_2022} aims at ensuring the absence of unreasonable risk due to hazards from functional insufficiencies of the system and requires quantitative acceptance criteria or validation targets for each hazard. 
The concept can be applied to AI-based functions, but these acceptance criteria are not explicitly defined~\cite{burton2022safety}.
Moreover, specific guidance for designing AI-based functionality is missing.

As a result, these standards face challenges in addressing safety requirements for data-driven deep learning systems~\cite{nuissl}. 
Although there is ongoing work on the ISO/AWI PAS 8800 - \textit{Road Vehicles - Safety and Artificial Intelligence} \cite{iso8800}, its scope and guidance remain unclear due to it still being in a development phase.
In general, there is also a relatively high level of mistrust in society regarding AD.
The American Automobile Association's survey on autonomous vehicles (AV) indicates that 68\% of drivers in the United States are wary of AVs~\cite{AAA}, and AI has been identified as one of the key factors contributing to the non-acceptance of AVs in society \cite{Reig2018studyAIAD}. 

A promising approach to address these problems is explainable AI (XAI). 
XAI aims to provide human-understandable insights into the behaviour of the AI and the development of XAI methods could be beneficial for different kinds of stakeholders~\cite{langer2021we}. 
First, it may become an essential tool for AI developers to identify and debug malfunctions~\cite{Dwivedi2023Debug}. 
Second, XAI could help users calibrate their trust in automated systems in line with the actual capabilities of AVs~\cite{Weitz2019TrustXAI}, thereby preventing misuse. 
Lastly, assurance companies and regulatory bodies may also benefit, as the increased transparency due to XAI could enable traceability that allows for a more accurate assessment of due diligence and liability in case of accidents~\cite{ColumbiaLaw2019xAI}.
Muhammad et al.~\cite{Muhammad2021SafeAD} go as far as to say that in the future XAI could be necessary in terms of regulatory compliance including fairness, accountability and transparency in DL for AD. 
Given the increasing size of literature on XAI specifically for AD, it is necessary to systematically review which XAI techniques exist and how they are applied to enhance the safety and trustworthiness of AD.

\subsection{Previous Reviews on XAI for AD} \label{sec:intro:previous_survey}
Indeed some reviews of XAI for AD already exist and we give a brief overview of each in this subsection.
These works provide a good overview of the challenges and stakeholders of the field but have some crucial shortcomings:
\begin{enumerate}
    \item Lack of a systematic literature review methodology, leading to potential bias and incomplete coverage;
    \item No focus on the specific benefits and drawbacks of XAI on the safety and trustworthiness of AD;
    \item No review of frameworks for integrating XAI with AD.
\end{enumerate}

The work of Omeiza et al.~\cite{omeiza2021explanations} was the first notable survey in the field. 
They provide a holistic look at XAI for AD, covering the different needs for explanations, regulations, standards, and stakeholders, and an overview of some explainability methods applied in AD. 
They review the challenges involved in designing useful XAI systems for AD and the associated literature, however, this review is neither reproducible nor complete, especially for the perception and planning tasks.

In addition, Atakishiyev et al.~\cite{atakishiyev2021explainable} covered very similar topics to Omeiza et al. with a slightly broader coverage of recent XAI technologies for AV perception and planning.
They propose an end-to-end (E2E) framework for integrating XAI with existing AD technologies, however, they did not elaborate how different XAI techniques can be integrated into the framework. 
Their literature review was also not described in sufficient detail to be repeatable.

Finally, the literature review of Zablocki et al.~\cite{zablocki2022explainability} identified potential stakeholders and why they might need explanations, the type of explanations useful for them, and when explanations need to be delivered.
Based on that, they examine the different methods in the literature. 
However, they do not focus on the impact of XAI in meeting the requirements for safe and trustworthy AI. 
Furthermore, the survey has some shortcomings regarding completeness, since they only focus on vision-based methods for E2E systems.
Accordingly, they do not consider XAI methods for planning and perception that can be applied to modular AD pipelines.

\subsection{Main Contributions}
In light of the existing works and the increasing importance of XAI for AD, we make the following contributions:
\begin{enumerate}
    \item We discuss detailed requirements for AI in AD and highlight the importance of XAI in fulfilling them; %
    \item Using a structured, systematic, and repeatable review methodology, we survey XAI methods applied for AD with a focus on environmental perception, planning and prediction, and control;
    \item Based on our review, we identify five paradigms of XAI techniques applied for safe and trustworthy AD which include interpretable design, interpretable surrogate models, interpretable monitoring, auxiliary explanations, and interpretable validation. Moreover, we discuss each paradigm using concrete examples;
    \item We analyse the limitations of existing modular XAI frameworks for AD and then propose a conceptual framework called \textit{SafeX} that is designed to be readily used with the summarized XAI techniques.
\end{enumerate}

\subsection{Scope and Structure} 
Our study gives a comprehensive view of the current state-of-the-art XAI approaches for AD encompassing both modular and E2E pipelines, focusing on perception, planning and prediction, and control.
We also present a conceptual modular framework to incorporate XAI into the design of AVs.
Our survey does not identify stakeholders nor aims to give background on mathematical foundations such as DNNs or reinforcement learning as existing surveys in~\cref{sec:intro:previous_survey} provide good coverage of these topics.

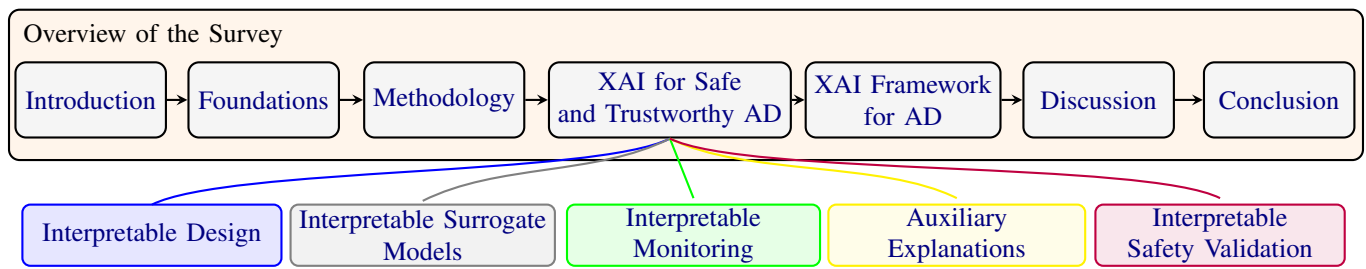
\begin{figure*}[!t]
    \centering
    \begin{tikzpicture}[node distance=2cm,every node/.style={font=\normalsize}]
    \node (outer) [level0]{};
    \node[anchor=north west, inner sep=2mm] at (outer.north west) {Overview of the Survey};
    
    \node (introduction) at ([xshift=1cm]outer.west) [level1, align=left, xshift=0.1cm, yshift=-0.2cm]{\hyperref[sec:intro]{Introduction}};
    \node (background) [level1, right of=introduction, xshift=0.3cm] {\hyperref[sec:background]{Foundations}};
    \node (methodology) [level1, right of=background, xshift=0.4cm] {\hyperref[sec:methodology]{Methodology}};
    \node (XAI) [level1, right of=methodology, align=center, xshift=1cm] {\hyperref[sec:cat]{XAI for Safe} \\ \hyperref[sec:cat]{and Trustworthy AD}};
    \node (framework) [level1, right of=XAI, align=center, xshift=1.1cm] {\hyperref[sec:framework]{XAI Framework} \\ \hyperref[sec:framework]{for AD}};
    \node (discussion) [level1, right of=framework, xshift=0.6cm] {\hyperref[sec:discussion]{Discussion}};
    \node (conclusion) [level1, right of=discussion, xshift=0.4cm] {\hyperref[sec:conclusion]{Conclusion}};

    \node (category1) [ca1, below of=outer, xshift=-7.1cm ] {};
    \draw [draw=blue, thick, fill=blue!10] (category1.south west) [rounded corners=3pt] -- (category1.north west) [rounded corners=3pt] -- (category1.north east) [rounded corners=3pt] -- (category1.south east) [rounded corners=3pt] -- cycle;
    \node (category1) [ca1, below of=outer, xshift=-7.1cm ] {\hyperref[ssec:cat:interp]{Interpretable Design}};

    \node (category2) [ca2, right of=category1, xshift=1.6cm ] {};
    \draw [draw=gray, thick, fill=gray!10] (category2.south west) [rounded corners=3pt] -- (category2.north west) [rounded corners=3pt] -- (category2.north east) [rounded corners=3pt] -- (category2.south east) [rounded corners=3pt] -- cycle;
    \node (category2) [ca2, right of=category1, align=center, xshift=1.6cm ] {\hyperref[ssec:cat:surr]{Interpretable Surrogate} \\ \hyperref[ssec:cat:surr]{Models}};

    \node (category3) [ca3, right of=category2, xshift=1.6cm] {};
    \draw [draw=green, thick, fill=green!10] (category3.south west) [rounded corners=3pt] -- (category3.north west) [rounded corners=3pt] -- (category3.north east) [rounded corners=3pt] -- (category3.south east) [rounded corners=3pt] -- cycle;
    \node (category3) [ca3, right of=category2, align=center, xshift=1.6cm] {\hyperref[ssec:cat:monitor]{Interpretable} \\ \hyperref[ssec:cat:monitor]{Monitoring}};

    \node (category4) [ca4, right of=category3, xshift=1.5cm] {};
    \draw [draw=yellow, thick, fill=yellow!10] (category4.south west) [rounded corners=3pt] -- (category4.north west) [rounded corners=3pt] -- (category4.north east) [rounded corners=3pt] -- (category4.south east) [rounded corners=3pt] -- cycle;
    \node (category4) [ca4, right of=category3,  align=center, xshift=1.5cm] {\hyperref[ssec:cat:aux]{Auxiliary} \\ \hyperref[ssec:cat:aux]{Explanations}};

    \node (category5) [ca5, right of=category4, xshift=1.5cm] {};
    \draw [draw=purple, thick, fill=purple!10] (category5.south west) [rounded corners=3pt] -- (category5.north west) [rounded corners=3pt] -- (category5.north east) [rounded corners=3pt] -- (category5.south east) [rounded corners=3pt] -- cycle;
    \node (category5) [ca5, right of=category4, align=center, xshift=1.5cm] {\hyperref[ssec:cat:valid]{Interpretable} \\ \hyperref[ssec:cat:valid]{Safety Validation}};
    
    \draw[blue, thick] (XAI.south) .. controls ([xshift=-1cm, yshift=-0.5cm]XAI.south) 
              and ([xshift=1cm, yshift=0.5cm]category1.north) .. (category1.north);
    \draw[gray, thick] (XAI.south) .. controls ([xshift=-1cm, yshift=-0.5cm]XAI.south) 
              and ([xshift=1cm, yshift=0.5cm]category2.north) .. (category2.north);
    \draw[green, thick] (XAI.south) .. controls ([xshift=0.2cm, yshift=-0.5cm]XAI.south) 
              and ([xshift=-0.2cm, yshift=0.5cm]category3.north) .. (category3.north);      
    \draw[yellow, thick] (XAI.south) .. controls ([xshift=1cm, yshift=-0.5cm]XAI.south) 
              and ([xshift=-1cm, yshift=0.5cm]category4.north) .. (category4.north);  
    \draw[purple, thick] (XAI.south) .. controls ([xshift=1cm, yshift=-0.5cm]XAI.south) 
              and ([xshift=-1cm, yshift=0.5cm]category5.north) .. (category5.north);  
    \draw [arrow] (introduction) -- (background);
    \draw [arrow] (background) -- (methodology);
    \draw [arrow] (methodology) -- (XAI);
    \draw [arrow] (XAI) -- (framework);
    \draw [arrow] (framework) -- (discussion);
    \draw [arrow] (discussion) -- (conclusion);
    
    \end{tikzpicture}
    
    \caption{The structure of our survey. We present foundations into why we need trustworthy AI, an XAI taxonomy, and AD terminology. We then describe our survey methodology in detail so that our review is reproducible. In the analysis, we categorize existing XAI for AD approaches into five branches based on their different applications for AD: interpretable design, interpretable monitoring, interpretable surrogate models, auxiliary explanations and interpretable validation. A new conceptual framework is presented called SafeX based on our analysis. Finally, we discuss challenges and future directions.}
    \label{fig:structure}
\end{figure*}

The structure of our survey is illustrated in~\cref{fig:structure}.
In~\cref{sec:background}, we provide foundations, where we define trustworthy AI and identify requirements corresponding to the application of AI in AD. 
Moreover, we describe the various sources of explanations for AI systems and introduce a taxonomy of XAI concepts as well as terminology for AD components. 
\Cref{sec:methodology} describes our research questions and the methodology for the survey, assuring reproducibility. 
\Cref{sec:cat} surveys the literature divided into interpretable design, interpretable monitoring, interpretable surrogate models, auxiliary explanations, and interpretable validation.
In~\cref{sec:framework}, we review existing XAI frameworks in AD and propose our framework SafeX. 
In~\cref{sec:discussion} and~\cref{sec:conclusion}, we discuss our findings and identify future directions in light of the results of our survey.

\section{Foundations}\label{sec:background}
We begin in this section with an exploration of the need for trustworthy AI. 
We then examine the requirements for applying trustworthy AI in AD. 
Our analysis highlights the critical role of XAI in fulfilling these requirements and identifies the sources of explanations in AI systems.
We conclude by reviewing a taxonomy of XAI along with a detailed terminology of AD components.

\subsection{Trustworthy AI}\label{ssec:background:trustworthyAI}
Historically, AI was based on \textit{symbolic} representations, where information was encoded using well-defined mathematical symbols, such as propositional logic or program induction.
The first instances of successful AI applications were expert systems~\cite{buchananFundamentalsExpertSystems1988} which relied on such symbolic representations, lending themselves to varying degrees of inherent interpretability, that usually manifested in the form of causal chains of reasoning.

In contrast, current neural AI methods rely on \textit{sub-symbolic} representations. 
Under this paradigm, input data is mathematically transformed into output via the learning of millions of parameters from large swathes of training data.
This approach allows the modelling of highly complex multi-dimensional relationships which results in high performance.
Still, the outputs of sub-symbolic systems are not interpretable due to their sheer size and high levels of abstraction. 
Therefore, they are often likened to black boxes that lack transparency.

While this efficiency versus transparency trade-off is sometimes acceptable (arguably, even non-existent~\cite{rudinStopExplainingBlack2019}), highly complex safety-relevant systems such as AD cannot fully rely on black box systems, as they are not currently certifiable for safety.
This is in addition to the countless ethical, social, and legal reasons why neural methods may also be suspect~\cite{kaurTrustworthyArtificialIntelligence2022}.

Symptomatic of these issues is the lack of trust by users of AI systems.
To alleviate the many problems that stem from a lack of transparency, methods that automatically explain predictions and decisions to users have become popular~\cite{burkartSurveyExplainabilitySupervised2021}, forming the field of XAI.
However, achieving trustworthy AI is a much more complex issue than could be solved by merely imbuing AI systems with explainability.
Instead, trustworthy AI must consider a complex set of socio-technical requirements, among others, human agency, technical robustness and safety, privacy and data governance, diversity, non-discrimination, fairness, and societal and environmental well-being~\cite{gyevnarBridgingTransparencyGap2023}.
Our focus on XAI is not to suggest that we can achieve trustworthy AI just via explainable methods but as a necessary element among the many approaches that support \textit{human-centric AI} which strives to ensure that human values are central to how AI systems are developed, deployed, used, and monitored by ensuring respect for basic human rights.

In the following subsection, we explore in detail these requirements for trustworthy AI specifically for AD. 
Subsequently, we discuss from which sources and to what extent current AI methods are amenable to explanation and overview a taxonomy of XAI to organise our discussion.

\subsection{Requirements For Safe and Trustworthy AI in AD}\label{ssec:background:requirements}

Owing to the superior performance in high-dimensional tasks like image processing and object detection~\cite{feng2021review}, black box sub-symbolic methods are now the predominant approach to solving challenges in AD. 
Unlike many other robotic domains, incorrect behaviour by AVs can cause serious injury or death to humans, meaning safety is a top priority for all stakeholders.
Designing safe and trustworthy AI is, thus, becoming urgent for AVs, necessitating the definition of safety and trustworthiness requirements.

Unfortunately, no requirements are published specifically for AI in AD.
Instead, we need to take more general requirements for safe and trustworthy AI as a starting point. 
We discuss whether these map to AI systems in AD and whether new requirements for AI in AD should be defined in~\cref{tab:requirements}. 

Fortunately, several guidelines have been developed for safe and trustworthy AI which are well-suited to address the transparent and safety-critical operation of AVs.
One of the well-known AI regulations is the ethics guidelines released by the European Commission~\cite{EUTrustAI}, in which seven key requirements for trustworthy AI were defined. 
These are (i) human agency and oversight; (ii) technical robustness and safety; (iii) privacy and data governance; (iv) transparency; (v) diversity, non-discrimination and fairness; (vi) societal and environmental well-being; (vii) accountability.

Another AI risk management framework was developed by the American National Institute of Standards and Technology (NIST)~\cite{ai2023artificial}.
This also defined seven key characteristics so that trustworthy AI should be (i) valid and reliable; (ii) safe; (iii) secure and resilient; (iv) accountable and transparent; (v) explainable and interpretable; (vi) privacy-enhanced; (vii) fair with harmful bias management. 
According to this framework, validity and reliability are the bases for other characteristics, while accountability and transparency are overarching concepts related to all characteristics.

The requirements defined in these two proposals are derived from three main sources: data-, model-, and agency-related requirements.
We synthesise them in~\cref{tab:requirements}.
First, diverse data and data governance are essential to avoid unbiased decisions and protect privacy. 
Second, an AI model itself ought to be, among others, robust, safe, and accountable. 
Third, the deployed AI models must be overseen by humans for which human agency is required.
Similar requirements are proposed by other individual researchers. 
For instance, Alzubaidi et al.~\cite{alzubaidi2023towards} defined similar requirements for trustworthy AI. 
They considered accuracy and reproducibility as separate requirements while the EU assigned those requirements to robustness.
To avoid unnecessarily conflating conflicting definitions, we take the requirements derived from the two national-level proposals as our starting point, noting that other conceptions of trustworthy AI may be fit to these frameworks.

\begin{table*}[!ht] 
\caption{Summary of the defined requirements in the ethics guidelines (EU) and the AI risk management framework (USA) and discussion of their applicability to AD. The requirements are classified into three sources: data, model and agency.}
\centering
\begin{tabularx}{0.9\linewidth}{@{} p{2cm} p{5cm} p{5.3cm} p{2.5cm} @{}}
\toprule
\textbf{Sources}         & \textbf{Ethics Guidelines (EU)}           & \textbf{AI Risk Management Framework (USA)}    & \textbf{Transferable to AD?} \\ 
\midrule
\multirow{2}{*}{Data}    & Privacy and data governance               & Privacy-enhanced                  & Y   \\
                         & Diversity, non-discrimination, fairness   & Fair with harmful bias management & Y   \\ 
                         \midrule
\multirow{5}{*}{Model}   & Technical robustness and safety           & Safe                              & Y   \\
                         & Transparency                              & Accountable and transparent       & Y   \\
                         & Accountability                            & Valid and reliable                & Y   \\
                         & Societal and environmental well-being     & Secure and resilient              & Y   \\
                         & ---                                         & Explainable and interpretable     & Y   \\ \midrule
Agency                   & Human agency and oversight                & ---                                 & Y/N \\ 
\bottomrule
\end{tabularx}
\label{tab:requirements}
\end{table*}

\textbf{Requirements From Data}: 
Proper data governance is necessary for AD since privacy- and quality-sensitive data from drivers and external environments need to be processed. 
For instance, ML-based perception typically uses vision systems to perceive and understand the surroundings, where highly personal data such as pedestrian faces and license plates also appear.
However, it is not necessarily the case that technical data must be classified as non-personal~\cite{andravsko2021sustainable}. 
Under EU jurisdiction, for instance, the General Data Protection Regulation (GDPR)~\cite{GDPR2016a} may provide a legal basis for processing personal data when using AI-based functionalities, though it is unclear to what extent the unilateral and fully automated processing of personal data in AVs is covered by the GDPR.
Moreover, to avoid, among others, unfair bias, we should rely on diverse data to train AI models.
Particularly, the non-discrimination of pedestrians is an important requirement for ML-based systems in AD. 
Despite this, Li et al.~\cite{li2023dark} showed a bias for missing pedestrians who are children or have darker skin tones.
In general, the elicitation of safety-related requirements should be seen as a process that includes multiple stakeholder perspectives to increase diversity~\cite{burton2022safety}.

\textbf{Requirements From AI Models}: 
AD is a safety-critical application where a lack of robustness could lead to traffic accidents. 
The environment in which an AV operates is complex, uncertain, and changes over time and space. 
DL-based models need to be robust not only to variations in the physical driving condition (e.g., differing weather conditions, and changes to the car behaviour due to component wear) but also to variations in the behaviours of other drivers, including the possibility that adversarial road users may try to exploit AV systems~\cite{eykholt2018robust}.
In addition, adversarial perturbations can fool deep learning-based neural networks~\cite{yuan2019Adversarial}, leading to implausible results. 
Therefore, AI needs to be robust against, among others, noise, distribution shift, and adversarial attacks~\cite{Muhammad2021SafeAD} and must demonstrate safe decisions even in uncertain environments.
Moreover, AVs need to be sufficiently transparent for the involved stakeholders such that the decisions of the AV can be understood. 
For instance, developers need transparency to debug models and thus improve system robustness, while regulators need transparency to audit and certify systems. 
Furthermore, deployed AI models should be user-centric and designed in a way that all people can benefit from their services regardless of their situation.
Finally, establishing accountability for AD systems is important for determining liability in case of accidents~\cite{omeiza2021explanations}. 

\textbf{Requirements For Agency}: 
For level 3 AVs~\cite{sae_j3016_taxonomy_2021}, human drivers are allowed to do non-driving-related activities while the AV undertakes dynamic driving tasks (DDTs) as long as it remains within the predefined operational design domain (ODD). 
Nevertheless, a driver should be prepared to take over control at any moment if the system fails or when the ODD is exceeded. 
In contrast, for more advanced level 4 and 5 systems, human drivers no longer need to stay in the loop which diminishes their oversight, especially when the AI systems are inscrutable.
Without additional measures, human agency will suffer due to the use of black box systems. 
In particular, the users of AVs may have very little insight into the decision-making processes of AVs and could never hope to contest the decisions that may directly impact their bodily integrity.
However, obtaining recourse in these situations may not just be a matter of enabling intervention on the AD systems, but rather the provision of explanations that calibrate users' trust according to the system's capabilities.
Therefore, depending on the automation level of the system, the requirements for agency may or may not be addressed by existing systems.

In addition to the above three categories of requirements, safety assurance is imperative for AD~\cite{NascimentoSafety2020,ribeiro2022requirements}. 
Safety is fundamentally important, underpinning and complementing the above three high-level requirements of trustworthy AI for AD.
While requirements for trustworthy AI often include safety, safety assurance imposes strict constraints on the behaviour of AI systems as opposed to the more high-level criteria of other aspects of trustworthy AI. Therefore, safety could be viewed as a distinct set of requirements.

However, providing a comprehensive account of safety requirements would mandate its own publication, so instead we focus on one way of ensuring certain safety requirements that also align with many of the recommendations for trustworthy AI, namely explainable AI.
First, XAI contributes to transparency by delivering (intelligible) explanations of AI models' decisions.
To show compliance with data protection regulations, one may call on XAI to provide evidence that AD systems do not process personal data and that they can function without personally identifying features in the data.
Second, accountability through inquiry and traceability may be achieved, which is essential to show non-discrimination, determine failure cases, and establish a holistic case of their workings for legal proceedings or regulatory conformity.
Third, XAI is beneficial for the inspection, debugging, and auditing of AI models, which can contribute to improved robustness and better-calibrated trust in AVs~\cite{Sheh2021XAIrequirements}. 
Therefore, we must conclude, that XAI is an essential tool in meeting the requirements of safe and trustworthy AI for AD.

\subsection{Sources of Explanation in AI}\label{ssec:background:sources}
There are many ways in which an AI system may be amenable to revealing its decision-making process through explanation.
Crucially, how information is represented and then used in the AI system directly influences the ways we can generate explanations for them.
Understanding these different sources of explanation is essential to effectively discussing various methods of XAI because it helps not only to build a consistent vocabulary for XAI but also to understand the applicability of various XAI systems to AD and how they address the various requirements for safe and trustworthy AD.

We discuss here five sources of explanation from the XAI literature.
These are: interpretability, explainability, justifiability, traceability, and transparency.
These terms are related and not necessarily mutually exclusive properties of AI systems, however, they are often (incorrectly) used interchangeably.
Our discussion here is informed by~\cite{gyevnarBridgingTransparencyGap2023, angelov2021explainable, miller2019explanation}.

\textbf{Interpretability}: we call an AI system interpretable if it is sufficiently low in complexity such that a reasonably experienced user can understand the output of the system and the causal process that produced that output from the input~\cite{molnarInterpretableMachineLearning2023}.
Therefore, interpretability is an inherent quality of a system.
Interpretable systems are often argued to be better suited for safety-relevant applications due to the observable chain of causality that led to a decision~\cite{rudinStopExplainingBlack2019}.

\textbf{Explainability}: we call an AI system explainable if the output of the system is accompanied by an additional output that takes the syntactic form of an explanation.
The explanation should intelligibly communicate the reasoning process behind how the output was derived~\cite{schwalbe2023comprehensive}.
Explainability is not necessarily an inherent quality of the AI system, and may not accurately reflect the causal chain that produced the output.

\textbf{Justifiability}: an AI system's decision is justifiable if one can explain why an output was good without necessarily explaining how the output was computed~\cite{miller2019explanation}.
This property depends on a definition of goodness that will inherently depend on the application domain and the ethical framework the designers of the system see fit for use.

\textbf{Traceability}: an AI system is traceable if an external auditor can follow the causal chain of the full decision-making process from input to output.
Any system that relies on a black box is not traceable since causality is obscured by design.
A system might also only rely on white box systems but still be untraceable due to the sheer size of the models.

\textbf{Transparency} is a broad term that is often used (incorrectly) to mean any of the above definitions.
As discussed in~\cref{ssec:background:trustworthyAI}, transparency is not solely the property of the AI system achievable via XAI but the result of a range of measures that enable the understanding and informed use of the system through a combination of, among others, documentation, XAI, standardisation, and risk assessments~\cite{gyevnarBridgingTransparencyGap2023}.

\subsection{Taxonomy of XAI}\label{ssec:background:XAI-taxonomy}

\begin{figure}[t]
    \centering
    \includegraphics[width=\linewidth]{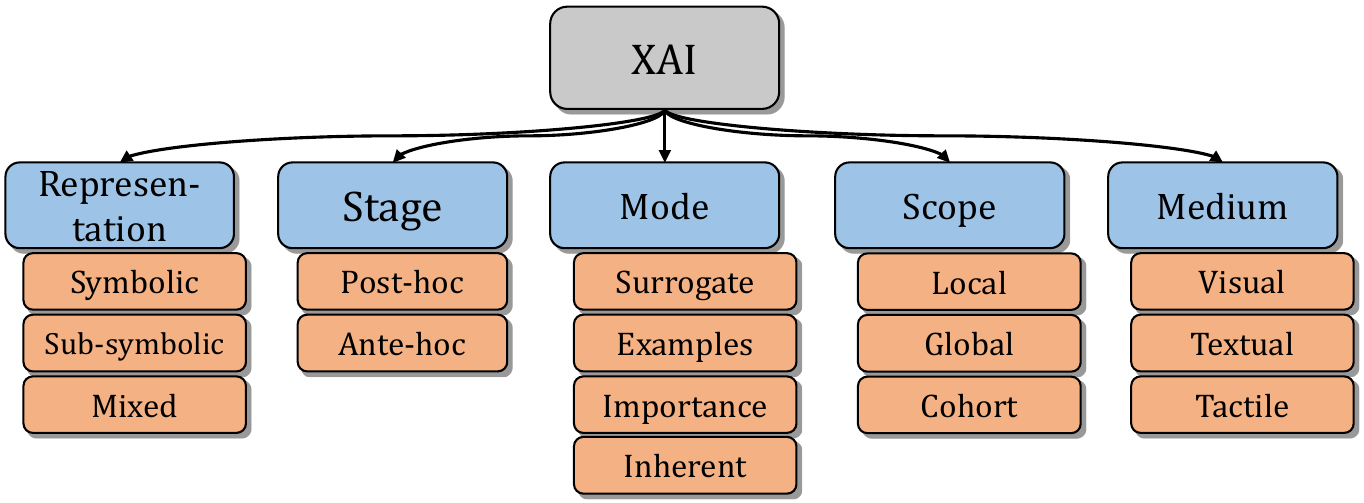}
    \caption{A taxonomy of XAI covering the most important concepts occurring in the literature of XAI for AD with terminology borrowed from Speith~\cite{speith2022review}.}
    \label{fig:taxonomy}
\end{figure}

We now provide a taxonomy of XAI visualised in~\cref{fig:taxonomy} which is used later to describe the reviewed methods in~\cref{sec:cat}.
Our six taxonomic categories are based on Speith~\cite{speith2022review}.

\textbf{Representation} of information within the decision-making model has a significant effect on the working of the XAI system.
We can differentiate between \textit{symbolic} (e.g., rule-based system, decision tree, etc.) and \textit{sub-symbolic} systems (e.g., deep learning), as well as \textit{mixed} systems that utilise both. 
Note that this category is usually determined by the design of the decision-making system, not the XAI system.
As discussed in~\cref{ssec:background:trustworthyAI}, it is more difficult to explain sub-symbolic systems and it may be more difficult to build trustworthy and safe systems that utilise such a representation.

\textbf{Stage} relates to when during the decision-making process an explanation is generated and from what representations. 
\textit{Post-hoc} XAI systems are run after a decision has been made and are  widely applicable to any AI method regardless of representation.
These methods are usually assumed to have access only to the decision-making system and its input and output.
In contrast, \textit{ante-hoc} XAI systems are constrained to AI methods with symbolic or mixed representations as these XAI systems generate explanations directly from the information represented in the decision-making process which would not be possible with a black box system.
These systems have more constrained applicability but are generally more trustworthy and verifiable.

\textbf{Mode} determines the syntactic and semantic form of the explanation.
While there is a large range of explanatory modes, three are particularly popular in XAI.
\textit{Surrogate} systems condense the overall workings of a more complex method into an interpretable model, however, it is difficult to quantify to what extent these models can faithfully represent their parent models.
In addition, representative and counterfactual \textit{examples} are a mode to explain some aspect of the decision-making process in terms of an input example, but these rely on the assumption that the user can understand and interpret the example correctly.
\textit{Importance}-based explanations explain which features of the input representations are most influential for the model when it makes a prediction. 
These models provide a way to shed some light on black box decision making but importance must not be conflated with actual causality as they can often be altered without affecting the output prediction~\cite{jain2019attention,kumarProblemsShapleyvaluebasedExplanations2020a}.
Finally, for interpretable ante-hoc methods the system itself \textit{inherently} serves as the mode of explanation, though understanding a model as the explanation itself requires significant cognitive processing and is unlikely to contribute to trustworthiness in most stakeholders.

\textbf{Scope} determines whether the explanation applies to a given input instance only (\textit{local}), to a group of instances (\textit{cohort}), or to the entire model as a whole (\textit{global}).
The scope of the explanation is tightly connected to its mode.
Example- and importance-based explanations are more suited for local and cohort explanations while surrogate models represent the entire decision-making process and are, thus, global explainers.

\textbf{Medium} is the channel through which the explanations are \textit{intelligibly} delivered to the stakeholders.
How explanations are delivered has a profound influence on efficacy and intelligibility.
It is a crucial design consideration and should complement a correct understanding of stakeholder requirements.
Unfortunately, it is a prevailing trend in XAI to offer an explanation ``as is'' (e.g., feature importance plots, decision tree visualisations, etc.) without further regard to how it should be communicated.

\subsection{Terminology of AD Components}

\begin{figure}
    \centering
    \includegraphics[width=0.85\linewidth]{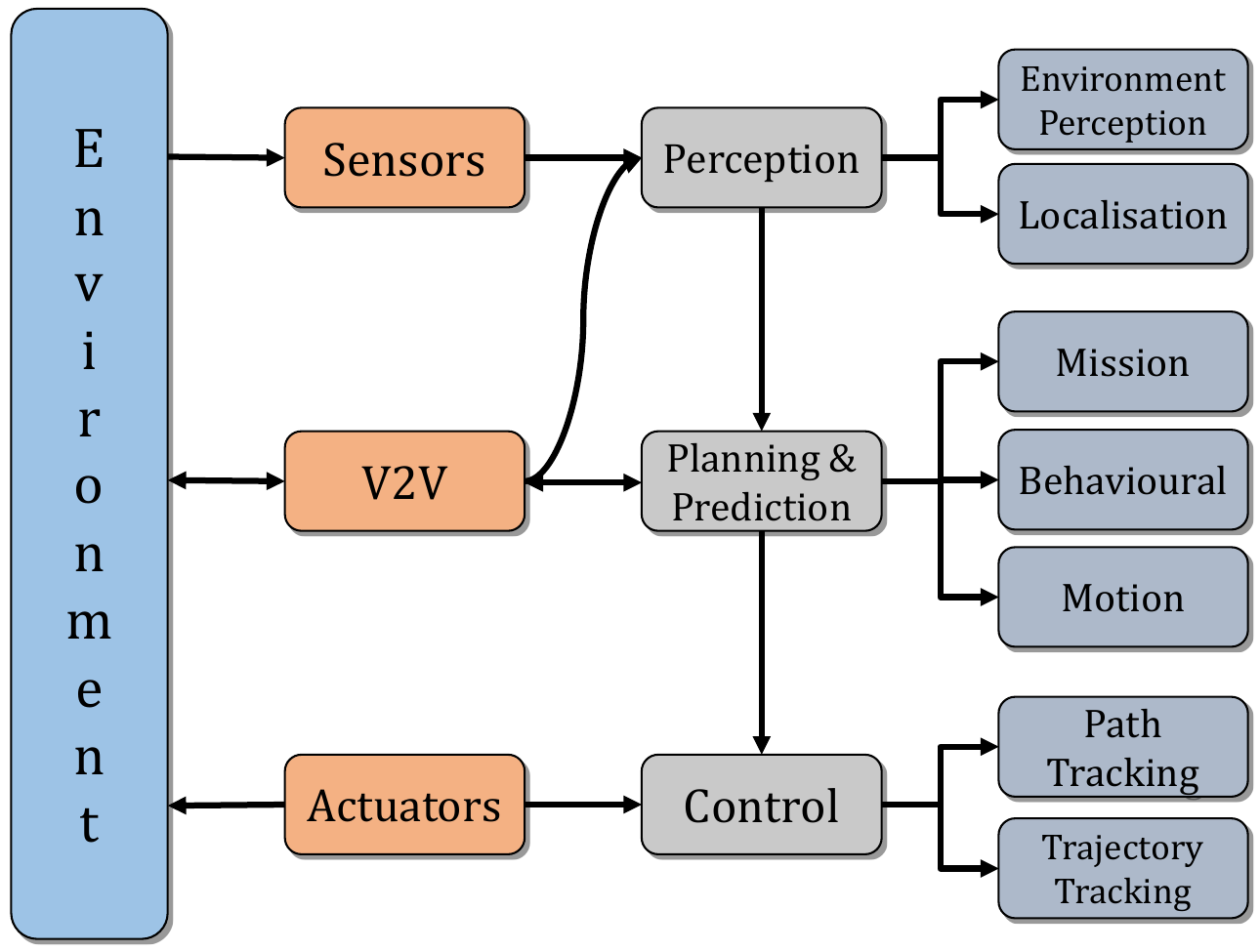}
    \caption{A typical system overview of autonomous driving systems\cite{pendleton2017perception}. Arrows denote the flow of information. Orange boxes are hardware, grey boxes are software components. (V2V: vehicle-to-vehicle communication.)}
    \label{fig:ADComponents}
\end{figure}

Although different divisions of AD components exist depending on the level of detail~\cite{van2018autonomous}, the core competencies of an AV can be generally categorized into three components, which are perception, planning and prediction, and control~\cite{pendleton2017perception,tuncali2019requirements}, as illustrated in \cref{fig:ADComponents}. 
Perception is the capability to gather data from the surroundings and derive meaningful insights or knowledge from that environmental information. 
Specifically, \textit{environmental perception} refers to the development of a contextual understanding of the environment, which encompasses the identification of obstacles, detection of road signs and markings, and classification of data based on their semantic significance.
\textit{Localisation} refers to the ability of the AV to determine its position within the environment. 

Planning and prediction involve the strategic process of making informed decisions based on predicted future trajectories of obstacles to achieve the vehicle's higher-order goals. 
This typically includes navigating the vehicle from a starting point to a desired destination, while simultaneously avoiding obstacles and optimizing performance based on pre-designed constraints. 
According to~\cite{paden2016survey}, planning can be further divided into mission planning, behaviour planning and motion planning.
\textit{Mission planning} represents the selection of a route from its current position to the predefined destination based on the road network.
\textit{Behavior planning} is responsible for determining the appropriate driving behaviour at any point of time along the selected route, given the perceived behaviour of other traffic participants and road conditions, etc.
Lastly, \textit{motion planning} aims to find a collision-free, comfortable, and dynamically feasible path or trajectory once the behaviour layer signals a driving behaviour in the current driving context.

Finally, control competency denotes its proficiency in executing planned actions, which are formulated by its higher-level processing modules.
In \textit{path tracking}, the vehicle is required to converge to and follow a path generated by motion planning without including a temporal law~\cite{altafini2002following}. 
In contrast, \textit{Trajectory tracking} refers to the following feasible "state-space" trajectories, which specify the time evolution of the potion, orientation, and linear and angular velocities~\cite{frazzoli2000trajectory}. 

The above modular system description enables the separate development of each component. 
In addition to modular approaches, there are E2E systems that replace the AD architecture with a single neural network~\cite{zablocki2022explainability}, though often the control part is separated and the E2E network only comprises the planning and perception components. 
The motivation for E2E architectures relies on its simple design by avoiding the consideration of different interconnections between different modules and instead focusing on joint feature optimization of individual modules~\cite{chen2023endtoend}.
In contrast to modular pipelines, E2E networks are much less interpretable, so ensuring their safety is more challenging. 
It is easier to trace the source of errors in modular approaches~\cite{Tampuu2022endtoend}.

\section{Review Methodology}\label{sec:methodology}
Considering the requirements and challenges in implementing XAI for AD, the field has been growing in popularity.
To comprehensively explore the published methods, we perform a systematic literature review following the recommendations of Kitchenham and Charters~\cite{kitchenhamGuidelinesPerformingSystematic2007} and the review methodology section of Stepin et al.~\cite{stepinSurveyContrastiveCounterfactual2021}.
A structured review allows us to systematically explore the field by combining increasingly more fine-grained queries with online indexing databases, while our description of this process enables the repeatability of our search which can verify the validity of our work and help obtain an updated look of the field in the future.

To give an overview of the review process, first, we defined two primary research questions based on which we developed a query hierarchy.
We used the resulting queries to search three indexing databases -- Scopus, Web of Science, and IEEE Xplore -- and applied a three-step process to arrive at a final set of 84 publications. 
We describe the full process below.

\subsection{Research Questions}
\begin{itemize}
    \item[\textbf{RQ1}] What are the current methods of XAI that address requirements of safety and/or trustworthiness, and what are their key contributions to meeting these requirements?
    \item[\textbf{RQ2}] What concrete general frameworks are proposed for integrating XAI with autonomous driving?
\end{itemize}

\subsection{Search Process}
\begin{figure}
    \centering
    \includegraphics[width=\linewidth]{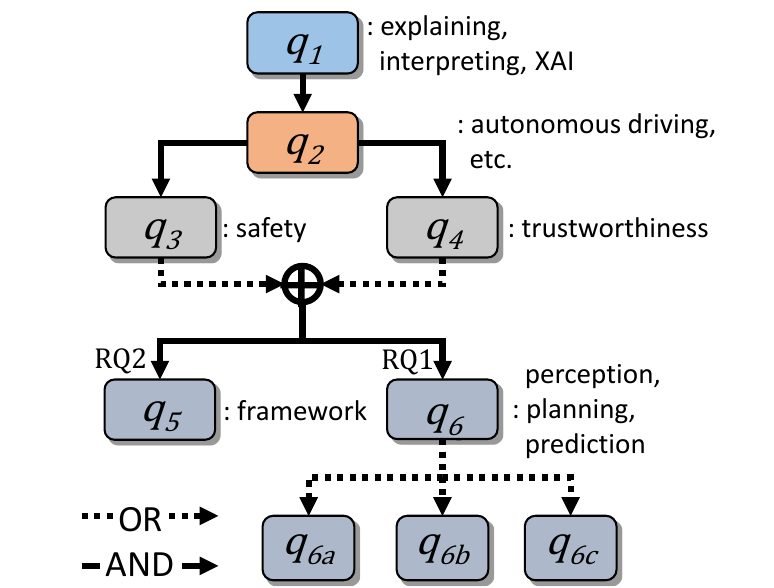}
    \caption{The query hierarchy used for the survey with a representative list of keywords corresponding to each query. Colours signal various depths of the search hierarchy.}
    \label{fig:query}
\end{figure}

We chose the Scopus, Web of Science (WoS), and IEEE Xplore online indexing databases to perform our review, as these platforms provide extensive coverage of both technical and non-technical venues as well as the ability to construct and refine detailed queries.
To obtain a list of candidate papers, we constructed a search hierarchy as shown in~\cref{fig:query}.
Each level of depth in this tree corresponds to increasingly more refined search terms such that the final list of candidate papers was a set of highly relevant publications with manageable counts.
The queries are shown below in the WoS notation, and equivalent queries were constructed for both Scopus and IEEE Xplore. 
The queries were applied to the title, author keywords, and abstract field of each indexed publication, and the search was carried out between 22 to 26 September 2023. 

\begin{itemize}
    \item $q_1$: expla* OR interp* or XAI
    \item $q_2$: $q_1$ AND (auto* AND (driv* OR vehicle* OR car*) OR self driving);
    \item $q_3$: $q_2$ AND safe*;
    \item $q_4$: $q_2$ AND trust*;
    \item $q_5$: ($q_3$ OR $q_4$) AND (pipeline OR architecture OR framework);
    \item $q_6$: ($q_3$ OR $q_4$) AND \dots
    \begin{itemize}
        \item $q_{6a}$: sense OR perception OR computer vision OR object detection OR semantic segmentation;
        \item $q_{6b}$: prediction OR plan*;
        \item $q_{6c}$: control*.
    \end{itemize}
\end{itemize}

Our choice for $q_1$ selects all papers that are related to explaining, interpretation, or any papers that mention XAI.
At this point, we did not constrain our search with keywords relating to a particular subject area (e.g., autonomous driving) to build a large foundation of papers to select from.
We narrowed our search to focus on autonomous driving (and related keywords) using $q_2$, and then further filtered papers based on whether they contain keywords relating to trust or safety.
To answer \textbf{RQ1}, we take this set of papers and sort them based on whether they relate to a particular subsystem of the AD stack as shown in~\cref{fig:ADComponents}.
To answer \textbf{RQ2}, we filter the collected set of papers based on keywords that relate to frameworks or architectures.

\begin{figure}
    \centering
    \includegraphics[width=\linewidth]{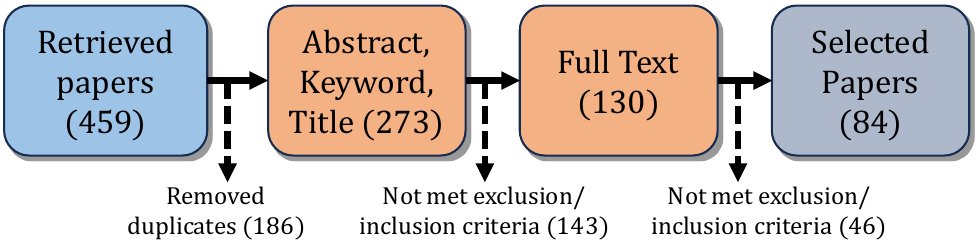}
    \caption{Overview of the review process as a flowchart. First, papers were retrieved according to the query hierarchy (blue box), then twice filtered by content (orange boxes). Numbers in parentheses show papers at the end of each step.}
    \label{fig:flowchart}
\end{figure}

\begin{table}[]
    \caption{The number of papers collected for queries corresponding to \textbf{RQ1} ($q_6$) and \textbf{RQ2} ($q_5$), while not showing $q_{1-4}$ as these queries had multiple thousands of papers.}
    \centering
    \begin{tabular}{llllll}
    \toprule
         {}   & WoS & IEEE & Scopus & Duplicates & Total (w/o dups.) \\
         \midrule
         \textbf{RQ1} &  130 & 135 & 169 & 178 & 256 \\
         \textbf{RQ2} & 7 & 11 & 7 & 8 & 17 \\
         \bottomrule
    \end{tabular}
    \label{tab:paper-counts}
\end{table}

The search and selection process was conducted as indicated in~\cref{fig:flowchart} and explained below. 
The total numbers of papers retrieved for the research questions are shown in~\cref{tab:paper-counts}.
After querying for papers we have removed all duplicate papers. 
The remaining set was then filtered based on our exclusion and inclusion criteria (detailed in~\cref{ssec:method:criteria}).
We then proceeded to filter the remaining papers based on their full text and re-applied the same exclusion and inclusion criteria to determine which papers to include in our final list.

\subsection{Inclusion and Exclusion Criteria}\label{ssec:method:criteria}
We now describe the inclusion and exclusion criteria that were used for both research questions at each stage of the search process to arrive at the final list.
At each stage in the filtering process, we first applied a list of inclusion criteria to determine which papers to keep at that stage.
All of these inclusion criteria must have been fulfilled by the paper to pass this stage.
We included papers where:
\begin{itemize}
    \item The paper was -- in part or fully -- motivated by a need for safer or more trustworthy technologies; AND
    \item The paper proposed a concrete system, algorithm, framework, or novel artefact related to artificial intelligence;
\end{itemize}

After the inclusion process, we applied a list of exclusion criteria which specified more detailed requirements on the papers.
We filtered out papers if they met at least one of the exclusion criteria.
We excluded papers where:
\begin{itemize}
    \item It showed no attempt to address any of the sources of explanations (as described in~\cref{ssec:background:sources}); OR
    \item The main domain of application or evaluation was not autonomous driving; OR
    \item The paper did not address perception, planning, prediction, or control for autonomous driving;
\end{itemize}

\section{XAI For Safe and Trustworthy AD}\label{sec:cat}
\begin{figure*}
    \centering
    \includegraphics[width=\linewidth]{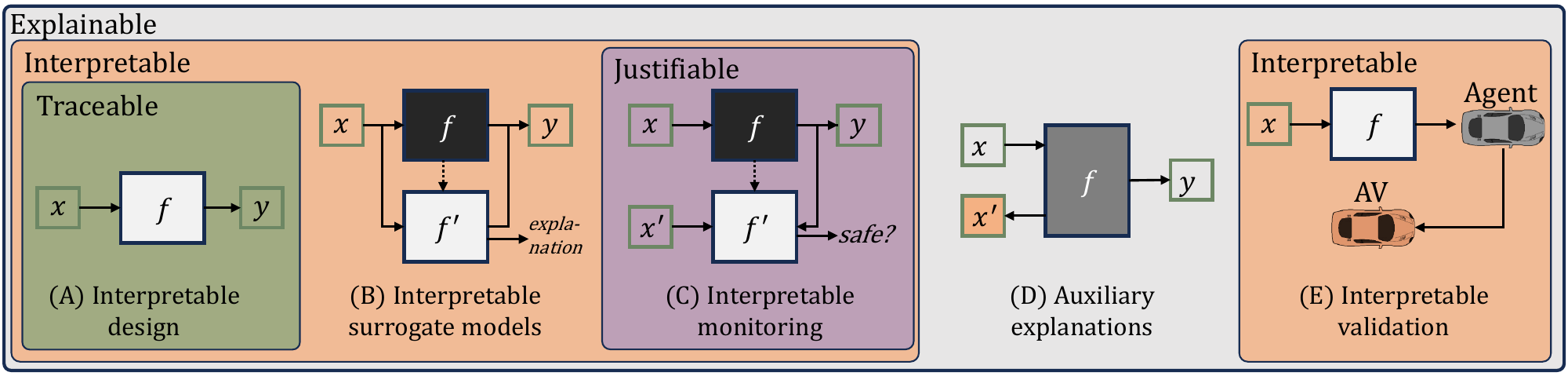}
    \caption{Types of XAI methods for AD as grouped by the sources of explanations (\cref{ssec:background:sources}). \textbf{(A)} the AI algorithm is inherently interpretable (white box); \textbf{(B)} surrogate models are used to approximate opaque AI models (black box) and generate explanations for their outputs; \textbf{(C)} a transformed input is fed into an interpretable monitor system (white box) to runtime check the safety of opaque AI models (black box); \textbf{(D)} the AI model can deliver some intermediate information as explanation of its designed functions (gray box); \textbf{(E)} interpretable (white box) algorithms are employed to control agents for validating AVs.}
    \label{fig:XAIModels}
\end{figure*}

We found five main categories into which we sorted the reviewed papers.
These are interpretable design, interpretable surrogate models, interpretable monitoring, auxiliary explanations, and interpretable validation.\footnote{In contrast to our definitions of the sources of explanation in~\cref{ssec:background:sources}, Du et al.~\cite{Du2019} grouped interpretable machine learning into intrinsic and post-hoc interpretability. Our collected publications include both types of methods.}
An overview of the advantages and disadvantages of methods for each category is given in~\cref{tab:method-summary} and visual illustrations of these categories are shown in~\cref{fig:XAIModels}.
In this section, we present an overview of different methods in each category and analyse their relevance to achieving safer and more trustworthy AD.
For each category -- except interpretable validation which has only five works -- we present a table of summary (\cref{tab:inherently-interpretable,tab:interpretable-surrogate,tab:interpretable-monitoring,tab:auxiliary-explanations}) of methods according to the XAI taxonomy presented in~\cref{ssec:background:XAI-taxonomy}.

\begin{table*}
    \centering
    \caption{Definition, advantages (\textcolor{ForestGreen}{+}), and disadvantages (\textcolor{BrickRed}{--}) of each of the five categories of reviewed methods.}
    \label{tab:method-summary}    
    \begin{tabular}{@{}lp{14cm}@{}}
         \toprule
         \textbf{Method} & \textbf{Summary} \\
         \midrule

         \multirow{2}{3.5cm}{\textsc{Interpretable by design} (\cref{ssec:cat:interp})} & Inherently interpretable method whose design reveals the explicit causal relationship between input(s) and output(s). Often these methods rely on decision trees, Bayesian networks, interpretable latent space, or rule-based algorithms. \\
          {} & \textcolor{ForestGreen}{+ May be used to verify formal claims of safety about the algorithm;} \\
          {} & \textcolor{ForestGreen}{+ Uses meaningful state abstractions which offer intelligible explanations.} \\
          {} & \textcolor{BrickRed}{-- Requires extensive domain knowledge to engineer meaningful representations or high-level driving maneuvers;} \\
          {} & \textcolor{BrickRed}{-- Analysing interpretable systems requires significant AI expertise which is not available to all stakeholders.} \\
          \midrule

         \multirow{2}{3.5cm}{\textsc{Interpretable surrogate model} (\cref{ssec:cat:surr})} & An interpretable-by-design algorithm approximates the behaviour of a black box model such that its primary goal is to provide intelligible explanations of the black box model to some stakeholder(s). \\
          {} & \textcolor{ForestGreen}{+ Perception errors can be analyzed thoroughly contributing to safety analyses;} \\
          {} & \textcolor{ForestGreen}{+ Can apply many readily applicable tools from XAI (e.g., SHAP, LIME, GradCAM);} \\
          {} & \textcolor{ForestGreen}{+ May combine with natural language generation to create intelligible explanations.} \\
          {} & \textcolor{BrickRed}{-- Feature attribution methods can be inconsistent, hard to interpret, and incorrect;} \\
          {} & \textcolor{BrickRed}{-- Cannot give formal guarantees on safety.} \\
          \midrule

         \multirow{2}{3.5cm}{\textsc{Interpretable monitoring} (\cref{ssec:cat:monitor})} & An interpretable-by-design model is used to verify the decision-making algorithm's output such that it ensures safer AI deployment for AVs.  \\
          {} & \textcolor{ForestGreen}{+ Greatly improves (perceived) safety of the AV;} \\
          {} & \textcolor{ForestGreen}{+ Can be deployed with existing decision-making/perception systems.} \\
          {} & \textcolor{BrickRed}{-- May introduce prohibitive computational overhead;} \\
          {} & \textcolor{BrickRed}{-- Not suited for standalone deployment.} \\
          {} & \textcolor{BrickRed}{-- May fail to generalise or correctly identify unsafe actions if the interpretable model is too simple, incorrect, or biased.} \\         
          \midrule

         \multirow{2}{3.5cm}{\textsc{Auxiliary explanations} (\cref{ssec:cat:aux})} & The execution of the decision-making algorithm creates auxiliary information that provides information about how the algorithm works. Methods using attention mechanisms are very common here. \\
          {} & \textcolor{ForestGreen}{+ Applicable to most decision-making algorithms;} \\
          {} & \textcolor{ForestGreen}{+ Low overhead on generation as the explanation is a by-product of the decision-making process.} \\
          {} & \textcolor{BrickRed}{-- Usually does not reveal enough information about the decision-making process to improve safety or trustworthiness;} \\
          {} & \textcolor{BrickRed}{-- Requires careful manual analysis to interpret;} \\   
          {} & \textcolor{BrickRed}{-- Attention-based auxiliary explanations may look plausible despite being incorrect.} \\   
          {} & \textcolor{BrickRed}{-- Heat maps can be fragile and unreliable} \\ 
          \midrule

         \multirow{2}{3.5cm}{\textsc{Interpretable safety validation} (\cref{ssec:cat:valid})} & Provides an interpretable way to generate adversarial behaviours of surrounding agents for the validation of AVs. \\
          {} & \textcolor{ForestGreen}{+ Greatly improves safety of AV;} \\
          {} & \textcolor{ForestGreen}{+ Extracts unique failuer and challenge scenarios that can be used for further AD assessment.} \\
          {} & \textcolor{BrickRed}{-- Very challenging to generate unique scenarios;} \\
          {} & \textcolor{BrickRed}{-- Applicable only during pre-deployment due to excessive runtime requirements.} \\         
         \bottomrule
    \end{tabular}
\end{table*}

\subsection{Interpretable By Design}\label{ssec:cat:interp}
\begin{definition}[Interpretable By Design]
\textit{We call an algorithm interpretable by design if it is inherently interpretable such that its design reveals the explicit causal relationship between its input(s) and output(s)~\cite{molnarInterpretableMachineLearning2023}.}
\end{definition}

\begin{table*}
    \centering
    \caption{Summary of interpretable-by-design methods using the XAI taxonomy of~\cref{ssec:background:XAI-taxonomy}. All methods in this category are ante-hoc (Stage). Evaluation methods are based on fixed simulated scenarios, randomised simulations, or datasets (names listed). Missing entry (---) means not applicable. No E2E methods were found for this category. *~Conceptual framework}
    \label{tab:inherently-interpretable}
    \begin{tabular}{@{}llp{2.75cm}llllp{2.45cm}l@{}}
         \toprule
         {} & \textbf{Paper} & \textbf{Method} & {}    & {}   & {}        & {} & \textbf{Evaluation}      & {}       \\
         {} &             &  \textit{Task}      & \textit{Representation}  & \textit{Mode} & \textit{Scope} & \textit{Medium} & \textit{Method}               & \textit{User study} \\ 
         \midrule
         \multirow{7}{*}{\textsc{Perception}} & 
         \cite{chaghazardiXAITrafficSign} & traffic sign detection & symbolic & inherent & global & --- & self-curated dataset & no \\
         {} &
         \cite{FeifelSafetyImpactInterpretableDNN} & pedestrian detection & subsymbolic & inherent & local & --- & CityPersons~\cite{Cordts2016Cityscapes} & no \\ 
         {} & \cite{Losch2021SB} & semantic segmentation & subsymbolic & inherent & global & --- & CityScapes~\cite{Cordts2016Cityscapes} & no \\ 
         {} & \cite{PlebeAutoencoder} & semantic segmentation & subsymbolic & inherent & global & --- & SYNTHIA~\cite{ros2016synthia} & no \\ 
         {} & \cite{Oltramari2020NeurosymbolicAF} & context understanding & mixed & inherent & local & --- & nuScenes~\cite{nuscenes2019} & no \\
         {} & \cite{MartinezCapsulate} & eye fixation & subsymbolic & inherent & global & --- & DR(eye)VE~\cite{palazzi2019dreye} & no \\
         {} & \cite{YonakaSunGlare} & sun-glare recognition & mixed & inherent & local & --- & self-curated dataset & no \\
         \midrule
         \multirow{13}{1.5cm}{\textsc{Planning \& \newline Prediction}} & \cite{albrechtInterpretableGoalbasedPrediction2021,hannaInterpretableGoalRecognition2021} & motion planning, trajectory prediction & symbolic & inherent & global & --- & scenarios & no \\
         {} & \cite{antonello2022flash} & trajectory prediction & mixed & inherent & global & --- & NGSIM~\cite{NGSIM} & no \\
         {} & \cite{brewitt2021grit,brewitt2023ogrit} & goal prediction & symbolic & inherent  & global, local & --- & inD~\cite{inDdataset}, rounD~\cite{rounDdataset}, openDD~\cite{breuer2020opendd} & no  \\ 
         {} & \cite{ghoulInterpretableGoalBasedModel2023} & trajectory prediction & mixed & importance & local & visual & INTERACTION~\cite{interactiondataset} & no \\
         {} & \cite{gyevnar2022humancentric}          & motion planning & symbolic & surrogate & global & textual & scenarios & no \\
         {} & \cite{omeiza2022spoken} & motion planning & symbolic & inherent & global & textual & self-curated dataset & yes \\ 
         {} & \cite{henzeHowCanAutomated2022}* & motion planning & mixed & importance & local & textual & --- & yes \\ 
         {} & \cite{kleinInterpretableClassifiersBased2023} & lane-change prediction & mixed & importance & local & --- & HighD~\cite{highDdataset} & no \\
         {} & \cite{kridalukmanaSelfExplainingAbilitiesIntelligent2022} & motion planning & symbolic & importance & global & textual & CARLA~\cite{Dosovitskiy17} & yes \\
         {} & \cite{muschollEMIDASExplainableSocial2021a} & pedestrian prediction & symbolic & importance & global & --- & scenarios & no \\
         {} & \cite{neumeierVariationalAutoencoderBasedVehicle2021a} & lane-change prediction & mixed & inherent & local & --- & HighD~\cite{highDdataset} & no \\
         {} & \cite{wuHybridDrivingDecisionMaking2023} & lane-change prediction & mixed & surrogate & global & --- & simulation & no \\
         \midrule
         \textsc{Control} & \cite{zheng2023towards} & safe control & mixed & inherent & local & --- & simulation & no \\
         \bottomrule
    \end{tabular}
\end{table*}

\subsubsection{Interpretable By Design -- Perception}
Chaghazardi et al.~\cite{chaghazardiXAITrafficSign} introduced an inductive logic programming approach for traffic sign classification where firstly high-level features such as colour, shape, etc. are extracted and then a hypothesis is learned. The design increases transparency and reliability. Moreover, a higher robustness against adversarial attacks compared to other state-of-the-art algorithms was shown.
In \cite{FeifelSafetyImpactInterpretableDNN}, Feifel et al. proposed a structured interpretable latent space in a DNN for pedestrian detection which learns to extract specific prototypes. The learned prototypes in the latent space can be clustered in a projected 2D-plane via a principal component analysis \cite{PCABishop1999} or a t-SNE projection \cite{tSNELaurenz2008}. Due to the interpretably designed DNN, an ante-hoc analysis is possible which supports the safety argumentation. Plebe et al. \cite{PlebeAutoencoder} developed a temporal autoencoder for lane and car detections in semantic segmentation consisting of an organized latent space where semantic concepts of lane and car segments are learned. Similarly, Losch et al. \cite{Losch2021SB} proposed semantic bottleneck models for semantic segmentation tasks which aligned every channel with a human interpretable visual concept. The introduction of semantic concepts in the latent space additionally increases transparency in the prediction by the DNN.
Oltramari et al. \cite{Oltramari2020NeurosymbolicAF} developed a hybrid AI framework for perceptual scene understanding via instructing the latent space of DNNs with knowledge graphs that are extracted from clustering the labelled training data. 
Martinez et al. \cite{MartinezCapsulate} developed an interpretable latent space in the DNN by using capsule networks \cite{capsulesHinton2011} to predict eye fixations in AD scenarios and contextual conditions. 
With these capsules, it is possible to express interpretable relationships between features and contextual conditions on frame-level and pixel-level.
In \cite{YonakaSunGlare}, Yoneda et al. trained a CNN to identify the presence of sun-glare in the AD environment. Subsequently, heat maps with a Gradient-weighted activation map approach (Grad-CAM) \cite{selvarajuGradCAMVisualExplanations2017} were calculated to identify the regions of sun glare in the image.
The developed heat map approach increases transparency in the decision-making process.

\subsubsection{Interpretable By Design -- Planning \& Prediction}
Methods in this category create explanations for mainly three purposes: goal/trajectory prediction, lane-change intention prediction, and motion planning.
One of the most common design choices is the use of high-level, interpretable driving maneuvers~\cite{albrechtInterpretableGoalbasedPrediction2021,hannaInterpretableGoalRecognition2021,kridalukmanaSelfExplainingAbilitiesIntelligent2022,muschollEMIDASExplainableSocial2021a,brewitt2021grit,brewitt2023ogrit}.
These maneuvers provide a convenient abstraction over lower-level state variables (e.g., acceleration and steering) and render the decision-making process tractable and interpretable.

However, interpretable-by-design approaches rely on extensive domain knowledge which may not be scalable to more complex decision-making.
Moreover, some methods rely on BNs to define probabilistic models of the decision-making process~\cite{kridalukmanaSelfExplainingAbilitiesIntelligent2022,muschollEMIDASExplainableSocial2021a,gyevnar2022humancentric}, which needs a good understanding of and assumptions about the causal processes behind driving decisions. 
The benefit of BNs is that they provide a principled mathematical framework to reason about causality, however, the relience on expert knowledge has the potential to introduce human modelling errors and biases.

Another approach for motion planning is to rely on Monte Carlo Tree Search (MCTS) over high-level driving maneuvers to create a shallow search tree via simulations that is interpretable~\cite{albrechtInterpretableGoalbasedPrediction2021,hannaInterpretableGoalRecognition2021,gyevnar2022humancentric}. 
MCTS has the benefit of covering only relevant parts of the search space while avoiding unsafe actions, however, it relies on trajectory predictions for other traffic participants and is computationally expensive.

Interpretable goal and trajectory prediction have also received attention.
These methods rely either on rational (Bayesian) inverse planning~\cite{antonello2022flash,albrechtInterpretableGoalbasedPrediction2021}, decision trees~\cite{brewitt2021grit,brewitt2023ogrit}, or discrete choice models~\cite{ghoulInterpretableGoalBasedModel2023}.
There is significant work on explaining lane change (LC) predictions using a variety of methods which include time-series motifs~\cite{kleinInterpretableClassifiersBased2023}, rule extraction~\cite{wuHybridDrivingDecisionMaking2023}, and interpretable variational auto-encoders~\cite{neumeierVariationalAutoencoderBasedVehicle2021a}.
One work stands out for predicting pedestrian intentions~\cite{muschollEMIDASExplainableSocial2021a} using a dynamic BN derived from annotated real-world image data.

Only three papers ran a user study~\cite{henzeHowCanAutomated2022,kridalukmanaSelfExplainingAbilitiesIntelligent2022,omeiza2022spoken} and only one of those provided a more thorough investigation of the effects of explanations on users~\cite{omeiza2022spoken}.
This latter work is also one of the only methods in this category which elicited stakeholders' requirements in detail along the axes of intelligibility, accountability, and trust.
However, a major limitation of their proposed method is the use of a highly specialised dataset with annotations for high-level semantic and structured explanations which may not be readily available.

\subsubsection{Interpretable By Design -- Control}
Zheng et al.~\cite{zheng2023towards} proposed an ante-hoc explainable controller in which the output of a neural network-based controller with control barrier function filters is projected onto a safe set in an interpretable manner via quadratic programs and a gauge map.
Their method provides good traceability of the control process and discusses in-depth the effects of interpretable control on safety validation.

\subsubsection{Interpretable By Design -- Summary}
In the field of perception, the majority of algorithms that are interpretable by design rely on the construction of an interpretable latent space within the DNN. By compelling the algorithm to learn semantic concepts, the automatic feature extraction becomes more interpretable. Another possibility is to modularize the perception algorithm into multiple algorithms that learn to identify different semantic concepts. By training the perception algorithm to extract semantic concepts, the algorithm is forced to learn an interpretable feature extraction. However, it is challenging to define the concepts. Additionally, a dataset which contains semantic concepts of an object as a label is often required. Lastly, it is also possible to use the interpretability of a classifier to localize objects~\cite{YonakaSunGlare}.  

In planning and prediction, it is common to hand-craft high-level interpretable features or maneuvers to abstract the low-level state space.
These abstractions may then be used in various algorithms for decision-making, such as Monte Carlo Tree Search (MCTS), Bayesian networks (BN), or decision trees.
However, creating these abstractions requires significant domain knowledge and careful engineering.
For the control task, constraint-satisfaction algorithms can be applied to map the output of a DNN onto a safe-by-construction control domain in an interpretable way.

These methods are intrinsic explainable and have the potential to improve both the trustworthiness and the safety of AD. 
Interpretable algorithms provide a clear causal link between the input and the output of the algorithm which may enable safety validation, while meaningful high-level abstractions may be easily understood by people which can contribute to accurate trust calibration.
Unfortunately, very few methods investigate the algorithms' efficacy with actual stakeholders leaving many of the motivating claims of these works unaddressed.
This is especially clear from the lack of methods that generate concrete, intelligible explanations (i.e., explainable systems as defined in~\cref{ssec:background:sources}).
Instead, papers usually offer scientific analyses of the interpretable components of their systems, but these are impossible to scale for varied stakeholders and do not consider the requirements for achieving more trustworthy AD (cf.~\cref{ssec:background:requirements}).
The missing motivation for trustworthiness is also clear from the lack of user studies that evaluate the benefit of explanations on stakeholders.

\subsection{Interpretable Surrogate Models}\label{ssec:cat:surr}
\begin{definition}[Interpretable Surrogate Models]
\textit{We call a system an interpretable surrogate model if an interpretable-by-design model approximates the behaviour of a black box algorithm such that this provides intelligible explanations of the black box algorithm~\cite{schwalbe2023comprehensive}.}
\end{definition}

\begin{table*}[]
    \centering
    \caption{Summary of interpretable surrogate methods using in part the taxonomy of~\cref{ssec:background:XAI-taxonomy}. All methods in this category are post-hoc (Stage) and subsymbolic (Representation). The Surrogate field refers to the specific surrogate method used to approximate the underlying black box. Missing entry (---) means not applicable.}
    \label{tab:interpretable-surrogate}
    \begin{tabular}{@{}lllllllp{2cm}l@{}}
         \toprule
         {} & \textbf{Paper} & \textbf{Method}   & {} & {}   & {}      & {}   & \textbf{Evaluation}    & {}       \\
         {} &             &  \textit{Task}     & \textit{Surrogate} & \textit{Mode} & \textit{Scope} & \textit{Medium} & \textit{Method}               & \textit{User study} \\ 
         \midrule
         \textsc{Perception} & \cite{ponn2020identification} & object detection & SHAP, RF & importance & local & --- & nuScenes~\cite{nuscenes2019} & no \\
         \midrule
         \multirow{6}{1.5cm}{\textsc{Planning \& \newline Prediction}} & \cite{cui2022interpretation} & vehicle following & SHAP, RF &  importance & global, local & --- & simulation & no \\
         {} & \cite{liExplainingMachineLearningLane2023} & lane-change prediction & max-entropy SHAP & importance & local & --- & HighD~\cite{highDdataset} & no \\
         {} & \cite{maLaneChangeAnalysis2021} & lane-change prediction & mean impact value & importance & local & --- & HighD~\cite{highDdataset} & no \\
         {} & \cite{omeiza2021explanations} & motion planning & decision tree & surrogate & global & textual & scenarios & yes \\
         {} & \cite{mishra2022not} & route planning & decision tree & surrogate & local & visual & simulation & yes \\
         {} & \cite{gyevnar2024cema}                  & motion planning & cognitive model & surrogate & global & textual & HEADD~\cite{gyevnar2024headd} & yes \\
         \midrule
         \textsc{Control} & \cite{xAI_control_Dassanayake} & position control & clustering & surrogate & cohort & --- & simulation & no \\
         \midrule
         \multirow{3}{1.5cm}{\textsc{E2E}} & \cite{Zemni2023OCTET} & action selection & generative model &importance & local & --- & BDD100K~\cite{yu2020bdd100k}, BDD-OIA~\cite{xAI_ObjectInduced_xu} & yes \\
         {} & \cite{shi2020self} & action selection & DNN & importance & local & --- & simulation & no \\
         \bottomrule
    \end{tabular}
\end{table*}

\subsubsection{Interpretable Surrogate Models -- Perception}
Ponn et al.~\cite{ponn2020identification} introduced a model-agnostic surrogate model for camera-based object detectors. 
A random forest is trained to predict a detection score according to meta-information about the environment in the training data. 
Afterwards, Shapley values are calculated to measure the impact of different features from the meta-information which helps interpret the results, such that the behaviour of the object detector under influencing factors in the environment can be estimated.

\subsubsection{Interpretable Surrogate Models -- Planning  \& Prediction}
Cui et al.~\cite{cui2022interpretation} combined SHAP and random forests to increase the transparency of decision-making driven by a deep reinforcement learning (DRL) algorithm. 
In their framework, SHAP determines the important features associated with the decision made by the DRL algorithm and an RF model is trained using these features to explain the decisions of the original DRL model. 
In addition, Li et al.~\cite{liExplainingMachineLearningLane2023} also relied on SHAP to understand the importance of features for LC predictions.
They propose a modified version of SHAP called Maximum Entropy SHAP (ME-SHAP) that they use to explain an XGBoost-based LC decision model.
Their evaluation shows that the ME-SHAP feature contributions may be rationalised in terms of intuitive driving actions, but the qualitative benefits of ME-SHAP for human understanding are not substantiated.  

Three works stand out which avoid using SHAP while also carrying out significant user studies.
First, Mishra et al.~\cite{mishra2022not} used a decision tree to explain an RL agent's actions based on states and corresponding actions determined by the optimal policy. 
They created a visual explanation interface and evaluated it with both students and experts using a wide range of qualitative and quantitative analyses.
They showed that their method is effective, applicable with experts, and more effective than textual explanations.
 
Second, Omeiza et al.~\cite{omeiza2021explanations} created natural language explanations by deriving decision trees from scene graphs.
Their algorithm is based on pre-defined meaningful features but their method description is limited which makes it difficult assess how well it would work in unseen scenarios.
An extensive user study was used to measure the effects of explanations on the perceived accountability of the AD system and on users' understanding of how the AD system works.

Finally, Gyevnar et al.~\cite{gyevnar2024cema} proposed a method called CEMA which is based on a cognitive model of how people select causes for explanations. 
They also generated natural language explanations, but unlike the previous methods which used a decision tree, their method relied on simulations based on a probabilistic planner.
They generated counterfactual worlds that were used to analyse the causal relationships affecting the motion planning of the AV.
They evaluated their explanations with more than 200 online participants against a baseline of human-written explanations called HEADD~\cite{gyevnar2024headd}.

\subsubsection{Interpretable Surrogate Models -- Control}
Surrogate models of control should be strongly focused on safety given their safety-critical hardware-level application.
We identified here one work~\cite{xAI_control_Dassanayake} in which the authors analysed the behaviour of a DNN-based controller that stabilizes the dynamic position of an AV under disturbing environmental conditions.
A cross-comparable clustering method for the time series data was introduced to interpret a response signal from a neural network, such that the internal model understanding and transparency were increased.
However, the specifications of the underlying neural network is not explained, which makes their claims hard to reproduce and generalise.

\subsubsection{Interpretable Surrogate Models -- End-to-end}
Zemni et al. \cite{Zemni2023OCTET} proposed an object-centric framework which generates counterfactual explanations for E2E decision models. 
The E2E decision model was designed to have an instance-based latent representation. 
Thereby, the generative model was able to produce new images with slightly changed objects from the original input image. By analyzing changes in the output, the framework helps to understand the influence of objects in the environment on the decisions of the network.
Shi et al.~\cite{shi2020self} proposed a self-supervised interpretable framework to produce an attention mask corresponding to the importance assigned to each pixel, which constitutes the most evidence for an agent's decisions. The core concept of the framework is a separate explanation model trained for vision-based RL.

\subsubsection{Interpretable Surrogate Models -- Summary}
An interpretable surrogate model consists of a meta-model with interpretability capabilities that approximates the behaviour of a black box model and thus supports understanding the internal working principle of a black box model. 
For perception, this may be done by training a different ML model, such as a random forest that can be interpreted by inspection or via, for example, Shapley values~\cite{ponn2020identification}. This type of interpretable surrogate model has the potential to increase the transparency and reliability of the network, as detection errors can be analyzed more thoroughly, thereby enhancing the understanding of the model's behaviour. 
It is similarly common to use Shapley values in planning and prediction, especially as implemented by SHAP~\cite{lundberg2017SHAP}, and only three works did not rely on SHAP.

Unfortunately, blindly relying on SHAP-based interpretability may not help achieve safer AD as this method is susceptible to a variety of issues that lead to inconsistent explanations when compared with other feature saliency methods~\cite{freyer2021shapley,bilodeau2024impossibility}.
This also means that care must be taken in their use when trying to improve the trustworthiness of AVs, and user studies with relevant stakeholders are essential to validate claims about trustworthiness.

For the control task, the output of a DNN can be interpreted in a post-hoc manner by analyzing the data over time with a clustering approach~\cite{xAI_control_Dassanayake}. In E2E learning, a separate generative model can be trained via the black box model \cite{Zemni2023OCTET}. The generative model can then be used to provide counterfactual explanations of the model. Moreover, it is possible to train a separate explanation module of the E2E network \cite{shi2020self}.

\subsection{Interpretable Monitoring}\label{ssec:cat:monitor}

\begin{table*}[]
    \centering
    \caption{Summary of interpretable monitoring methods using the taxonomy of~\cref{ssec:background:XAI-taxonomy}. No user studies were done in this category. Missing entry (---) means not applicable. No Control and E2E methods were identified in this category.}
    \label{tab:interpretable-monitoring}
    \begin{tabular}{@{}cllllllll@{}}
         \toprule
         {} & \textbf{Paper} & \textbf{Method} & {}    & {} & {}   & {}        &  {}      & \textbf{Evaluation}       \\
         {} &             &  \textit{Task}      & \textit{Representation}  & \textit{Stage} & \textit{Mode} & \textit{Scope} & \textit{Medium} & \textit{Method}  \\ 
         \midrule
         \multirow{2}{*}{\textsc{Perception}} & \cite{kronenberger2020dependency} & traffic sign recognition & subsymbolic & post-hoc & inherent & global & --- & GTSRB \cite{Houben-IJCNN-2013}\\
         {} & \cite{HackerSaliencyPlausibility} & traffic sign recognition & subsymbolic & post-hoc & importance & local & visual & simulation \\
         {} & \cite{keser2023interpretable} & object detection & subsymbolic & post-hoc & surrogate & global & --- & \makecell[lt]{COCO~\cite{lin2014microsoft}, \\ Broden~\cite{bau2017network}, \\ KITTI~\cite{Fritsch2013ITSC}} \\
         {} & \cite{fang2023toward} & anomaly detection & subsymbolic & post-hoc & importance & local & --- & self-curated dataset \\
         \midrule
         \multirow{8}{1.5cm}{\textsc{Planning \& \newline Prediction}} & \cite{baoDRIVEDeepReinforced2021} & accident prediction & mixed & post-hoc & importance & local & visual & DADA-2000~\cite{fang2021dada} \\
         {} & \cite{chen2023attention} & action selection & mixed & post-hoc & importance & local & --- & simulation \\
         {} & \cite{di2020interpretable,gall2021gaussian} & action selection & symbolic & ante-hoc & inherent & global, local & --- & self-curated dataset \\
         {} & \cite{gilpin2021explaining} & anomaly detection & symbolic & ante-hoc & inherent & local & textual & CARLA~\cite{Dosovitskiy17} \\
         {} & \cite{gorospeAnalyzingInterVehicleCollision2023} & collision prediction & symbolic & ante-hoc & inherent & global, local & --- & simulation  \\
         {} & \cite{karimExplainableArtificialIntelligence2022} & accident prediction & subsymbolic & post-hoc & importance & local & visual & CCD~\cite{BaoMM2020} \\
         {} & \cite{nahata2021assessing} & risk scoring & symbolic & ante-hoc & importance & local & --- & Lyft~\cite{houstonOneThousandOne2021} \\
         {} & \cite{schmidt2021can} &  motion planning & mixed & post-hoc & surrogate & global, local & --- & simulation \\
         \bottomrule
    \end{tabular}
\end{table*}

\begin{definition}[Interpretable Monitoring]
\textit{We call a system an interpretable monitoring system if an interpretable-by-design model is used to verify a decision-making algorithm's output such that this ensures safer deployment of AVs.}
\end{definition}

\subsubsection{Interpretable Monitoring -- Perception}
In \cite{kronenberger2020dependency}, Kronenberger et al. examined interpretable DNNs for traffic sign recognition. They introduced additional explanations of visual concepts such as colours, shapes and numbers or symbols. These visual concepts are used to verify the decision of the network. Hacker et al. \cite{HackerSaliencyPlausibility} also proposed a monitor for traffic sign recognition. The monitor consists of various mechanisms including an interpretable saliency detector. During operation, the saliency map is computed via occlusion sensitivity \cite{ZeilerOcclusion2014} and is compared by computing the Euclidean distance to an offline computed saliency map for each traffic sign category. Keser et al. \cite{keser2023interpretable} proposed an interpretable and model-agnostic monitor by introducing a concept bottleneck model (CBM) which is used for a plausibility check with the original DNN-based object detector. The interpretability of CBM is achieved by learning human-interpretable labels. 
Fang et al. \cite{fang2023toward} constructed a fault diagnosis framework to monitor a system's operational status, while the interpretability of the fault diagnosis is achieved by calculating the contribution of each input feature to the anomaly detection results.
The perceptual monitors enhance the reliability of the decision process for the detection algorithm in an interpretable manner. Moreover, robustness is increased. Besides detecting anomalous behaviour of the network, the monitor is also able to detect unsafe inputs.

\subsubsection{Interpretable Monitoring -- Planning \& Prediction}
Interpretable monitoring of AD planning and prediction systems are primarily concerned with two tasks: accident/collision prediction and safe action selection.
A majority of methods here rely on symbolic representations, predominantly decision trees~\cite{di2020interpretable,gall2021gaussian,gorospeAnalyzingInterVehicleCollision2023,nahata2021assessing,chen2023attention} to predict either a binary or scalar safety score for a fixed set of high-level actions.
These methods are useful to assess the safety of potential actions before they are executed, however, they rely only on the state description without considering other visual cues.

This shortcoming is addressed in other works that rely on raw perception data to assess the safety of driving maneuvers, for example by Karim et al.~\cite{karimExplainableArtificialIntelligence2022} who used GradCAM~\cite{selvarajuGradCAMVisualExplanations2017} to extract visual explanations for accident prediction.
More uniquely, Bao et al.~\cite{baoDRIVEDeepReinforced2021} proposed a two-stage design for traffic accident prediction based on visual attention informed by a Markov decision process designed based on human-like visual attention fixation.
Stage 1 uses saliency maps to show visual attention for both top-down (focus on a particular region) and bottom-up (consider everything) vision, while stage 2 is a stochastic Markov decision process in which an agent predicts the probability of an accident as well as the visual fixation area, such that this setup balances exploration through visual fixation with exploitation for more accurate accident prediction. 

In contrast to predicting the safety of a single action, Schmidt et al.~\cite{schmidt2021can} proposed a decision tree-based monitoring pipeline for full motion planning.
They used imitation learning to train a decision tree based on an RL teacher policy that was trained for safe driving under a constrained MDP. 
The method was shown to be verifiable and easily interpretable, although their evaluation was limited only to lane-change decisions.

Rather than monitoring the safety of motion planning on its own, Gilpin et al.~\cite{gilpin2021explaining} designed a high-level explanatory framework for holistic anomaly detection within the AV that can also create explanations for end-users using natural language.
They proposed a hierarchy of systems to first select explanations generated by lower-level systems (e.g., control and perception modules), and then synthesise higher-level explanations using first-order logic rules and common sense knowledge.
Unfortunately, they did not run a user study, so their explanations' practical usefulness was not assessed.

\subsubsection{Interpretable Monitoring -- Summary}
For the perception task, surrogate models that are interpretable by design can be used to verify the decision of a perception algorithm through their interpretable extraction of semantic concepts. Moreover, the internal workings of a perception algorithm can be monitored via a heat map monitor or an interpretable meta-model can be developed as a monitor to identify anomalies and their causes in the perception algorithm. 

For prediction and planning, most methods rely on decision trees to predict a binary safety label or scalar risk score for the ego vehicle's actions.
These methods only utilise the state description of the environment. 
However, two methods were proposed that consider visual cues from image data as well.

Monitoring methods are well suited to post-hoc address the safety concerns of AD, but they are not designed to calibrate trust.
A huge challenge in interpretable monitoring systems is the trade-off between the computational complexity and the performance of the monitoring algorithm, since the monitor should not take too much time to operate and should not avoid consuming the main system's resources~\cite{Yatbaz2024_Introspection}.

\subsection{Auxiliary Explanations}\label{ssec:cat:aux}

\begin{table*}[]
    \centering
    \caption{Summary of auxiliary explanation methods using the taxonomy of~\cref{ssec:background:XAI-taxonomy}. No user studies were performed in this category. No Control methods were identified in this category. }
    \label{tab:auxiliary-explanations}
    \begin{tabular}{@{}llp{2.75cm}lllllp{3.33cm}@{}}
         \toprule
         {} & \textbf{Paper} & \textbf{Method} & {}    & {} & {}   & {}        & {}      & \textbf{Evaluation}       \\
         {} &             &  \textit{Task}      & \textit{Representation}  & \textit{Stage} & \textit{Mode} & \textit{Scope} & \textit{Medium} & \textit{Method} \\ 
         \midrule
         \multirow{12}{*}{\textsc{Perception}} & \cite{kolekar2022explainable} & semantic segmentation & subsymbolic & post-hoc & importance & local & visual & IDD-lite~\cite{IDD2019_Varma} \\
         {} & \cite{SaravanarjanGradCam} & semantic segmentation &  subsymbolic & post-hoc & importance & local & visual & CamVid~\cite{Brostow2009} \\
         {} & \cite{AbukmeilVariationalAttention} & semantic segmentation &  subsymbolic & post-hoc & importance & local & visual &  SYNTHIA~\cite{ros2016synthia}, A2D2~\cite{geyer2020a2d2} \\
         {} & \cite{mankodiya2022od} & semantic segmentation &  subsymbolic & post-hoc & importance & local & visual & KITTI road ~\cite{Fritsch2013ITSC} \\
         {} & \cite{NowakChargerDetection} & 2D object detection & sybsymbolic & post-hoc & importance & local & visual & self-curated dataset\\
         {} & \cite{Schinagl_2022_CVPR} & 3D object detection & sybsymbolic & post-hoc & importance & local & visual & KITTI~\cite{Geiger2012CVPR}\\
         {} & \cite{BoschVATLD} & traffic light detection & sybsymbolic & post-hoc & inherent & global & visual & BSTLD~\cite{BehrendtNovak2017ICRA} \\
         {} & \cite{ShorrNeuroscope2021} & image classification, \newline semantic segmentation & subymbolic & post-hoc & importance & local & visual & nuScenes \cite{nuscenes2019} \\
         {} & \cite{WangSpatioTemporalVisualAnalytics} & 3D object detection & subsymbolic & post-hoc & importance& cohort & visual & KITTI \cite{Geiger2012CVPR} \\
         {} & \cite{HaedeckeScrutinAI2022} & 2D object detection,\newline semantic segmentation & subsymbolic & post-hoc & importance & cohort & visual & self-curated dataset\\
         \midrule
         \multirow{10}{1.5cm}{\textsc{Planning \& \newline Prediction}} & \cite{jiangIntentionAwareInteractiveTransformer2023} & trajectory prediction & subsymbolic & ante-hoc & importance & local & --- & NGSIM~\cite{NGSIM}, HighD~\cite{highDdataset}, self-curated dataset \\
         {} & \cite{kochakarnExplainableActionPrediction2023} & scene graph learning & subsymbolic &  ante-hoc & importance & local & --- & ROAD~\cite{singh2022road}, \newline Oxford RobotCar~\cite{RobotCarDatasetIJRR} \\
         {} & \cite{liu2023interpretable} & goal recognition, \newline motion planning & subsymbolic & post-hoc & importance & local & --- & Lyft~\cite{houstonOneThousandOne2021} \\
         {} & \cite{wangLaneChangeIntentionPrediction2022a} & lane-change prediction & mixed & post-hoc & importance & local & --- & NGSIM~\cite{NGSIM} \\
         {} & \cite{huTrajectoryPredictionNeural2023} & trajectory prediction & subsymbolic & ante-hoc & importance & local & --- & NGSIM~\cite{NGSIM}, HighD~\cite{highDdataset} \\
         {} & \cite{yuSceneGraphAugmentedDataDriven2022} & risk prediction & subsymbolic & ante-hoc & importance & local & --- & HDD~\cite{ramanishka2018CVPR}, CARLA \\
         {} & \cite{xtrajpred_Zhang2022} & trajectory prediction & subsymbolic & post-hoc & importance & local & visual & Lyft~\cite{houstonOneThousandOne2021} \\
         \midrule

         \multirow{13}{1.5cm}{\textsc{E2E}} & \makecell[lt]{\cite{kim2018textual,kuhn2023textual,Dong2022Transformer,Zhang2023Interrelation,fengNLEDMNaturalLanguageExplanations2023,zhang2023tactical}, \\ \cite{xAI_ObjectInduced_xu}} & action selection & subsymbolic & ante-hoc & importance & local & textual & \makecell[lt]{BDD-X \cite{kim2018textual}, PSI \cite{Chen2021PSIAP}, \\ BDD-OIA \cite{xAI_ObjectInduced_xu}, SAX \cite{gaddSenseAssessEXplain2020g}} \\
         {} & \cite{mori2019visual} & action selection & subsymbolic & ante-hoc & importance & local & visual & GTAV simulator \\
         {} & \cite{Wang2021End2EndInterpretable}   & motion planning & subsymbolic & ante-hoc & inherent &local & visual & NuScenes \cite{nuscenes2019}, CARLA \cite{Dosovitskiy17}, self-curated dataset \\
         {} & \cite{Chen2022End2EndLatent}& control & subsymbolic & ante-hoc & inherent & local & visual & CARLA \cite{Dosovitskiy17} \\
         {} & \cite{Yang2019SceneUnderstanding}, \cite{cultrera2020explaining} & steering & subsymbolic & post-hoc & importance & local & visual & TORCS \cite{Espi2005TORCSTO}, CARLA \cite{Dosovitskiy17} \\
         {} & \cite{AksoyAttentionModel} & braking & subsymbolic & post-hoc & importance & local & visual & \makecell[lt]{BDD-A \cite{BDDA_Xia2017}, \\ CAT2000 \cite{CAT2000_Borji2015}} \\
         {} & \cite{Chitta2021NEAT}  & trajectory planning & subsymbolic & post-hoc & importance & local & visual & CARLA \cite{Dosovitskiy17} \\
         {} & \cite{Sadat2020} & motion planning & subsymbolic & post-hoc & inherent & local & visual & self-curated dataset
         \\
         {} & \cite{wei2021perceive} & motion planning & subsymbolic & ante-hoc & importance & local & visual & nuScenes~\cite{nuscenes2019}, self-curated \\
         {} & \cite{Tashiro2023QuantizedActivation}  & steering & subsymbolic & ante-hoc & importance & local & visual & Udacity \cite{udacity2016} \\
         {} & \cite{teng2022hierarchical} & action selection & sybsymbolic & ante-hoc & inherent & local & visual & CARLA \cite{Dosovitskiy17} \\
         \bottomrule
    \end{tabular}
\end{table*}

\begin{definition}[Auxiliary Explanations]
\textit{We say that an algorithm can provide auxiliary explanations if the execution of the algorithm creates auxiliary information that provides information about how the algorithm produced its output.}
\end{definition}

\subsubsection{Auxiliary Explanations -- Perception}
In perception tasks, heat maps are often created to explain the prediction results by highlighting regions that influence the network's decision.
A widely used model-specific approach is Grad-CAM \cite{selvarajuGradCAMVisualExplanations2017} which visualizes the activation, typically in the last layer. Kolekar et al. \cite{kolekar2022explainable} applied Grad-CAM to a DNN for camera-based semantic segmentation. 
Saravanarajan et al. \cite{SaravanarjanGradCam} also inspected the behaviour of a DNN for semantic segmentation via Grad-CAM under the synthetically generated haze. 
In addition to the last layer, Grad-CAM was also applied to two layers in the encoder and one in the decoder resulting in four different heat maps, thus increasing transparency in the decision understanding of the DNN.

Abukmeil et al. \cite{AbukmeilVariationalAttention} proposed a variational autoencoder for a semantic segmentation task and generated multiple heat maps by computing the second-order derivatives between the encoder layers and the latent space. The resulting attention maps are aggregated and fused with the last decoder layer to improve the results. Mankodiya et al.~\cite{mankodiya2022od} defined a framework to determine the important area of an image contributing to the outcomes of semantic road segmentation, while the XAI methods used here were Grad-CAM and saliency maps.

In \cite{NowakChargerDetection}, Nowak et al. computed attention heat maps for a DNN-based bus charger detection. Additionally, these heat maps are used to identify spurious predictions and are further used for training via data augmentation to increase robustness. Besides providing transparency due to the heat maps, the robustness of the DNN is also increased.   
The aforementioned approaches only focused on camera-based perception tasks. 
Schinagl et al. \cite{Schinagl_2022_CVPR} proposed a model-agnostic attribution map generation method for LiDAR-based 3D object detection. The heat maps are generated perturbation-based via systematically removing LiDAR points and observing the output changes.
They also propose various visual analysis tools which help identify potential misbehaviour of a DNN-based perception system in an interpretable manner.
This way, more transparency in the model working is given and the whole development process of the ML system becomes safer. 

Gou et al. \cite{BoschVATLD} developed the framework Vatld to examine traffic light detection algorithms by analyzing input-output data as well as intermediate representations. Disentangled representation learning was used to extract semantic concepts in the latent representation such as color, background, rotation etc.. Therefore, the analysis tool heavily relies on DNNs that are based on representation learning.

In \cite{ShorrNeuroscope2021}, Schorr et al. developed a toolbox with various state-of-the-art visualisation algorithms of a CNN for image classification and semantic segmentation including Grad-CAM and its extensions, saliency maps \cite{kadir2001saliency} and guided back-propagation \cite{Springenberg2015deconv}.
Wang et al. \cite{WangSpatioTemporalVisualAnalytics} proposed a framework to interpret 3D-object detection failures by combining macro-level spatiotemporal information and micro-level CNN features. 
For the micro-level feature extraction, the heat map algorithm Grad-CAM and the aforementioned Vatld framework were used. 
Haedecke et al. \cite{HaedeckeScrutinAI2022} introduced the analysis toolbox ScrutinAI for semantic segmentation and object detection tasks by offering several visualisation tools. 
Particularly, ScrutinAI may distinguish between metadata in the input (e.g., different observable body parts in an image for pedestrian detection) to explicitly identify model weaknesses related to semantic concepts of objects.

\subsubsection{Auxiliary Explanations -- Planning \& Prediction}
The overwhelming majority of methods generating auxiliary explanations rely on the attention mechanism to gain some insight into how the algorithms reached their output.
The recurring design pattern here is that a recurrent neural network (RNN) is proposed onto which an auxiliary attention mechanism is bolted.
Alternatively, instead of an RNN, a transformer architecture is proposed, in which case the attention mechanism is built into the neural network from the start.
The explainability analyses of these methods are then performed by looking at the attention scores that the model assigns to either the input or some interpretable input embedding.
Finally, the attention scores are sometimes visualised using heat maps or bar graphs.

For example, Jiang et al.~\cite{jiangIntentionAwareInteractiveTransformer2023} proposed a transformer-based method for inter-vehicle trajectory interaction analysis.
Their evaluation showed that the proposed model is significantly faster than similar methods and performs competitively as compared to baselines, with the added benefit of some interpretability analysis.
In addition, Kochakarn et al.~\cite{kochakarnExplainableActionPrediction2023} designed an algorithm with spatial and temporal attention for road scene understanding.
A self-supervised scene-graph learning algorithm is used to create spatiotemporal embeddings of scene graphs based on graph contrastive learning, which is then used for driver action prediction as a downstream task.
As the final stage of graph embedding, an attention layer is used to highlight the most important spatial and temporal factors in the scene graph sequence as a form of post-hoc explainability.
Yu et al.~\cite{yuSceneGraphAugmentedDataDriven2022} also used an attention mechanism with scene graph embeddings as well as image data to predict binary risk prediction. 
However, no quantitative evaluation is given and only one qualitatively interesting example is presented of the impact of attention mechanisms on explainability on safety prediction. 
Finally, explainable trajectory prediction has also received some attention along the similar neural-attention methodology~\cite{huTrajectoryPredictionNeural2023,xtrajpred_Zhang2022}. 

However, due to the unreliability of attention-based explanations, it is interesting to look at methods which do not rely on attention weights.
Liu et al. \cite{liu2023interpretable} used a post-hoc heatmap to infer different potential goals on a map, which then guides a neural network-based planner to capture planning uncertainties. 
Additionally, Wang et al.~\cite{wangLaneChangeIntentionPrediction2022a} combined bi-directional long short-term memory with a conditional random field (CRF) predictor to provide scores for interpretable hand-crafted features in LC scenarios.
Their model also enforces interpretable hard and soft rules that the system must satisfy.
However, their evaluation is limited and no qualitative discussion is given of how the CRF improves the interpretability of the system as a whole.

\subsubsection{Auxiliary Explanations -- End-to-end}
Kim et al. \cite{kim2018textual} proposed the generation of textual explanations for E2E driving tasks. They introduced a dataset called BDD-X (Berkeley DeepDrive eXplanation) with driving videos annotated with driving descriptions and action explanations. In addition to the E2E control system, a second attention-based model was trained to predict textual explanations from video sequences. The attention maps of both models were aligned to create a dependency between the controller and the explanations. Based on that, Kühn et al. \cite{kuhn2023textual} evaluated the developed baseline on a new dataset called SAX~\cite{gaddSenseAssessEXplain2020g} and proposed some improvements over the baseline.
They utilized video frames as input and generated natural language action descriptions and explanations using an opaque neural network. 
Building on this architecture, Mori et al.~\cite{mori2019visual} incorporated throttle into the control in addition to steering and developed an attention map for visual explanations of AV decisions. 
Xu et al. \cite{xAI_ObjectInduced_xu} introduced the dataset BDD-OIA (object-induced actions) which extracted complicated scenarios from BDD-X and annotated them with new explanations focusing on objects which influence the decision. 
Additionally, they proposed a DNN architecture which jointly learns action prediction and textual generation.  
Dong et al.~\cite{Dong2022Transformer} extended the approach by introducing a transformer architecture for the E2E network. 
In this way, the decision and reason generator could include the feature extractor and the attention zones of the transformer architecture. 
For the decision and reason generator task, Zhang et al. \cite{Zhang2023Interrelation} introduced an additional interrelation module in the network expressing interrelationships among the ego vehicle and other traffic-related objects. 
This module is then combined with global features of the E2E network to provide more reliable actions and explanations. 

Feng et al. \cite{fengNLEDMNaturalLanguageExplanations2023} proposed to expand the textual reasoning about the driving actions with explanations including the surrounding environment based on semantic segmentation by extending the BDD-OIA dataset with additional annotations, although they did not qualitatively show the added benefit of the new annotations.
In \cite{zhang2023tactical}, Zhang et al. extended the BDD dataset by BDD-3AA by providing explanations and corresponding object segmentations. 
The interpretation was provided by importance value scores for the objects on the image. 
Human evaluation showed that object-level explanations are more persuasive than pixel-level explanations while the additional textual explanations increased trust for users and manufacturers.
However, the decisions and explanations do not necessarily correlate, and the explanations need to be validated for reliability. 

Wang et al.~\cite{Wang2021End2EndInterpretable} proposed intermediate outputs in the E2E design to improve interoperability. 
Besides the planned trajectory as an output, they also provide future semantic maps from the intermediate perception part in Birds-Eye-View (BEV). 
A similar approach was proposed by Chen et al. \cite{Chen2022End2EndLatent} where a semantic BEV mask containing a map, ego state, surrounding objects and routing was delivered. 
Yang et al. \cite{Yang2019SceneUnderstanding} proposed two frameworks generating attention maps of E2E controllers to better understand scenes.
The first one was model-specific and produced feature maps from the convolutional layer. In contrast, a second model-agnostic approach was proposed which compared the controller outputs between the raw input images and occluded ones. By examining changes in the output, a pixel-wise heat map was created. 

Cultrera et al. \cite{cultrera2020explaining} proposed attention blocks in the DNN-based E2E controller to 
create attention maps. 
Aksoy and Yazici~\cite{AksoyAttentionModel} developed an E2E controller which explicitly provided a saliency map prediction as an intermediate output and as an input for the action prediction. Chitta et al. \cite{Chitta2021NEAT} proposed an E2E system which provides a trajectory and a BEV semantic prediction as an output. Moreover, attention maps of the DNN are computed to increase interpretability. Similarly, Sadat et al. \cite{Sadat2020} introduced an E2E motion planner that provides semantic occupancy forecasting as an interpretable intermediate representation resulting from the perception and prediction task. The intermediate output consists of a semantic occupancy map including, motion predictions of different agents. Moreover, the cost function of the motion planner takes occupancy forecasting as an additional input to increase safety of the generated trajectories. 

Wei et al. \cite{wei2021perceive} trained an E2E method that directly plans the future trajectory for the ego vehicle. Their method includes an attention mask over a CNN backbone that they claim can increase the safety and interpretability of the system by allowing the inspection of the LiDAR input data. However, their evaluation does not analyse the benefits of this system. 
Tashiro et al. \cite{Tashiro2023QuantizedActivation} also produced heat maps as an intermediate output for an E2E controller. 
For the heat map generation, they quantised the network activations to pay limited attention to specific bits and showed improved performance to other attention map generation methods. 
In addition, the visual intermediate outputs lead to a similar transparency that modular AD architectures can provide. 
This could also help identify errors in complex E2E systems more accurately. 
However, the reliability of the intermediate output is not guaranteed and the intermediate explanations do not necessarily help in understanding the behaviour of the E2E system.

Teng et al. \cite{teng2022hierarchical} leveraged a Bird's Eye View (BEV) mask, which provided scene semantic information. 
They argued that the BEV mask can demonstrate how an AV understands the scenarios and thus promote interoperability. 

\subsubsection{Auxiliary Explanations -- Summary}
A prominent approach in perception is generating heat maps, which visually explain the regions in the input that the black box algorithm has focused on. 
The heat maps can be utilized for local post-hoc explanations. However, heat maps can be fragile and unreliable. Moreover, it is difficult to evaluate the correctness of the provided explanation \cite{molnarInterpretableMachineLearning2023}. 
Planning and prediction algorithms that generate auxiliary explanations all rely on an attention mechanism either as part of the transformer architecture or in conjunction with a recurrent neural network (RNN).
Attention weights are then manually interpreted giving some insight into how the algorithm transformed the input into a decision.

The benefit of attention-based methods is that their results are interpretable through the analysis of the attention weights, but the proposed systems are highly model-specific and require detailed knowledge of the underlying architecture of the neural network.
In addition, attention weights are widely known to be inconsistent and difficult to interpret as explanations~\cite{jain2019attention,wiegreffeAttentionNotNot2019b}.
The major problem here is that attention weights may provide a ``plausible'' -- i.e., intuitively correct -- explanation of the decision-making algorithm despite not being ``faithful'', i.e., factually correct.
This may then wrongly calibrate people's trust in the system leading to over- or under-reliance.

For E2E learning, there are various possibilities to provide auxiliary explanations. As in perception, heat maps can be provided which highlight important regions for the algorithm's action in the input. These visual explanations can be expanded by textual explanations, that generate reasons based on the input for the chosen action. Lastly, one can also provide intermediate outputs which visualise the perception or prediction part inside the E2E network. This gives a better insight into the internal workings of the E2E network, but the intermediate outputs cannot directly explain the network's decisions. 

\subsection{Interpretable Safety Validation}\label{ssec:cat:valid}

\begin{definition}[Interpretable Safety Validation]
\textit{We say that an algorithm} provides \textit{interpretable safety validation if it uses an interpretable algorithm to generate adversarial behaviours of other traffic participants for the validation of an AV.}
\end{definition}

\subsubsection{Interpretable Safety Validation -- Summary}
These interpretable methods focus on supporting safety assurance via post-hoc explainability by either generating failure cases or by extracting accident scenarios to be used in safety assessments.
Safety validation also differs from the standard perception, planning, and other runtime functions as it is executed during the verification and validation phase (offline).
For reinforcement learning, temporal logic can be inserted into policies to ensure safe behaviour.
For critical scenarios in prediction and perception, heat maps can be further analyzed to extract and interpret critical factors in the corresponding scenarios.

Corso and Kochenderfer \cite{corso2020interpretable} utilized signal temporal logic (STL) to generate high-likelihood failures for AVs, while they argued STL is easily understood because of its logical description between temporal variables.
DeCastro et al.~\cite{decastro2020interpretable} leveraged parametric STL (pSTL) to construct an interpretable view on modeling a relationship between policy parameters to the emergent behaviours from deploying that policy, while the behaviour outcome is expressed by pSTL formulas. 
As pSTL provides a way to construct formulas that describe the relationships between spatial and temporal properties of a signal, the formally-specifiable outcome can be obtained by configuring the parameters, allowing proactively generating various desired behaviour of an agent for testing AVs.

Kang et al. \cite{kang2022vision} proposed a visual transformer to predict collisions supplemented by attention maps. 
Subsequently, a time series of attention maps is further analysed to identify spatiotemporal characteristics and based on the situation interpretation, accident scenarios for safety assessment are extracted. The extraction is based on the definition of functional scenarios by the PEGASUS project~\cite{MenzelPegasusScenarios2019} on 6-layer information including road levels, traffic infrastructure, events, objects, environments, and digital information.

In~\cite{LIDLHybridFramwework2020}, Li et al. introduced a risk assessment phase for the perception and prediction of dangerous vehicles as well as traffic lights. 
A visual explanation for the classification is provided by computing saliency maps via RISE algorithm \cite{petsiuk2018rise}, which supports safety assurance in the risk assessment phase. 
Shao et al. \cite{shao2023safety} also output the intermediate interpretable features for semantic explanation, aiming to enhance safety for the downstream controller.

\section{XAI Framework for AD}\label{sec:framework}
We now provide an overview of existing XAI frameworks for AD and analyze their limitations. 
As part of our systematic review, we identified three relevant XAI frameworks, which illustrate high-level AD modules and describe various ways to integrate them.
Subsequently, we propose our XAI framework for AD -- \textit{SafeX}: a framework for safe and explainable AD --  based on the concrete XAI methods summarized in~\cref{sec:cat}.

\subsection{Existing XAI frameworks}

\begin{figure}
    \centering
    \includegraphics[width=\linewidth]{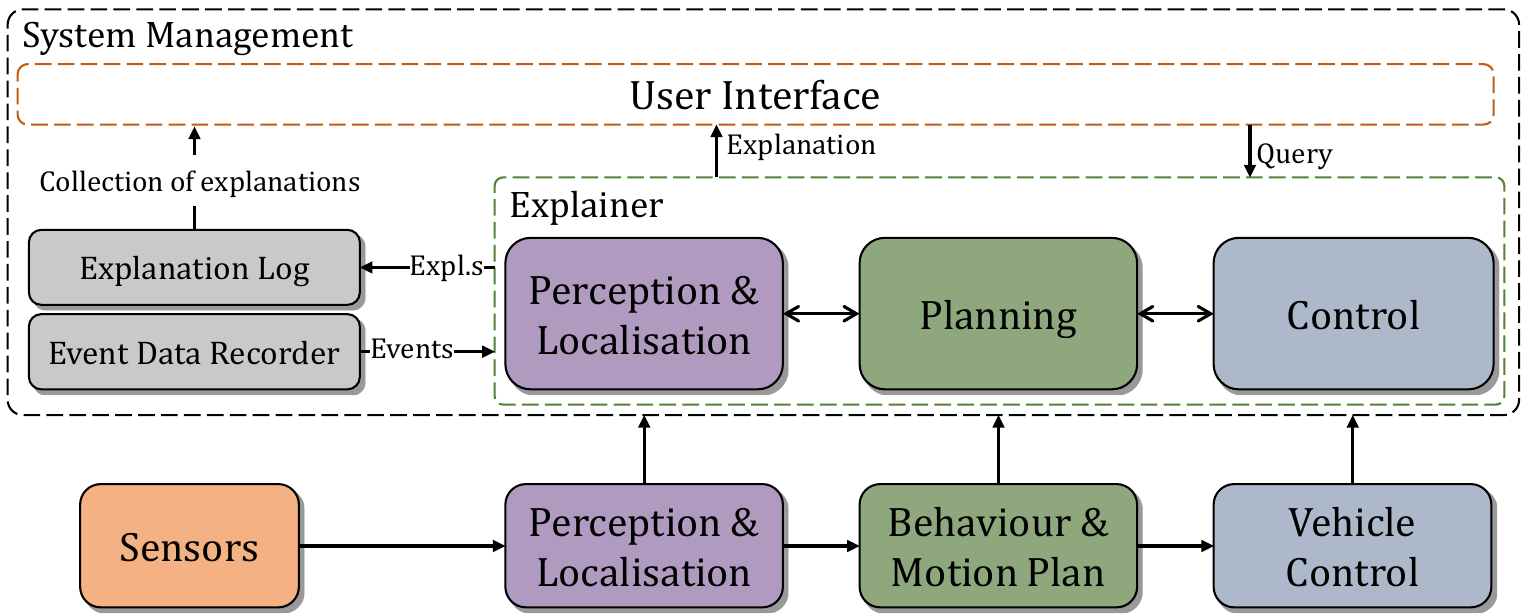}
    \caption{Reproduction of the framework from \cite{omeiza2021explanations} with three main components: the user interface, the explainer, and the AD modules. The explainer serves as a middleware between users and AD modules and interacts with them.}
    \label{fig:XAIgeneralforAV}
\end{figure}

Omeiza et al.~\cite{omeiza2021explanations} defined an explainer as the bridge between an AV and users, allowing explanations to users' queries based on the information from AD modules, as shown in \cref{fig:XAIgeneralforAV}. 
Instead of focusing on a specific AV function, their framework remains at a high level to illustrate the general role of XAI in AD. 
Atakishiyev et al.~\cite{atakishiyev2021explainable} introduced a similar conceptual framework for E2E autonomous control systems by including XAI components that realise safety-regulatory compliance.
In this framework, an XAI component aims to provide explanations of each driving action taken in the given environment. 
Regulatory compliance is confirmed by simulation and real-world testing based on these explanations. 

The framework defined by Brajovic et al.~\cite{brajovic2023model} consists of four steps for the entire development cycle of AI. 
These are use case definition, data collection, model development, and model operation. 
The use case describes the task that the AI aims to solve, while the data affects whether the AI is biased and robust. 
The developed model is aimed to achieve an appropriate level of accuracy, robustness, explainability, and other desirable requirements.
Finally, the model operation shall be equipped with a monitoring system that is proportionate to the nature of the AI and its associated risks. 
Although this framework provides useful guidance, its application to AVs is not addressed and users' queries are not considered.

\begin{figure*}[ht]
    \centering
        \subfloat[]{
        \includegraphics[width=0.55\linewidth]{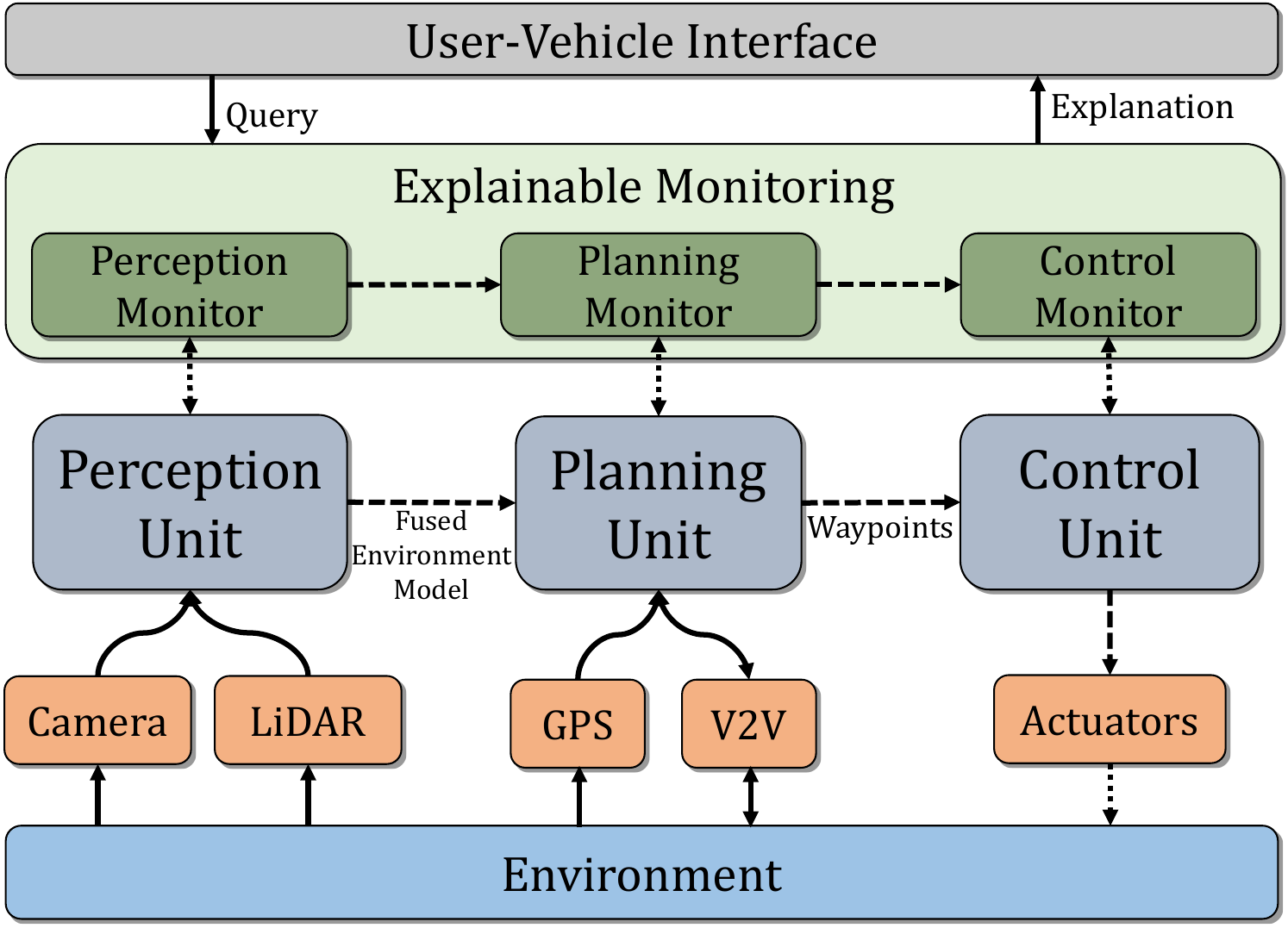}
        \label{fig:ourframeworksub1}
    }
    \hspace{1.5em}%
       \subfloat[]{
        \includegraphics[width=0.26\linewidth]{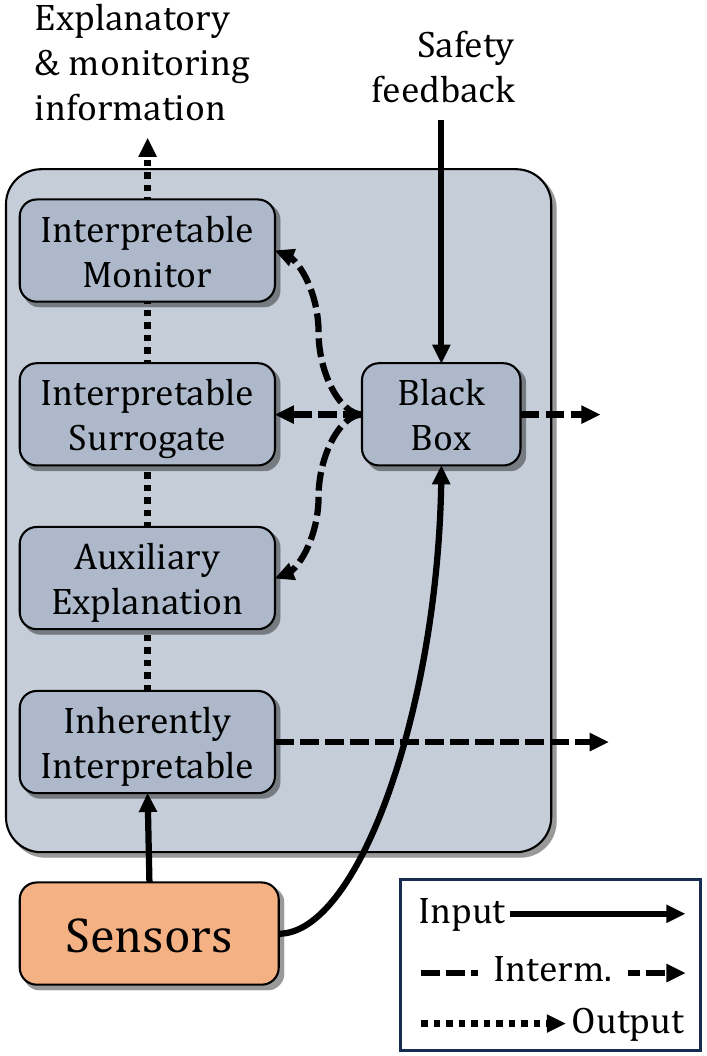}
        \label{fig:ourframeworksub2}
    }
    \caption{\textit{SafeX}: our framework for safe and explainable AI for AD integrates concrete XAI techniques with AD in an actionable way. \textbf{(a)} Explainable monitoring generates intelligible explanations to users' queries using information from AD modules to achieve trustworthiness while monitoring the output of each module by providing safety feedback to ensure safety; \textbf{(b)} Each unit in SafeX may contain interpretable surrogate monitors and models, while auxiliary explanations can be applied for the black boxes in each AD module. Alternatively, the functions in AD modules are designed to be inherently interpretable.}
    \label{fig:ourframework}
\end{figure*}

\subsection{Conceptual Framework: SafeX}
We propose a novel conceptual framework for safe and explainable AD shown in \cref{fig:ourframework}, which we call SafeX.
Different from the frameworks proposed in previous work, we present a more fine-grained application of XAI to AD, focusing on the integration of the concrete surveyed methods within the full AD stack in a way that also enables safety monitoring and intelligible explanation delivery.

The overall structure of SafeX is shown in \cref{fig:ourframeworksub1}.
We define an explainable monitoring system (EMS) as a bridge between users and an AV. 
On one hand, the EMS generates intelligible explanations to users based on their queries by extracting the necessary information from the AV. 
On the other hand, it includes a monitor for each AD module to deliver safety feedback regarding the module's output. 
These two functions of the EMS are not only aimed at increasing a user's understanding and trust in the AV but also at providing a safer AV for the user.
To accomplish the two roles of the EMS, each AD module must be carefully designed. 
\Cref{fig:ourframeworksub2} uses the four identified XAI categories from our survey to deliver explanatory and monitoring information to the upstream EMS for each AD module. 
For black boxes in an AD module, interpretable monitors, interpretable surrogate models, and auxiliary explanations can be applied. 
In addition, the functions in the module can also be inherently interpretable to deliver traceable explanatory information if the interpretable functions meet the performance requirements.
They may also serve as a fallback if the monitoring systems report unexpected and unverifiable behaviour from the black box systems.

In contrast to existing frameworks, SafeX is based on concrete state-of-the-art methods, and we design SafeX according to the modular components identified in~\cref{sec:cat}. 
We display two variants of how SafeX could be realized.

\textit{Variant 1}: all three units in \cref{fig:ourframeworksub1} are deep learning-based black box modules. For camera-based perception, heat maps can be generated as an auxiliary explanation \cite{WangSpatioTemporalVisualAnalytics}, highlighting pixels in the camera image that are relevant to the black box model's prediction. Moreover, a random forest can be applied to identify perception errors based on meta-information from the environment \cite{ponn2020identification}. With the corresponding Shapley values, an interpretable surrogate model can be applied. Additionally, the concept-bottleneck model for pedestrian detection can be used for the EMS as proposed in \cite{keser2023interpretable}. The perception monitor can verify the predictions of the object detector for the safety-relevant object class pedestrians. Regarding deep learning-based motion planning algorithms, they take the environmental representation provided by the object detector as an input and provide a planned trajectory of the AV as an output. Similarly to the perception unit, a heat map highlighting different potential goals on the map \cite{liu2023interpretable} can be generated as an auxiliary explanation. As a surrogate model, the cognitive model introduced by Gyevnar et. al. \cite{gyevnar2024cema} based on the planning and prediction model~\cite{albrechtInterpretableGoalbasedPrediction2021} can be utilized to provide causal explanations for decision-making. For the EMS, an inherently interpretable decision tree can be trained to verify the decisions of the black box motion planner, as proposed in \cite{schmidt2021can}. Lastly, a DNN-based controller can be applied, which can be combined with cross-comparable clustering as a surrogate to aid in interpreting the control signals. 

\textit{Variant 2}: AD modules are inherently interpretable. The perception module could be trained as a concept-bottleneck model \cite{Losch2021SB}. Extracting semantic concepts in the segmented environment makes the algorithm more interpretable and reliable. For the subsequent planning unit, a Monte Carlo Tree Search over high-level driving maneuvres can be applied \cite{albrechtInterpretableGoalbasedPrediction2021}. The resulting search tree helps in interpreting the planner's decision. Lastly, a neural network-based controller with an interpretable projection mechanism \cite{zheng2023towards} can be used as an explainable control unit in the framework.

The resulting modular design allows future research and development to focus on deeply investigating and refining specific components independently.
By stacking multiple forms of XAI methods, we can enable developers to integrate the appropriate methods with their AD stack based on specific stakeholder and regulatory requirements, and the desired degree of safety.
Moreover, the proposed EMS can simultaneously achieve both the safety monitoring of the AV and the delivery of intelligible explanations to users' queries.

\section{Discussion}\label{sec:discussion}
We set out to answer two research questions based on a systematic literature review.
In closely scrutinising the retrieved publications for \textbf{RQ1}, we found that state-of-the-art literature is trying to resolve the challenge of safe and trustworthy AI in AD by focusing on five core XAI design paradigms, namely interpretable design, interpretable surrogate models, interpretable monitoring, auxiliary explanations, and interpretable validation. 

It is interesting to note that there is a significant imbalance in the number of publications among each of the driving tasks of perception, planning and prediction, and control.
Control is consistently more neglected across all five XAI design paradigms than perception and planning, despite the intensive research into neural network-based safe RL control methods~\cite{waschleReviewAISafety2022}. 
Furthermore, XAI for LiDAR-based perception and various fusion approaches remains highly unexplored compared to camera-based detectors. 
This is noteworthy even though the majority of state-of-the-art perception architectures incorporate LiDAR sensors due to their provision of accurate depth information~\cite{feng2021review}. 
In contrast, E2E methods enjoy significant attention from the field, however, most methods for these systems are constrained to auxiliary explanations. 

However, herein lies an important challenge.
It has been shown many times, that the post-hoc analysis methods of auxiliary explanations based on Shapley values, attention maps, or saliency maps are neither consistent nor necessarily correct (see for some examples \cite{jain2019attention,kumarProblemsShapleyvaluebasedExplanations2020a,wangDataBanzhafRobust2023}).
While these methods are undoubtedly useful for building explanations, they are also not sufficient, if our requirements of trustworthy and safe AI are to be upheld in, for example, regulations and courts.
This challenge is then further exacerbated by the fact that the evaluation of auxiliary methods is usually cursory with hard-to-interpret quantitative metrics and no qualitative insights at all.
One way to increase safety for AD is to integrate multiple XAI techniques into one framework in a ``Swiss cheese'' model of safety that assures that malfunctions do not go unnoticed through the AD stack.

This is why our analysis of \textbf{RQ2} is relevant, and why we propose a new framework called SafeX to integrate concrete XAI methods with AD.
We found that the number of existing works about frameworks or pipelines is limited and these provide only a very high-level overview of the ways XAI may be integrated with AD on a lower level of the AD stack.
Given these limitations and the urgent need for safe and trustworthy AI for the AD stack, our framework SafeX modularly integrates the identified techniques of XAI with each AD module.
A modular approach in SafeX allows the combination of multiple sources of explanations in a way that may reduce the risks of using AI for AD.
One may also combine multiple modalities of predictions which, when used with our proposed explainable monitoring system, can act both as a bridge between users and the AD system and as a tool for comprehensive safety guarantees.
The EMS is, thus, designed to enable the delivery of explanations to users while ensuring the safety of each AD module through runtime monitoring. 

We also observed that interpretable safety validation, one of the five XAI design paradigms, has received less attention in the field. 
This is relevant because the safety testing of AVs is one of the most pertinent and difficult challenges that currently faces the AD field due to the heavy-tailed distribution of driving scenarios \cite{wang2023application}.
As we saw in~\cref{ssec:cat:valid}, one way to mitigate this problem is the extraction of varied scenarios from real driving data that is achieved through an interpretability analysis uncovering the relevant factors of the environment in the scenario.
Through interpretability, we can also understand the causal factors in the scenarios so that we can manipulate them and extract new scenarios.

In our study, we narrowed our focus on perception, planning and prediction, and control, while not considering studies about data diversity, ethics, or AI model oversight. 
This is because the former three are arguably the most pressing if we aim to address the requirements of safe and trustworthy AI in a way that also translates to more deployable and reliable AVs.
While the latter three are undoubtedly important, their solution may present less of a stride towards creating real-world AVs.

In addition, natural language understanding and generation for interacting with users and delivering intelligible explanations were not considered, though our review has picked up on a few methods~\cite{omeiza2021explanations,gyevnar2024cema,kridalukmanaSelfExplainingAbilitiesIntelligent2022} that directly consider human-robot interactions as a significant part of the explanatory process.
What this suggests, is that there exists a disconnect between research that focuses on the needs of end users and research that addresses explainability of the driving stack.
The problem with this gap is that explanations ought to change depending on the requirements of the user and the design of explanations need to take this dependency into account otherwise risking invoking mistrust or confusion in users. This necessitates the study of evaluating methods for explainability with humans as actual stakeholders, for example, as summarized by Vilone and Longo~\cite{vilone2020explainable}.

Furthermore, our focus on XAI is only a partial measure of how safe and trustworthy AI should be achieved.
As discussed in \cref{ssec:background:trustworthyAI}, trustworthiness, safety, and transparency are overarching concepts that require the cross-disciplinary collaboration of people. 
Other measures such as uncertainty quantification, rigorous testing, thorough documentation, standardisation, etc. are also necessary.
Still, we have also seen that XAI is a diverse and popular field that addresses some of the key requirements of trustworthy and safe AI.

Finally, it is worth noting that generative methods did not appear prominently in our review, even though our systematic search did not exclude these papers. On one hand, this is not surprising because generative methods further exacerbate the issues around black-box decision-making algorithms such that existing XAI algorithms cannot be applied to them. On the other hand, generative methods, especially multi-modal methods which combine sensing, control, and language (e.g., \cite{marcu2024lingoqavideoquestionanswering}), have the potential to self-explain their decisions. In addition, the issue of the long-tail disttribution of critical scenarios may also be aleviated by generative methods, though their efficacy at this is yet to be verified. At the very least, generative modelling is an area that requires further investigation and may be a promising direction for future research.

To summarise, we identify the following challenges and recommendations for the field of XAI for AD:
\begin{itemize}
    \item \textit{Explainable perception architecture}: investigate more explainable approaches for other sensors such as LiDAR and Radar not just camera-based perception; explore XAI for various fusion architectures, particularly combining XAI methods for different sensors that are integrated; 
    \item \textit{Rigorous testing for auxiliary methods}: auxiliary explanations methods like Shapley values, saliency and attention maps are prone to gaming, inaccuracies, and misinterpretation. It is necessary to thoroughly evaluate these methods not just quantitatively but also with extensive qualitative insights that focus especially on the failure cases of the methods;
    \item \textit{Modular and layered monitoring}: to improve the safety of AD, one method does not suffice. Our proposed framework, SafeX, instead suggests that multiple layers of independent and co-supervisory explanatory functions should verify and monitor the workings of underlying black box systems and each other, potentially providing fallback options in emergencies;
    \item \textit{Cross-disciplinary collaboration}: XAI methods are usually developed in isolation. To better understand stakeholder requirements and to adapt explanations to the varied socio-technical interactions of the real world, it is crucial to develop methods that are rooted in actual problems and not merely motivated by a vague sense of need for safety and trustworthiness.
    \item \textit{Generative methods}: generative methods are a promising direction for future research in XAI for AD. They have the potential to self-explain their decisions and may be able to alleviate the issue of the long-tail distribution of critical scenarios. However, their efficacy at this is yet to be verified.
\end{itemize}

\section{Conclusion}\label{sec:conclusion}
In this paper, we investigated the applications of XAI for safe and trustworthy AD. 
We began the survey by defining requirements for trustworthy AI in AD, noting that XAI is a promising field for addressing several of these requirements. 
Subsequently, we gave an overview of the sources of explanations in AI and presented the taxonomy of XAI.
Based on a systematic literature survey founded on two research questions, we derived five key applications of XAI for safe and trustworthy AI in AD and an appropriate framework to integrate these applications into AD.
Our key findings are:

\begin{itemize}
    \item Actual XAI for AD research can be sorted into five categories: interpretable design, interpretable surrogate models, interpretable monitoring, auxiliary explanations, and interpretable validation;
    \item There is a lack of detailed general XAI for AD frameworks that address safety requirements and are also rooted in concrete research. We propose to fill this gap with a new framework SafeX that can incorporate all categories of XAI methods designed for AD;
    \item XAI for AD, as an emerging topic, is gaining increasing attention according to the published literature per year. We expect that the number of studies will further increase with the development of AI.
\end{itemize} 

Looking to the future, we can expect legal and social pressures to increase on the development of AD.
Growing up to this challenge will require joined initiatives from multiple disciplines and the involvement of various stakeholders.
Here, we expect XAI to act as a bridge that could connect cross-disciplinary gaps.
Emerging fields will also continue to influence the field. 
With the advent of large language model-based (LLM) systems, there will be a pronounced need for XAI more than ever, as models continue to improve and emergent behaviour is discovered every day.
Calls for this in other fields are already emerging (e.g., mechanistic interpretability~\cite{elhage2021mathematical}), however, the use of LLMs in AD further complicates the black box problem.
In addition, LLMs themselves could one day become the explainers, but it will only be through the involvement of various stakeholders and disciplines that this may become a reality for safe and trustworthy AD.

\bibliographystyle{IEEEtran}
\bibliography{bibs/taxonomy, bibs/introduction, bibs/literature, bibs/requirements}

\begin{thebibliography}{100}
\providecommand{\url}[1]{#1}
\csname url@samestyle\endcsname
\providecommand{\newblock}{\relax}
\providecommand{\bibinfo}[2]{#2}
\providecommand{\BIBentrySTDinterwordspacing}{\spaceskip=0pt\relax}
\providecommand{\BIBentryALTinterwordstretchfactor}{4}
\providecommand{\BIBentryALTinterwordspacing}{\spaceskip=\fontdimen2\font plus
\BIBentryALTinterwordstretchfactor\fontdimen3\font minus
  \fontdimen4\font\relax}
\providecommand{\BIBforeignlanguage}[2]{{%
\expandafter\ifx\csname l@#1\endcsname\relax
\typeout{** WARNING: IEEEtran.bst: No hyphenation pattern has been}%
\typeout{** loaded for the language `#1'. Using the pattern for}%
\typeout{** the default language instead.}%
\else
\language=\csname l@#1\endcsname
\fi
#2}}
\providecommand{\BIBdecl}{\relax}
\BIBdecl

\bibitem{Mathew2020DL}
A.~Mathew, P.~Amudha, and S.~Sivakumari, ``Deep learning techniques: an
  overview,'' \emph{Advanced Machine Learning Technologies and Applications:
  Proceedings of AMLTA 2020}, pp. 599--608, 2021.

\bibitem{Jammal2023HumanMachine}
A.~A. Jammal, A.~C. Thompson, E.~B. Mariottoni, S.~I. Berchuck, C.~N. Urata,
  T.~Estrela, S.~M. Wakil, V.~P. Costa, and F.~A. Medeiros, ``Human versus
  machine: comparing a deep learning algorithm to human gradings for detecting
  glaucoma on fundus photographs,'' \emph{American journal of ophthalmology},
  vol. 211, pp. 123--131, 2020.

\bibitem{willers2020safety}
O.~Willers, S.~Sudholt, S.~Raafatnia, and S.~Abrecht, ``Safety concerns and
  mitigation approaches regarding the use of deep learning in safety-critical
  perception tasks,'' in \emph{International Conference on Computer Safety,
  Reliability, and Security}, 2020, pp. 336--350.

\bibitem{iso_iso_2018}
\BIBentryALTinterwordspacing
{ISO}, ``{ISO} 26262-1:2018(en), {Road} vehicles — {Functional} safety,''
  2018. [Online]. Available: \url{https://www.iso.org/standard/43464.html}
\BIBentrySTDinterwordspacing

\bibitem{salay2018using}
R.~Salay and K.~Czarnecki, ``Using machine learning safely in automotive
  software: An assessment and adaption of software process requirements in iso
  26262,'' \emph{arXiv preprint arXiv:1808.01614}, 2018.

\bibitem{iso_iso_2022}
\BIBentryALTinterwordspacing
ISO, ``{ISO} 21448:2022: {Road} vehicles—{Safety} of the intended
  functionality,'' 2022. [Online]. Available:
  \url{https://www.iso.org/standard/77490.html}
\BIBentrySTDinterwordspacing

\bibitem{burton2022safety}
S.~Burton, C.~Hellert, F.~H{\"u}ger, M.~Mock, and A.~Rohatschek, ``Safety
  assurance of machine learning for perception functions,'' in \emph{Deep
  Neural Networks and Data for Automated Driving: Robustness, Uncertainty
  Quantification, and Insights Towards Safety}.\hskip 1em plus 0.5em minus
  0.4em\relax Springer International Publishing Cham, 2022, pp. 335--358.

\bibitem{nuissl}
D.~Gesmann-Nuissl and I.~Tacke, ``Funktionale sicherheit ki-basierter systeme
  im automobilsektor,'' in \emph{the 14th Workshop Fahrerassistenz und
  automatisiertes Fahren}, 2022, pp. 85--98.

\bibitem{iso8800}
\BIBentryALTinterwordspacing
{ISO}, ``{ISO/CD PAS} 8800:road vehicles - safety and artificial
  intelligence,'' 2023. [Online]. Available:
  \url{https://www.iso.org/standard/83303.html}
\BIBentrySTDinterwordspacing

\bibitem{AAA}
\BIBentryALTinterwordspacing
B.~Moye. (2023) Aaa: Fear of self-driving cars on the rise. [Online].
  Available:
  \url{https://newsroom.aaa.com/2023/03/aaa-fear-of-self-driving-cars-on-the-rise/}
\BIBentrySTDinterwordspacing

\bibitem{Reig2018studyAIAD}
S.~Reig, S.~Norman, C.~G. Morales, S.~Das, A.~Steinfeld, and J.~Forlizzi, ``A
  field study of pedestrians and autonomous vehicles,'' in \emph{Proceedings of
  the 10th international conference on automotive user interfaces and
  interactive vehicular applications}, 2018, pp. 198--209.

\bibitem{langer2021we}
M.~Langer, D.~Oster, T.~Speith, H.~Hermanns, L.~K{\"a}stner, E.~Schmidt,
  A.~Sesing, and K.~Baum, ``What do we want from explainable artificial
  intelligence (xai)?--a stakeholder perspective on xai and a conceptual model
  guiding interdisciplinary xai research,'' \emph{Artificial Intelligence},
  vol. 296, p. 103473, 2021.

\bibitem{Dwivedi2023Debug}
R.~Dwivedi, D.~Dave, H.~Naik, S.~Singhal, R.~Omer, P.~Patel, B.~Qian, Z.~Wen,
  T.~Shah, G.~Morgan, and R.~Ranjan, ``Explainable ai (xai): Core ideas,
  techniques, and solutions,'' \emph{ACM Computing Surveys}, vol.~55, no.~9,
  pp. 1--33, 2023.

\bibitem{Weitz2019TrustXAI}
K.~Weitz, D.~Schiller, R.~Schlagowski, T.~Huber, and E.~Andr{\'e}, ``" do you
  trust me?" increasing user-trust by integrating virtual agents in explainable
  ai interaction design,'' in \emph{Proceedings of the 19th ACM International
  Conference on Intelligent Virtual Agents}, 2019, pp. 7--9.

\bibitem{ColumbiaLaw2019xAI}
A.~Deeks, ``The judicial demand for explainable artificial intelligence,''
  \emph{Columbia Law Review}, vol. 119, no.~7, pp. 1829--1850, 2019.

\bibitem{Muhammad2021SafeAD}
K.~Muhammad, A.~Ullah, J.~Lloret, J.~D. Ser, and V.~H.~C. de~Albuquerque,
  ``Deep learning for safe autonomous driving: Current challenges and future
  directions,'' \emph{IEEE Transactions on Intelligent Transportation Systems},
  vol.~22, no.~7, pp. 4316--4336, 2021.

\bibitem{omeiza2021explanations}
D.~Omeiza, H.~Webb, M.~Jirotka, and L.~Kunze, ``Explanations in autonomous
  driving: A survey,'' \emph{IEEE Transactions on Intelligent Transportation
  Systems}, vol.~23, no.~8, pp. 10\,142--10\,162, 2021.

\bibitem{atakishiyev2021explainable}
S.~Atakishiyev, M.~Salameh, H.~Yao, and R.~Goebel, ``Explainable artificial
  intelligence for autonomous driving: A comprehensive overview and field guide
  for future research directions,'' \emph{arXiv preprint arXiv:2112.11561},
  2021.

\bibitem{zablocki2022explainability}
{\'E}.~Zablocki, H.~Ben-Younes, P.~P{\'e}rez, and M.~Cord, ``Explainability of
  deep vision-based autonomous driving systems: Review and challenges,''
  \emph{International Journal of Computer Vision}, vol. 130, no.~10, pp.
  2425--2452, 2022.

\bibitem{buchananFundamentalsExpertSystems1988}
B.~G. Buchanan and R.~G. Smith, ``Fundamentals of expert systems,'' in
  \emph{Annual Review of Computer Science: Vol. 3, 1988}.\hskip 1em plus 0.5em
  minus 0.4em\relax {USA}: {Annual Reviews Inc.}, Sep. 1988, pp. 23--58.

\bibitem{rudinStopExplainingBlack2019}
C.~Rudin, ``Stop explaining black box machine learning models for high stakes
  decisions and use interpretable models instead,'' \emph{Nature Machine
  Intelligence}, vol.~1, no.~5, pp. 206--215, May 2019.

\bibitem{kaurTrustworthyArtificialIntelligence2022}
D.~Kaur, S.~Uslu, K.~J. Rittichier, and A.~Durresi, ``Trustworthy {{Artificial
  Intelligence}}: {{A Review}},'' \emph{ACM Computing Surveys}, vol.~55, no.~2,
  pp. 39:1--39:38, Jan. 2022.

\bibitem{burkartSurveyExplainabilitySupervised2021}
N.~Burkart and M.~F. Huber, ``A {{Survey}} on the {{Explainability}} of
  {{Supervised Machine Learning}},'' \emph{Journal of Artificial Intelligence
  Research}, vol.~70, pp. 245--317, May 2021.

\bibitem{gyevnarBridgingTransparencyGap2023}
B.~Gyevnar, N.~Ferguson, and B.~Schafer, ``Bridging the {{Transparency Gap}}:
  {{What Can Explainable AI Learn From}} the {{AI Act}}?'' in \emph{Proceedings
  of the 26th {{European Conference}} on {{Artificial Intelligence ECAI}}
  2023}, {Krakow, Poland}, Oct. 2023.

\bibitem{feng2021review}
D.~Feng, A.~Harakeh, S.~L. Waslander, and K.~Dietmayer, ``A review and
  comparative study on probabilistic object detection in autonomous driving,''
  \emph{IEEE Transactions on Intelligent Transportation Systems}, vol.~23,
  no.~8, pp. 9961--9980, 2021.

\bibitem{EUTrustAI}
\BIBentryALTinterwordspacing
{European Comission}. (2019) Ethics guidelines for trustworthy ai. [Online].
  Available:
  \url{https://digital-strategy.ec.europa.eu/en/library/ethics-guidelines-trustworthy-ai}
\BIBentrySTDinterwordspacing

\bibitem{ai2023artificial}
\BIBentryALTinterwordspacing
{NIST AI}. (2023) Artificial intelligence risk management framework (ai rmf
  1.0). [Online]. Available:
  \url{https://airc.nist.gov/AI_RMF_Knowledge_Base/Playbook}
\BIBentrySTDinterwordspacing

\bibitem{alzubaidi2023towards}
L.~Alzubaidi, A.~Al-Sabaawi, J.~Bai, A.~Dukhan, A.~H. Alkenani, A.~Al-Asadi,
  H.~A. Alwzwazy, M.~Manoufali, M.~A. Fadhel, A.~Albahri \emph{et~al.},
  ``Towards risk-free trustworthy artificial intelligence: Significance and
  requirements,'' \emph{International Journal of Intelligent Systems}, vol.
  2023, 2023.

\bibitem{andravsko2021sustainable}
J.~Andra{\v{s}}ko, O.~Hamul'{\'a}k, M.~Mesar{\v{c}}{\'\i}k, T.~Kerikm{\"a}e,
  and A.~Kajander, ``Sustainable data governance for cooperative, connected and
  automated mobility in the european union,'' \emph{Sustainability}, vol.~13,
  no.~19, p. 10610, 2021.

\bibitem{GDPR2016a}
\BIBentryALTinterwordspacing
E.~Parliament, ``Regulation ({EU}) 2016/679 of the {European Parliament} and of
  the {Council} of 27 {April} 2016 on the {Protection} of {Natural Persons With
  Regard} to the {Processing} of {Personal Data} and on the {Free Movement} of
  {Such Data} and {Repealing Directive} 95/46/{EC} ({General Data Protection
  Regulation}),'' 2016. [Online]. Available:
  \url{https://data.europa.eu/eli/reg/2016/679/oj}
\BIBentrySTDinterwordspacing

\bibitem{li2023dark}
X.~Li, Z.~Chen, J.~M. Zhang, F.~Sarro, Y.~Zhang, and X.~Liu, ``Dark-skin
  individuals are at more risk on the street: Unmasking fairness issues of
  autonomous driving systems,'' \emph{arXiv preprint arXiv:2308.02935}, 2023.

\bibitem{eykholt2018robust}
K.~Eykholt, I.~Evtimov, E.~Fernandes, B.~Li, A.~Rahmati, C.~Xiao, A.~Prakash,
  T.~Kohno, and D.~Song, ``Robust physical-world attacks on deep learning
  visual classification,'' in \emph{Proceedings of the IEEE conference on
  computer vision and pattern recognition}, 2018, pp. 1625--1634.

\bibitem{yuan2019Adversarial}
X.~Yuan, P.~He, Q.~Zhu, and X.~Li, ``Adversarial examples: Attacks and defenses
  for deep learning,'' \emph{IEEE Transactions on Neural Networks and Learning
  Systems}, vol.~30, no.~9, pp. 2805--2824, 2019.

\bibitem{sae_j3016_taxonomy_2021}
\BIBentryALTinterwordspacing
{SAE J3016}, ``Taxonomy and {Definitions} for {Terms} {Related} to {Driving}
  {Automation} {Systems} for {On}-{Road} {Motor} {Vehicles},'' Apr. 2021.
  [Online]. Available: \url{https://doi.org/10.4271/J3016_202104}
\BIBentrySTDinterwordspacing

\bibitem{NascimentoSafety2020}
A.~M. Nascimento, L.~F. Vismari, C.~B. S.~T. Molina, P.~S. Cugnasca, J.~B.
  Camargo, J.~R.~d. Almeida, R.~Inam, E.~Fersman, M.~V. Marquezini, and A.~Y.
  Hata, ``A systematic literature review about the impact of artificial
  intelligence on autonomous vehicle safety,'' \emph{IEEE Transactions on
  Intelligent Transportation Systems}, vol.~21, no.~12, pp. 4928--4946, 2020.

\bibitem{ribeiro2022requirements}
Q.~A. Ribeiro, M.~Ribeiro, and J.~Castro, ``Requirements engineering for
  autonomous vehicles: a systematic literature review,'' in \emph{Proceedings
  of the 37th ACM/SIGAPP Symposium on Applied Computing}, 2022, pp. 1299--1308.

\bibitem{Sheh2021XAIrequirements}
R.~Sheh, ``Explainable artificial intelligence requirements for safe,
  intelligent robots,'' in \emph{2021 IEEE International Conference on
  Intelligence and Safety for Robotics (ISR)}, 2021, pp. 382--387.

\bibitem{angelov2021explainable}
P.~P. Angelov, E.~A. Soares, R.~Jiang, N.~I. Arnold, and P.~M. Atkinson,
  ``Explainable artificial intelligence: an analytical review,'' \emph{Wiley
  Interdisciplinary Reviews: Data Mining and Knowledge Discovery}, vol.~11,
  no.~5, p. e1424, 2021.

\bibitem{miller2019explanation}
T.~Miller, ``Explanation in artificial intelligence: Insights from the social
  sciences,'' \emph{Artificial intelligence}, vol. 267, pp. 1--38, 2019.

\bibitem{molnarInterpretableMachineLearning2023}
\BIBentryALTinterwordspacing
C.~Molnar, \emph{Interpretable {{Machine Learning}}}.\hskip 1em plus 0.5em
  minus 0.4em\relax Christoph Molnar, 2023. [Online]. Available:
  \url{https://christophm.github.io/interpretable-ml-book/}
\BIBentrySTDinterwordspacing

\bibitem{schwalbe2023comprehensive}
G.~Schwalbe and B.~Finzel, ``A comprehensive taxonomy for explainable
  artificial intelligence: a systematic survey of surveys on methods and
  concepts,'' \emph{Data Mining and Knowledge Discovery}, pp. 1--59, 2023.

\bibitem{speith2022review}
T.~Speith, ``A review of taxonomies of explainable artificial intelligence
  (xai) methods,'' in \emph{Proceedings of the 2022 ACM Conference on Fairness,
  Accountability, and Transparency}, 2022, pp. 2239--2250.

\bibitem{jain2019attention}
S.~Jain and B.~C. Wallace, ``Attention is not explanation,'' \emph{arXiv
  preprint arXiv:1902.10186}, 2019.

\bibitem{kumarProblemsShapleyvaluebasedExplanations2020a}
I.~E. Kumar, S.~Venkatasubramanian, C.~Scheidegger, and S.~Friedler, ``Problems
  with {{Shapley-value-based}} explanations as feature importance measures,''
  in \emph{Proceedings of the 37th {{International Conference}} on {{Machine
  Learning}}}.\hskip 1em plus 0.5em minus 0.4em\relax {PMLR}, Nov. 2020, pp.
  5491--5500.

\bibitem{pendleton2017perception}
S.~D. Pendleton, H.~Andersen, X.~Du, X.~Shen, M.~Meghjani, Y.~H. Eng, D.~Rus,
  and M.~H. Ang~Jr, ``Perception, planning, control, and coordination for
  autonomous vehicles,'' \emph{Machines}, vol.~5, no.~1, p.~6, 2017.

\bibitem{van2018autonomous}
J.~Van~Brummelen, M.~O’Brien, D.~Gruyer, and H.~Najjaran, ``Autonomous
  vehicle perception: The technology of today and tomorrow,''
  \emph{Transportation research part C: emerging technologies}, vol.~89, pp.
  384--406, 2018.

\bibitem{tuncali2019requirements}
C.~E. Tuncali, G.~Fainekos, D.~Prokhorov, H.~Ito, and J.~Kapinski,
  ``Requirements-driven test generation for autonomous vehicles with machine
  learning components,'' \emph{IEEE Transactions on Intelligent Vehicles},
  vol.~5, no.~2, pp. 265--280, 2019.

\bibitem{paden2016survey}
B.~Paden, M.~{\v{C}}{\'a}p, S.~Z. Yong, D.~Yershov, and E.~Frazzoli, ``A survey
  of motion planning and control techniques for self-driving urban vehicles,''
  \emph{IEEE Transactions on intelligent vehicles}, vol.~1, no.~1, pp. 33--55,
  2016.

\bibitem{altafini2002following}
C.~Altafini, ``Following a path of varying curvature as an output regulation
  problem,'' \emph{IEEE Transactions on Automatic Control}, vol.~47, no.~9, pp.
  1551--1556, 2002.

\bibitem{frazzoli2000trajectory}
E.~Frazzoli, M.~A. Dahleh, and E.~Feron, ``Trajectory tracking control design
  for autonomous helicopters using a backstepping algorithm,'' in
  \emph{Proceedings of the 2000 American Control Conference. ACC (IEEE Cat. No.
  00CH36334)}, vol.~6.\hskip 1em plus 0.5em minus 0.4em\relax IEEE, 2000, pp.
  4102--4107.

\bibitem{chen2023endtoend}
L.~Chen, P.~Wu, K.~Chitta, B.~Jaeger, A.~Geiger, and H.~Li, ``End-to-end
  autonomous driving: Challenges and frontiers,'' \emph{arXiv preprint
  arXiv:2306.16927}, 2023.

\bibitem{Tampuu2022endtoend}
A.~Tampuu, T.~Matiisen, M.~Semikin, D.~Fishman, and N.~Muhammad, ``A survey of
  end-to-end driving: Architectures and training methods,'' \emph{IEEE
  Transactions on Neural Networks and Learning Systems}, vol.~33, no.~4, pp.
  1364--1384, 2022.

\bibitem{kitchenhamGuidelinesPerformingSystematic2007}
B.~Kitchenham and S.~Charters, ``Guidelines for performing {{Systematic
  Literature Reviews}} in {{Software Engineering}},'' {School of Computer
  Science and Mathematics, Keele University}, {Keele, UK}, {{EBSE Technical
  Report}} EBSE-2007-01, Jul. 2007.

\bibitem{stepinSurveyContrastiveCounterfactual2021}
I.~Stepin, J.~M. Alonso, A.~Catala, and M.~{Pereira-Fari{\~n}a}, ``A {{Survey}}
  of {{Contrastive}} and {{Counterfactual Explanation Generation Methods}} for
  {{Explainable Artificial Intelligence}},'' \emph{IEEE Access}, vol.~9, pp.
  11\,974--12\,001, 2021.

\bibitem{Du2019}
\BIBentryALTinterwordspacing
M.~Du, N.~Liu, and X.~Hu, ``Techniques for interpretable machine learning,''
  \emph{Communications of the ACM}, vol.~63, no.~1, p. 68–77, Dec. 2019.
  [Online]. Available: \url{http://dx.doi.org/10.1145/3359786}
\BIBentrySTDinterwordspacing

\bibitem{chaghazardiXAITrafficSign}
Z.~Chaghazardi, S.~Fallah, and A.~Tamaddoni-Nezhad, ``Explainable and
  trustworthy traffic sign detection for safe autonomous driving: An inductive
  logic programming approach,'' \emph{Electronic Proceedings in Theoretical
  Computer Science}, vol. 385, pp. 201--212, 08 2023.

\bibitem{FeifelSafetyImpactInterpretableDNN}
P.~Feifel, F.~Bonarens, and F.~Köster, ``Reevaluating the safety impact of
  inherent interpretability on deep neural networks for pedestrian detection,''
  in \emph{2021 IEEE/CVF Conference on Computer Vision and Pattern Recognition
  Workshops (CVPRW)}, 2021, pp. 29--37.

\bibitem{Cordts2016Cityscapes}
M.~Cordts, M.~Omran, S.~Ramos, T.~Rehfeld, M.~Enzweiler, R.~Benenson,
  U.~Franke, S.~Roth, and B.~Schiele, ``The cityscapes dataset for semantic
  urban scene understanding,'' in \emph{Proc. of the IEEE Conference on
  Computer Vision and Pattern Recognition (CVPR)}, 2016.

\bibitem{Losch2021SB}
M.~Losch, M.~Fritz, and B.~Schiele, ``Semantic bottlenecks: Quantifying and
  improving inspectability of deep representations,'' \emph{International
  Journal of Computer Vision}, vol. 129, pp. 3136--3153, 2021.

\bibitem{PlebeAutoencoder}
A.~Plebe and M.~D. Lio, ``On the road with 16 neurons: Towards interpretable
  and manipulable latent representations for visual predictions in driving
  scenarios,'' \emph{IEEE Access}, vol.~8, pp. 179\,716--179\,734, 2020.

\bibitem{ros2016synthia}
G.~Ros, L.~Sellart, J.~Materzynska, D.~Vazquez, and A.~M. Lopez, ``The synthia
  dataset: A large collection of synthetic images for semantic segmentation of
  urban scenes,'' in \emph{Proceedings of the IEEE conference on computer
  vision and pattern recognition}, 2016, pp. 3234--3243.

\bibitem{Oltramari2020NeurosymbolicAF}
A.~Oltramari, J.~Francis, C.~Henson, K.~Ma, and R.~Wickramarachchi,
  ``Neuro-symbolic architectures for context understanding,'' in
  \emph{Knowledge Graphs for eXplainable Artificial Intelligence: Foundations,
  Applications and Challenges}.\hskip 1em plus 0.5em minus 0.4em\relax IOS
  Press, 2020, pp. 143--160.

\bibitem{nuscenes2019}
H.~Caesar, V.~Bankiti, A.~H. Lang, S.~Vora, V.~E. Liong, Q.~Xu, A.~Krishnan,
  Y.~Pan, G.~Baldan, and O.~Beijbom, ``nuscenes: A multimodal dataset for
  autonomous driving,'' \emph{arXiv preprint arXiv:1903.11027}, 2019.

\bibitem{MartinezCapsulate}
J.~Mart\'{i}nez-Cebri\'{a}n, M.-A. Fern\'{a}ndez-Torres, and
  F.~D\'{i}az-De-Mar\'{i}a, ``Interpretable global-local dynamics for the
  prediction of eye fixations in autonomous driving scenarios,'' \emph{IEEE
  Access}, vol.~8, pp. 217\,068--217\,085, 2020.

\bibitem{palazzi2019dreye}
A.~Palazzi, D.~Abati, s.~Calderara, F.~Solera, and R.~Cucchiara, ``Predicting
  the driver's focus of attention: The dr(eye)ve project,'' \emph{IEEE
  Transactions on Pattern Analysis and Machine Intelligence}, vol.~41, no.~7,
  pp. 1720--1733, 2019.

\bibitem{YonakaSunGlare}
K.~Yoneda, N.~Ichihara, H.~Kawanishi, T.~Okuno, L.~Cao, and N.~Suganuma,
  ``Sun-glare region recognition using visual explanations for traffic light
  detection,'' in \emph{2021 IEEE Intelligent Vehicles Symposium (IV)}, 2021,
  pp. 1464--1469.

\bibitem{albrechtInterpretableGoalbasedPrediction2021}
S.~V. Albrecht, C.~Brewitt, J.~Wilhelm, B.~Gyevnar, F.~Eiras, M.~Dobre, and
  S.~Ramamoorthy, ``Interpretable {{Goal-based Prediction}} and {{Planning}}
  for {{Autonomous Driving}},'' in \emph{2021 {{IEEE International Conference}}
  on {{Robotics}} and {{Automation}} ({{ICRA}})}, May 2021, pp. 1043--1049.

\bibitem{hannaInterpretableGoalRecognition2021}
J.~P. Hanna, A.~Rahman, E.~Fosong, F.~Eiras, M.~Dobre, J.~Redford,
  S.~Ramamoorthy, and S.~V. Albrecht, ``Interpretable {{Goal Recognition}} in
  the {{Presence}} of {{Occluded Factors}} for {{Autonomous Vehicles}},'' in
  \emph{2021 {{IEEE}}/{{RSJ International Conference}} on {{Intelligent
  Robots}} and {{Systems}} ({{IROS}})}, Sep. 2021, pp. 7044--7051.

\bibitem{antonello2022flash}
M.~Antonello, M.~Dobre, S.~V. Albrecht, J.~Redford, and S.~Ramamoorthy,
  ``Flash: Fast and light motion prediction for autonomous driving with
  {Bayesian} inverse planning and learned motion profiles,'' in \emph{2022
  IEEE/RSJ International Conference on Intelligent Robots and Systems
  (IROS)}.\hskip 1em plus 0.5em minus 0.4em\relax IEEE, 2022, pp. 9829--9836.

\bibitem{NGSIM}
{U.S. Department Of Transportation Federal Highway Administration}, ``Next
  {{Generation Simulation}} ({{NGSIM}}) {{Vehicle Trajectories}} and
  {{Supporting Data}},'' 2017.

\bibitem{brewitt2021grit}
C.~Brewitt, B.~Gyevnar, S.~Garcin, and S.~V. Albrecht, ``{GRIT:} fast,
  interpretable, and verifiable goal recognition with learned decision trees
  for autonomous driving,'' in \emph{IEEE/RSJ International Conference on
  Intelligent Robots and Systems (IROS)}, 2021.

\bibitem{brewitt2023ogrit}
C.~Brewitt, M.~Tamborski, C.~Wang, and S.~V. Albrecht, ``Verifiable goal
  recognition for autonomous driving with occlusions,'' IEEE/RSJ International
  Conference on Intelligent Robots and Systems, 2023.

\bibitem{inDdataset}
J.~Bock, R.~Krajewski, T.~Moers, S.~Runde, L.~Vater, and L.~Eckstein, ``The ind
  dataset: A drone dataset of naturalistic road user trajectories at german
  intersections,'' in \emph{2020 IEEE Intelligent Vehicles Symposium (IV)},
  2020, pp. 1929--1934.

\bibitem{rounDdataset}
R.~Krajewski, T.~Moers, J.~Bock, L.~Vater, and L.~Eckstein, ``The round
  dataset: A drone dataset of road user trajectories at roundabouts in
  germany,'' in \emph{2020 IEEE 23rd International Conference on Intelligent
  Transportation Systems (ITSC)}, 2020, pp. 1--6.

\bibitem{breuer2020opendd}
A.~Breuer, J.-A. Term{\"o}hlen, S.~Homoceanu, and T.~Fingscheidt, ``opendd: A
  large-scale roundabout drone dataset,'' in \emph{2020 IEEE 23rd International
  Conference on Intelligent Transportation Systems (ITSC)}.\hskip 1em plus
  0.5em minus 0.4em\relax IEEE, 2020, pp. 1--6.

\bibitem{ghoulInterpretableGoalBasedModel2023}
A.~Ghoul, I.~Yahiaoui, A.~{Verroust-Blondet}, and F.~Nashashibi,
  ``Interpretable {{Goal-Based}} model for {{Vehicle Trajectory Prediction}} in
  {{Interactive Scenarios}},'' in \emph{2023 {{IEEE Intelligent Vehicles
  Symposium}} ({{IV}})}, Jun. 2023, pp. 1--6.

\bibitem{interactiondataset}
W.~Zhan, L.~Sun, D.~Wang, H.~Shi, A.~Clausse, M.~Naumann, J.~K\"ummerle,
  H.~K\"onigshof, C.~Stiller, A.~de~La~Fortelle, and M.~Tomizuka,
  ``{INTERACTION} {Dataset}: {An} {INTERnational}, {Adversarial} and
  {Cooperative} {moTION} {Dataset} in {Interactive} {Driving} {Scenarios} with
  {Semantic} {Maps},'' \emph{arXiv:1910.03088 [cs, eess]}, Sep. 2019.

\bibitem{gyevnar2022humancentric}
B.~Gyevnar, M.~Tamborski, C.~Wang, C.~G. Lucas, S.~B. Cohen, and S.~V.
  Albrecht, ``A human-centric method for generating causal explanations in
  natural language for autonomous vehicle motion planning,'' in \emph{IJCAI
  Workshop on Artificial Intelligence for Autonomous Driving}, 2022.

\bibitem{omeiza2022spoken}
D.~Omeiza, S.~Anjomshoae, H.~Webb, M.~Jirotka, and L.~Kunze, ``From spoken
  thoughts to automated driving commentary: Predicting and explaining
  intelligent vehicles’ actions,'' in \emph{2022 IEEE Intelligent Vehicles
  Symposium (IV)}.\hskip 1em plus 0.5em minus 0.4em\relax IEEE, 2022, pp.
  1040--1047.

\bibitem{henzeHowCanAutomated2022}
F.~Henze, D.~Fa{\ss}bender, and C.~Stiller, ``How {{Can Automated Vehicles
  Explain Their Driving Decisions}}? {{Generating Clarifying Summaries
  Automatically}},'' in \emph{2022 {{IEEE Intelligent Vehicles Symposium}}
  ({{IV}})}, Jun. 2022, pp. 935--942.

\bibitem{kleinInterpretableClassifiersBased2023}
K.~Klein, O.~De~Candido, and W.~Utschick, ``Interpretable {{Classifiers Based}}
  on {{Time-Series Motifs}} for {{Lane Change Prediction}},'' \emph{IEEE
  Transactions on Intelligent Vehicles}, vol.~8, no.~7, pp. 3954--3961, Jul.
  2023.

\bibitem{highDdataset}
R.~Krajewski, J.~Bock, L.~Kloeker, and L.~Eckstein, ``The highd dataset: A
  drone dataset of naturalistic vehicle trajectories on german highways for
  validation of highly automated driving systems,'' in \emph{2018 21st
  International Conference on Intelligent Transportation Systems (ITSC)}, 2018,
  pp. 2118--2125.

\bibitem{kridalukmanaSelfExplainingAbilitiesIntelligent2022}
R.~Kridalukmana, H.~Lu, and M.~Naderpour, ``Self-{{Explaining Abilities}} of an
  {{Intelligent Agent}} for {{Transparency}} in a {{Collaborative Driving
  Context}},'' \emph{IEEE Transactions on Human-Machine Systems}, vol.~52,
  no.~6, pp. 1155--1165, Dec. 2022.

\bibitem{Dosovitskiy17}
A.~Dosovitskiy, G.~Ros, F.~Codevilla, A.~Lopez, and V.~Koltun, ``{CARLA}: {An}
  open urban driving simulator,'' in \emph{Proceedings of the 1st Annual
  Conference on Robot Learning}, 2017, pp. 1--16.

\bibitem{muschollEMIDASExplainableSocial2021a}
N.~Muscholl, M.~Klusch, P.~Gebhard, and T.~Schneeberger, ``{{EMIDAS}}:
  {{Explainable}} social interaction-based pedestrian intention detection
  across street,'' in \emph{Proceedings of the {{ACM Symposium}} on {{Applied
  Computing}}}, 2021, pp. 107--115.

\bibitem{neumeierVariationalAutoencoderBasedVehicle2021a}
M.~Neumeier, M.~Botsch, A.~Tollk{\"u}hn, and T.~Berberich, ``Variational
  {{Autoencoder-Based Vehicle Trajectory Prediction}} with an {{Interpretable
  Latent Space}},'' in \emph{2021 {{IEEE International Intelligent
  Transportation Systems Conference}} ({{ITSC}})}, Sep. 2021, pp. 820--827.

\bibitem{wuHybridDrivingDecisionMaking2023}
M.~Wu, F.~R. Yu, P.~X. Liu, and Y.~He, ``A {{Hybrid Driving Decision-Making
  System Integrating Markov Logic Networks}} and {{Connectionist AI}},''
  \emph{IEEE Transactions on Intelligent Transportation Systems}, vol.~24,
  no.~3, pp. 3514--3527, Mar. 2023.

\bibitem{zheng2023towards}
H.~Zheng, Z.~Zang, S.~Yang, and R.~Mangharam, ``Towards explainability in
  modular autonomous system software,'' in \emph{2023 IEEE Intelligent Vehicles
  Symposium (IV)}.\hskip 1em plus 0.5em minus 0.4em\relax IEEE, 2023, pp. 1--8.

\bibitem{PCABishop1999}
M.~E. Tipping and C.~M. Bishop, ``Probabilistic principal component analysis,''
  \emph{Journal of the Royal Statistical Society Series B: Statistical
  Methodology}, vol.~61, no.~3, pp. 611--622, 1999.

\bibitem{tSNELaurenz2008}
L.~Van~der Maaten and G.~Hinton, ``Visualizing data using t-sne.''
  \emph{Journal of machine learning research}, vol.~9, no.~11, 2008.

\bibitem{capsulesHinton2011}
G.~E. Hinton, A.~Krizhevsky, and S.~D. Wang, ``Transforming auto-encoders,'' in
  \emph{Artificial Neural Networks and Machine Learning -- ICANN 2011},
  T.~Honkela, W.~Duch, M.~Girolami, and S.~Kaski, Eds.\hskip 1em plus 0.5em
  minus 0.4em\relax Berlin, Heidelberg: Springer Berlin Heidelberg, 2011, pp.
  44--51.

\bibitem{selvarajuGradCAMVisualExplanations2017}
R.~R. Selvaraju, M.~Cogswell, A.~Das, R.~Vedantam, D.~Parikh, and D.~Batra,
  ``Grad-{{CAM}}: {{Visual Explanations}} from {{Deep Networks}} via
  {{Gradient-Based Localization}},'' in \emph{2017 {{IEEE International
  Conference}} on {{Computer Vision}} ({{ICCV}})}, Oct. 2017, pp. 618--626.

\bibitem{ponn2020identification}
T.~Ponn, T.~Kr{\"o}ger, and F.~Diermeyer, ``Identification and explanation of
  challenging conditions for camera-based object detection of automated
  vehicles,'' \emph{Sensors}, vol.~20, no.~13, p. 3699, 2020.

\bibitem{cui2022interpretation}
Z.~Cui, M.~Li, Y.~Huang, Y.~Wang, and H.~Chen, ``An interpretation framework
  for autonomous vehicles decision-making via shap and rf,'' in \emph{2022 6th
  CAA International Conference on Vehicular Control and Intelligence
  (CVCI)}.\hskip 1em plus 0.5em minus 0.4em\relax IEEE, 2022, pp. 1--7.

\bibitem{liExplainingMachineLearningLane2023}
M.~Li, Y.~Wang, H.~Sun, Z.~Cui, Y.~Huang, and H.~Chen, ``Explaining a
  {{Machine-Learning Lane Change Model With Maximum Entropy Shapley Values}},''
  \emph{IEEE Transactions on Intelligent Vehicles}, vol.~8, no.~6, pp.
  3620--3628, 2023.

\bibitem{maLaneChangeAnalysis2021}
Y.~Ma, S.~Song, L.~Zhang, L.~Xiong, and J.~Chen, ``Lane {{Change Analysis}} and
  {{Prediction Using Mean Impact Value Method}} and {{Logistic Regression
  Model}},'' in \emph{2021 {{IEEE International Intelligent Transportation
  Systems Conference}} ({{ITSC}})}, Sep. 2021, pp. 1346--1352.

\bibitem{mishra2022not}
A.~Mishra, U.~Soni, J.~Huang, and C.~Bryan, ``Why? why not? when? visual
  explanations of agent behaviour in reinforcement learning,'' in \emph{2022
  IEEE 15th Pacific Visualization Symposium (PacificVis)}.\hskip 1em plus 0.5em
  minus 0.4em\relax IEEE, 2022, pp. 111--120.

\bibitem{gyevnar2024cema}
B.~Gyevnar, C.~Wang, C.~G. Lucas, S.~B. Cohen, and S.~V. Albrecht, ``Causal
  explanations for sequential decision-making in multi-agent systems,'' in
  \emph{International Conference on Autonomous Agents and Multi-Agent Systems
  (AAMAS)}, 2024.

\bibitem{gyevnar2024headd}
B.~Gyevnar, S.~Droop, T.~Quillien, S.~B. Cohen, N.~R. Bramley, C.~G. Lucas, and
  S.~V. Albrecht, ``People attribute purpose to autonomous vehicles when
  explaining their behavior,'' 2024.

\bibitem{xAI_control_Dassanayake}
P.~M. Dassanayake, A.~Anjum, A.~K. Bashir, J.~Bacon, R.~Saleem, and W.~Manning,
  ``A deep learning based explainable control system for reconfigurable
  networks of edge devices,'' \emph{IEEE Transactions on Network Science and
  Engineering}, vol.~9, no.~1, pp. 7--19, 2022.

\bibitem{Zemni2023OCTET}
M.~Zemni, M.~Chen, {\'E}.~Zablocki, H.~Ben-Younes, P.~P{\'e}rez, and M.~Cord,
  ``Octet: Object-aware counterfactual explanations,'' in \emph{Proceedings of
  the IEEE/CVF Conference on Computer Vision and Pattern Recognition}, 2023,
  pp. 15\,062--15\,071.

\bibitem{yu2020bdd100k}
F.~Yu, H.~Chen, X.~Wang, W.~Xian, Y.~Chen, F.~Liu, V.~Madhavan, and T.~Darrell,
  ``Bdd100k: A diverse driving dataset for heterogeneous multitask learning,''
  in \emph{Proceedings of the IEEE/CVF conference on computer vision and
  pattern recognition}, 2020, pp. 2636--2645.

\bibitem{xAI_ObjectInduced_xu}
Y.~Xu, X.~Yang, L.~Gong, H.-C. Lin, T.-Y. Wu, Y.~Li, and N.~Vasconcelos,
  ``Explainable object-induced action decision for autonomous vehicles,'' in
  \emph{2020 IEEE/CVF Conference on Computer Vision and Pattern Recognition
  (CVPR)}, 2020, pp. 9520--9529.

\bibitem{shi2020self}
W.~Shi, G.~Huang, S.~Song, Z.~Wang, T.~Lin, and C.~Wu, ``Self-supervised
  discovering of interpretable features for reinforcement learning,''
  \emph{IEEE Transactions on Pattern Analysis and Machine Intelligence},
  vol.~44, no.~5, pp. 2712--2724, 2020.

\bibitem{lundberg2017SHAP}
S.~M. Lundberg and S.-I. Lee, ``A unified approach to interpreting model
  predictions,'' \emph{Advances in neural information processing systems},
  vol.~30, pp. 4765--4774, 2017.

\bibitem{freyer2021shapley}
D.~Fryer, I.~Strümke, and H.~Nguyen, ``Shapley values for feature selection:
  The good, the bad, and the axioms,'' \emph{IEEE Access}, vol.~9, pp.
  144\,352--144\,360, 2021.

\bibitem{bilodeau2024impossibility}
B.~Bilodeau, N.~Jaques, P.~W. Koh, and B.~Kim, ``Impossibility theorems for
  feature attribution,'' \emph{Proceedings of the National Academy of
  Sciences}, vol. 121, no.~2, p. e2304406120, 2024.

\bibitem{kronenberger2020dependency}
J.~Kronenberger and A.~Haselhoff, ``Dependency decomposition and a reject
  option for explainable models,'' \emph{arXiv preprint arXiv:2012.06523},
  2020.

\bibitem{Houben-IJCNN-2013}
S.~Houben, J.~Stallkamp, J.~Salmen, M.~Schlipsing, and C.~Igel, ``Detection of
  traffic signs in real-world images: The {G}erman {T}raffic {S}ign {D}etection
  {B}enchmark,'' in \emph{International Joint Conference on Neural Networks},
  no. 1288, 2013.

\bibitem{HackerSaliencyPlausibility}
L.~Hacker and J.~Seewig, ``Insufficiency-driven dnn error detection in the
  context of sotif on traffic sign recognition use case,'' \emph{IEEE Open
  Journal of Intelligent Transportation Systems}, vol.~4, pp. 58--70, 2023.

\bibitem{keser2023interpretable}
M.~Keser, G.~Schwalbe, A.~Nowzad, and A.~Knoll, ``Interpretable model-agnostic
  plausibility verification for 2d object detectors using domain-invariant
  concept bottleneck models,'' in \emph{Proceedings of the IEEE/CVF Conference
  on Computer Vision and Pattern Recognition}, 2023, pp. 3890--3899.

\bibitem{lin2014microsoft}
T.-Y. Lin, M.~Maire, S.~Belongie, J.~Hays, P.~Perona, D.~Ramanan,
  P.~Doll{\'a}r, and C.~L. Zitnick, ``Microsoft coco: Common objects in
  context,'' in \emph{Computer Vision--ECCV 2014: 13th European Conference,
  Zurich, Switzerland, September 6-12, 2014, Proceedings, Part V 13}.\hskip 1em
  plus 0.5em minus 0.4em\relax Springer, 2014, pp. 740--755.

\bibitem{bau2017network}
D.~Bau, B.~Zhou, A.~Khosla, A.~Oliva, and A.~Torralba, ``Network dissection:
  Quantifying interpretability of deep visual representations,'' in
  \emph{Proceedings of the IEEE conference on computer vision and pattern
  recognition}, 2017, pp. 6541--6549.

\bibitem{Fritsch2013ITSC}
J.~Fritsch, T.~Kuehnl, and A.~Geiger, ``A new performance measure and
  evaluation benchmark for road detection algorithms,'' in \emph{International
  Conference on Intelligent Transportation Systems (ITSC)}, 2013.

\bibitem{fang2023toward}
Y.~Fang, H.~Min, X.~Wu, X.~Lei, S.~Chen, R.~Teixeira, and X.~Zhao, ``Toward
  interpretability in fault diagnosis for autonomous vehicles: Interpretation
  of sensor data anomalies,'' \emph{IEEE Sensors Journal}, vol.~23, no.~5, pp.
  5014--5027, 2023.

\bibitem{baoDRIVEDeepReinforced2021}
W.~Bao, Q.~Yu, and Y.~Kong, ``{{DRIVE}}: {{Deep Reinforced Accident
  Anticipation}} with {{Visual Explanation}},'' in \emph{2021 {{IEEE}}/{{CVF
  International Conference}} on {{Computer Vision}} ({{ICCV}})}, Oct. 2021, pp.
  7599--7608.

\bibitem{fang2021dada}
J.~Fang, D.~Yan, J.~Qiao, J.~Xue, and H.~Yu, ``Dada: Driver attention
  prediction in driving accident scenarios,'' \emph{IEEE Transactions on
  Intelligent Transportation Systems}, vol.~23, no.~6, pp. 4959--4971, 2021.

\bibitem{chen2023attention}
G.~Chen, Y.~Zhang, and X.~Li, ``Attention-based highway safety planner for
  autonomous driving via deep reinforcement learning,'' \emph{IEEE Transactions
  on Vehicular Technology}, 2023.

\bibitem{di2020interpretable}
C.~Di~Franco and N.~Bezzo, ``Interpretable run-time monitoring and replanning
  for safe autonomous systems operations,'' \emph{IEEE Robotics and Automation
  Letters}, vol.~5, no.~2, pp. 2427--2434, 2020.

\bibitem{gall2021gaussian}
C.~Gall and N.~Bezzo, ``Gaussian process-based interpretable runtime adaptation
  for safe autonomous systems operations in unstructured environments,'' in
  \emph{2021 IEEE/RSJ International Conference on Intelligent Robots and
  Systems (IROS)}.\hskip 1em plus 0.5em minus 0.4em\relax IEEE, 2021, pp.
  123--129.

\bibitem{gilpin2021explaining}
L.~H. Gilpin, V.~Penubarthi, and L.~Kagal, ``Explaining multimodal errors in
  autonomous vehicles,'' in \emph{2021 IEEE 8th International Conference on
  Data Science and Advanced Analytics (DSAA)}.\hskip 1em plus 0.5em minus
  0.4em\relax IEEE, 2021, pp. 1--10.

\bibitem{gorospeAnalyzingInterVehicleCollision2023}
J.~Gorospe, S.~Hasan, M.~R. Islam, A.~A. G{\'o}mez, S.~Girs, and E.~Uhlemann,
  ``Analyzing {{Inter-Vehicle Collision Predictions}} during {{Emergency
  Braking}} with {{Automated Vehicles}},'' in \emph{2023 19th {{International
  Conference}} on {{Wireless}} and {{Mobile Computing}}, {{Networking}} and
  {{Communications}} ({{WiMob}})}, Jun. 2023, pp. 411--418.

\bibitem{karimExplainableArtificialIntelligence2022}
M.~Karim, Y.~Li, and R.~Qin, ``Toward {{Explainable Artificial Intelligence}}
  for {{Early Anticipation}} of {{Traffic Accidents}},'' \emph{Transportation
  Research Record}, vol. 2676, no.~6, pp. 743--755, 2022.

\bibitem{BaoMM2020}
W.~Bao, Q.~Yu, and Y.~Kong, ``Uncertainty-based traffic accident anticipation
  with spatio-temporal relational learning,'' in \emph{Proceedings of the 28th
  ACM International Conference on Multimedia}, 2020, pp. 2682--2690.

\bibitem{nahata2021assessing}
R.~Nahata, D.~Omeiza, R.~Howard, and L.~Kunze, ``Assessing and explaining
  collision risk in dynamic environments for autonomous driving safety,'' in
  \emph{2021 IEEE International Intelligent Transportation Systems Conference
  (ITSC)}.\hskip 1em plus 0.5em minus 0.4em\relax IEEE, 2021, pp. 223--230.

\bibitem{houstonOneThousandOne2021}
J.~Houston, G.~Zuidhof, L.~Bergamini, Y.~Ye, L.~Chen, A.~Jain, S.~Omari,
  V.~Iglovikov, and P.~Ondruska, ``One {{Thousand}} and {{One Hours}}:
  {{Self-driving Motion Prediction Dataset}},'' in \emph{Proceedings of the
  2020 {{Conference}} on {{Robot Learning}}}.\hskip 1em plus 0.5em minus
  0.4em\relax PMLR, Oct. 2021, pp. 409--418.

\bibitem{schmidt2021can}
L.~M. Schmidt, G.~Kontes, A.~Plinge, and C.~Mutschler, ``Can you trust your
  autonomous car? interpretable and verifiably safe reinforcement learning,''
  in \emph{2021 IEEE Intelligent Vehicles Symposium (IV)}.\hskip 1em plus 0.5em
  minus 0.4em\relax IEEE, 2021, pp. 171--178.

\bibitem{ZeilerOcclusion2014}
M.~D. Zeiler and R.~Fergus, ``Visualizing and understanding convolutional
  networks,'' in \emph{Computer Vision -- ECCV 2014}, D.~Fleet, T.~Pajdla,
  B.~Schiele, and T.~Tuytelaars, Eds.\hskip 1em plus 0.5em minus 0.4em\relax
  Cham: Springer International Publishing, 2014, pp. 818--833.

\bibitem{Yatbaz2024_Introspection}
H.~Y. Yatbaz, M.~Dianati, and R.~Woodman, ``Introspection of dnn-based
  perception functions in automated driving systems: State-of-the-art and open
  research challenges,'' \emph{IEEE Transactions on Intelligent Transportation
  Systems}, vol.~25, no.~2, pp. 1112--1130, 2024.

\bibitem{kolekar2022explainable}
S.~Kolekar, S.~Gite, B.~Pradhan, and A.~Alamri, ``Explainable ai in scene
  understanding for autonomous vehicles in unstructured traffic environments on
  indian roads using the inception u-net model with grad-cam visualization,''
  \emph{Sensors}, vol.~22, no.~24, p. 9677, 2022.

\bibitem{IDD2019_Varma}
G.~Varma, A.~Subramanian, A.~Namboodiri, M.~Chandraker, and C.~Jawahar, ``Idd:
  A dataset for exploring problems of autonomous navigation in unconstrained
  environments,'' in \emph{2019 IEEE Winter Conference on Applications of
  Computer Vision (WACV)}, 2019, pp. 1743--1751.

\bibitem{SaravanarjanGradCam}
V.~S. Saravanarajan, R.-C. Chen, C.-H. Hsieh, and L.-S. Chen, ``Improving
  semantic segmentation under hazy weather for autonomous vehicles using
  explainable artificial intelligence and adaptive dehazing approach,''
  \emph{IEEE Access}, vol.~11, pp. 38\,194--38\,207, 2023.

\bibitem{Brostow2009}
\BIBentryALTinterwordspacing
G.~J. Brostow, J.~Fauqueur, and R.~Cipolla, ``Semantic object classes in video:
  A high-definition ground truth database,'' \emph{Pattern Recognition
  Letters}, vol.~30, no.~2, p. 88–97, Jan. 2009. [Online]. Available:
  \url{http://dx.doi.org/10.1016/j.patrec.2008.04.005}
\BIBentrySTDinterwordspacing

\bibitem{AbukmeilVariationalAttention}
M.~Abukmeil, A.~Genovese, V.~Piuri, F.~Rundo, and F.~Scotti, ``Towards
  explainable semantic segmentation for autonomous driving systems by
  multi-scale variational attention,'' in \emph{2021 IEEE International
  Conference on Autonomous Systems (ICAS)}, 2021, pp. 1--5.

\bibitem{geyer2020a2d2}
J.~Geyer, Y.~Kassahun, M.~Mahmudi, X.~Ricou, R.~Durgesh, A.~S. Chung,
  L.~Hauswald, V.~H. Pham, M.~M{\"u}hlegg, S.~Dorn \emph{et~al.}, ``A2d2: Audi
  autonomous driving dataset,'' \emph{arXiv preprint arXiv:2004.06320}, 2020.

\bibitem{mankodiya2022od}
H.~Mankodiya, D.~Jadav, R.~Gupta, S.~Tanwar, W.-C. Hong, and R.~Sharma,
  ``Od-xai: Explainable ai-based semantic object detection for autonomous
  vehicles,'' \emph{Applied Sciences}, vol.~12, no.~11, p. 5310, 2022.

\bibitem{NowakChargerDetection}
T.~Nowak, M.~R. Nowicki, K.~Ćwian, and P.~Skrzypczyński, ``How to improve
  object detection in a driver assistance system applying explainable deep
  learning,'' in \emph{2019 IEEE Intelligent Vehicles Symposium (IV)}, 2019,
  pp. 226--231.

\bibitem{Schinagl_2022_CVPR}
D.~Schinagl, G.~Krispel, H.~Possegger, P.~M. Roth, and H.~Bischof, ``Occam's
  laser: Occlusion-based attribution maps for 3d object detectors on lidar
  data,'' in \emph{Proceedings of the IEEE/CVF Conference on Computer Vision
  and Pattern Recognition (CVPR)}, June 2022, pp. 1141--1150.

\bibitem{Geiger2012CVPR}
A.~Geiger, P.~Lenz, and R.~Urtasun, ``Are we ready for autonomous driving? the
  kitti vision benchmark suite,'' in \emph{Conference on Computer Vision and
  Pattern Recognition (CVPR)}, 2012.

\bibitem{BoschVATLD}
L.~Gou, L.~Zou, N.~Li, M.~Hofmann, A.~K. Shekar, A.~Wendt, and L.~Ren, ``Vatld:
  A visual analytics system to assess, understand and improve traffic light
  detection,'' \emph{IEEE Transactions on Visualization and Computer Graphics},
  vol.~27, no.~2, pp. 261--271, 2021.

\bibitem{BehrendtNovak2017ICRA}
K.~Behrendt, L.~Novak, and R.~Botros, ``A deep learning approach to traffic
  lights: Detection, tracking, and classification,'' in \emph{2017 IEEE
  International Conference on Robotics and Automation (ICRA)}.\hskip 1em plus
  0.5em minus 0.4em\relax IEEE, 2017, pp. 1370--1377.

\bibitem{ShorrNeuroscope2021}
C.~Schorr, P.~Goodarzi, F.~Chen, and T.~Dahmen, ``Neuroscope: An explainable ai
  toolbox for semantic segmentation and image classification of convolutional
  neural nets,'' \emph{Applied Sciences}, vol.~11, no.~5, p. 2199, 2021.

\bibitem{WangSpatioTemporalVisualAnalytics}
J.~Wang, Y.~Li, Z.~Zhou, C.~Wang, Y.~Hou, L.~Zhang, X.~Xue, M.~Kamp, X.~L.
  Zhang, and S.~Chen, ``When, where and how does it fail? a spatial-temporal
  visual analytics approach for interpretable object detection in autonomous
  driving,'' \emph{IEEE Transactions on Visualization and Computer Graphics},
  vol.~29, no.~12, pp. 5033--5049, 2023.

\bibitem{HaedeckeScrutinAI2022}
E.~Haedecke, M.~Mock, and M.~Akila, ``{ScrutinAI: A Visual Analytics Approach
  for the Semantic Analysis of Deep Neural Network Predictions},'' in
  \emph{EuroVis Workshop on Visual Analytics (EuroVA)}, J.~Bernard and
  M.~Angelini, Eds.\hskip 1em plus 0.5em minus 0.4em\relax The Eurographics
  Association, 2022.

\bibitem{jiangIntentionAwareInteractiveTransformer2023}
T.~Jiang, Y.~Liu, Q.~Dong, and T.~Xu, ``Intention-{{Aware Interactive
  Transformer}} for {{Real-Time Vehicle Trajectory Prediction}} in {{Dense
  Traffic}},'' \emph{Transportation Research Record}, vol. 2677, no.~3, pp.
  946--960, Mar. 2023.

\bibitem{kochakarnExplainableActionPrediction2023}
P.~Kochakarn, D.~De~Martini, D.~Omeiza, and L.~Kunze, ``Explainable {{Action
  Prediction}} through {{Self-Supervision}} on {{Scene Graphs}},'' in
  \emph{2023 {{IEEE International Conference}} on {{Robotics}} and
  {{Automation}} ({{ICRA}})}, May 2023, pp. 1479--1485.

\bibitem{singh2022road}
G.~Singh, S.~Akrigg, M.~Di~Maio, V.~Fontana, R.~J. Alitappeh, S.~Saha,
  K.~Jeddisaravi, F.~Yousefi, J.~Culley, T.~Nicholson \emph{et~al.}, ``Road:
  The road event awareness dataset for autonomous driving,'' \emph{IEEE
  Transactions on Pattern Analysis \& Machine Intelligence}, no.~01, pp. 1--1,
  feb 5555.

\bibitem{RobotCarDatasetIJRR}
\BIBentryALTinterwordspacing
W.~Maddern, G.~Pascoe, C.~Linegar, and P.~Newman, ``{1 Year, 1000km: The Oxford
  RobotCar Dataset},'' \emph{The International Journal of Robotics Research
  (IJRR)}, vol.~36, no.~1, pp. 3--15, 2017. [Online]. Available:
  \url{http://dx.doi.org/10.1177/0278364916679498}
\BIBentrySTDinterwordspacing

\bibitem{liu2023interpretable}
H.~Liu, J.~Zhao, and L.~Zhang, ``Interpretable and flexible target-conditioned
  neural planners for autonomous vehicles,'' in \emph{2023 IEEE International
  Conference on Robotics and Automation (ICRA)}.\hskip 1em plus 0.5em minus
  0.4em\relax IEEE, 2023, pp. 10\,076--10\,082.

\bibitem{wangLaneChangeIntentionPrediction2022a}
K.~Wang, J.~Hou, and X.~Zeng, ``Lane-{{Change Intention Prediction}} of
  {{Surrounding Vehicles Using BiLSTM-CRF Models}} with {{Rule Embedding}},''
  in \emph{Proceedings - 2022 {{Chinese Automation Congress}}, {{CAC}} 2022},
  vol. 2022-January, 2022, pp. 2764--2769.

\bibitem{huTrajectoryPredictionNeural2023}
H.~Hu, Q.~Wang, M.~Cheng, and Z.~Gao, ``Trajectory {{Prediction Neural
  Network}} and {{Model Interpretation Based}} on {{Temporal Pattern
  Attention}},'' \emph{IEEE Transactions on Intelligent Transportation
  Systems}, vol.~24, no.~3, pp. 2746--2759, Mar. 2023.

\bibitem{yuSceneGraphAugmentedDataDriven2022}
S.-Y. Yu, A.~V. Malawade, D.~Muthirayan, P.~P. Khargonekar, and M.~A.~A.
  Faruque, ``Scene-{{Graph Augmented Data-Driven Risk Assessment}} of
  {{Autonomous Vehicle Decisions}},'' \emph{IEEE Transactions on Intelligent
  Transportation Systems}, vol.~23, no.~7, pp. 7941--7951, Jul. 2022.

\bibitem{ramanishka2018CVPR}
V.~Ramanishka, Y.-T. Chen, T.~Misu, and K.~Saenko, ``Toward driving scene
  understanding: A dataset for learning driver behavior and causal reasoning,''
  in \emph{Conference on Computer Vision and Pattern Recognition (CVPR)}, 2018.

\bibitem{xtrajpred_Zhang2022}
K.~Zhang and L.~Li, ``Explainable multimodal trajectory prediction using
  attention models,'' \emph{Transportation Research Part C: Emerging
  Technologies}, vol. 143, p. 103829, Oct. 2022.

\bibitem{kim2018textual}
J.~Kim, A.~Rohrbach, T.~Darrell, J.~Canny, and Z.~Akata, ``Textual explanations
  for self-driving vehicles,'' in \emph{Proceedings of the European conference
  on computer vision (ECCV)}, 2018, pp. 563--578.

\bibitem{kuhn2023textual}
M.~A. K{\"u}hn, D.~Omeiza, and L.~Kunze, ``Textual explanations for automated
  commentary driving,'' \emph{arXiv preprint arXiv:2304.08178}, 2023.

\bibitem{Dong2022Transformer}
J.~Dong, S.~Chen, M.~Miralinaghi, T.~Chen, and S.~Labi, ``Development and
  testing of an image transformer for explainable autonomous driving systems,''
  \emph{Journal of Intelligent and Connected Vehicles}, vol.~5, no.~3, pp.
  235--249, 2022.

\bibitem{Zhang2023Interrelation}
Z.~Zhang, R.~Tian, R.~Sherony, J.~Domeyer, and Z.~Ding, ``Attention-based
  interrelation modeling for explainable automated driving,'' \emph{IEEE
  Transactions on Intelligent Vehicles}, vol.~8, no.~2, pp. 1564--1573, 2023.

\bibitem{fengNLEDMNaturalLanguageExplanations2023}
Y.~Feng, W.~Hua, and Y.~Sun, ``{{NLE-DM}}: {{Natural-Language Explanations}}
  for {{Decision Making}} of {{Autonomous Driving Based}} on {{Semantic Scene
  Understanding}},'' \emph{IEEE Transactions on Intelligent Transportation
  Systems}, vol.~24, no.~9, pp. 9780--9791, Sep. 2023.

\bibitem{zhang2023tactical}
Y.~Zhang, W.~Wang, X.~Zhou, Q.~Wang, and X.~Sun, ``Tactical-level explanation
  is not enough: Effect of explaining av’s lane-changing decisions on
  drivers’ decision-making, trust, and emotional experience,''
  \emph{International Journal of Human--Computer Interaction}, vol.~39, no.~7,
  pp. 1438--1454, 2023.

\bibitem{Chen2021PSIAP}
\BIBentryALTinterwordspacing
T.~Chen, R.~Tian, Y.~Chen, J.~E. Domeyer, H.~Toyoda, R.~Sherony, T.~Jing, and
  Z.~Ding, ``Psi: A pedestrian behavior dataset for socially intelligent
  autonomous car,'' \emph{ArXiv}, vol. abs/2112.02604, 2021. [Online].
  Available: \url{https://api.semanticscholar.org/CorpusID:244909387}
\BIBentrySTDinterwordspacing

\bibitem{gaddSenseAssessEXplain2020g}
M.~Gadd, D.~{de Martini}, L.~Marchegiani, P.~Newman, and L.~Kunze,
  ``Sense\textendash{{Assess}}\textendash{{eXplain}} ({{SAX}}): {{Building
  Trust}} in {{Autonomous Vehicles}} in {{Challenging Real-World Driving
  Scenarios}},'' in \emph{2020 {{IEEE Intelligent Vehicles Symposium}}
  ({{IV}})}, Oct. 2020, pp. 150--155.

\bibitem{mori2019visual}
K.~Mori, H.~Fukui, T.~Murase, T.~Hirakawa, T.~Yamashita, and H.~Fujiyoshi,
  ``Visual explanation by attention branch network for end-to-end
  learning-based self-driving,'' in \emph{2019 IEEE intelligent vehicles
  symposium (IV)}.\hskip 1em plus 0.5em minus 0.4em\relax IEEE, 2019, pp.
  1577--1582.

\bibitem{Wang2021End2EndInterpretable}
H.~Wang, P.~Cai, Y.~Sun, L.~Wang, and M.~Liu, ``Learning interpretable
  end-to-end vision-based motion planning for autonomous driving with optical
  flow distillation,'' in \emph{2021 IEEE International Conference on Robotics
  and Automation (ICRA)}, 2021, pp. 13\,731--13\,737.

\bibitem{Chen2022End2EndLatent}
J.~Chen, S.~E. Li, and M.~Tomizuka, ``Interpretable end-to-end urban autonomous
  driving with latent deep reinforcement learning,'' \emph{IEEE Transactions on
  Intelligent Transportation Systems}, vol.~23, no.~6, pp. 5068--5078, 2022.

\bibitem{Yang2019SceneUnderstanding}
S.~Yang, W.~Wang, C.~Liu, and W.~Deng, ``Scene understanding in deep
  learning-based end-to-end controllers for autonomous vehicles,'' \emph{IEEE
  Transactions on Systems, Man, and Cybernetics: Systems}, vol.~49, no.~1, pp.
  53--63, 2019.

\bibitem{cultrera2020explaining}
L.~Cultrera, L.~Seidenari, F.~Becattini, P.~Pala, and A.~Del~Bimbo,
  ``Explaining autonomous driving by learning end-to-end visual attention,'' in
  \emph{Proceedings of the IEEE/CVF Conference on Computer Vision and Pattern
  Recognition Workshops}, 2020, pp. 340--341.

\bibitem{Espi2005TORCSTO}
B.~Wymann, E.~Espi{\'e}, C.~Guionneau, C.~Dimitrakakis, R.~Coulom, and
  A.~Sumner, ``Torcs, the open racing car simulator,'' \emph{Software available
  at http://torcs. sourceforge. net}, vol.~4, no.~6, p.~2, 2000.

\bibitem{AksoyAttentionModel}
E.~Aksoy and A.~Yazici, ``Attention model for extracting saliency map in
  driving videos,'' in \emph{2020 28th Signal Processing and Communications
  Applications Conference (SIU)}, 2020, pp. 1--4.

\bibitem{BDDA_Xia2017}
\BIBentryALTinterwordspacing
Y.~Xia, D.~Zhang, J.~Kim, K.~Nakayama, K.~Zipser, and D.~Whitney, ``Predicting
  driver attention in critical situations,'' 2017. [Online]. Available:
  \url{https://arxiv.org/abs/1711.06406}
\BIBentrySTDinterwordspacing

\bibitem{CAT2000_Borji2015}
\BIBentryALTinterwordspacing
A.~Borji and L.~Itti, ``Cat2000: A large scale fixation dataset for boosting
  saliency research,'' 2015. [Online]. Available:
  \url{https://arxiv.org/abs/1505.03581}
\BIBentrySTDinterwordspacing

\bibitem{Chitta2021NEAT}
K.~Chitta, A.~Prakash, and A.~Geiger, ``Neat: Neural attention fields for
  end-to-end autonomous driving,'' in \emph{2021 IEEE/CVF International
  Conference on Computer Vision (ICCV)}, 2021, pp. 15\,773--15\,783.

\bibitem{Sadat2020}
A.~Sadat, S.~Casas, M.~Ren, X.~Wu, P.~Dhawan, and R.~Urtasun, \emph{Perceive,
  Predict, and Plan: Safe Motion Planning Through Interpretable Semantic
  Representations}.\hskip 1em plus 0.5em minus 0.4em\relax Springer
  International Publishing, 2020, p. 414–430.

\bibitem{wei2021perceive}
B.~Wei, M.~Ren, W.~Zeng, M.~Liang, B.~Yang, and R.~Urtasun, ``Perceive, attend,
  and drive: Learning spatial attention for safe self-driving,'' in \emph{2021
  IEEE International Conference on Robotics and Automation (ICRA)}.\hskip 1em
  plus 0.5em minus 0.4em\relax IEEE, 2021, pp. 4875--4881.

\bibitem{Tashiro2023QuantizedActivation}
Y.~Tashiro and H.~Awano, ``Pay attention via quantization: Enhancing
  explainability of neural networks via quantized activation,'' \emph{IEEE
  Access}, vol.~11, pp. 34\,431--34\,439, 2023.

\bibitem{udacity2016}
{Udacity}, ``Udacity self-driving car driving data,'' Jun. 2020, [online]
  Available: \url{https://github.com/udacity/self-driving-car}.

\bibitem{teng2022hierarchical}
S.~Teng, L.~Chen, Y.~Ai, Y.~Zhou, Z.~Xuanyuan, and X.~Hu, ``Hierarchical
  interpretable imitation learning for end-to-end autonomous driving,''
  \emph{IEEE Transactions on Intelligent Vehicles}, vol.~8, no.~1, pp.
  673--683, 2022.

\bibitem{kadir2001saliency}
T.~Kadir and M.~Brady, ``Saliency, scale and image description,''
  \emph{International Journal of Computer Vision}, vol.~45, no.~2, pp. 83--105,
  2001.

\bibitem{Springenberg2015deconv}
J.~T. Springenberg, A.~Dosovitskiy, T.~Brox, and M.~A. Riedmiller, ``Striving
  for simplicity: The all convolutional net,'' in \emph{3rd International
  Conference on Learning Representations, {ICLR} 2015, San Diego, CA, USA, May
  7-9, 2015, Workshop Track Proceedings}, 2015.

\bibitem{wiegreffeAttentionNotNot2019b}
S.~Wiegreffe and Y.~Pinter, ``Attention is not not {{Explanation}},'' in
  \emph{Proceedings of the 2019 {{Conference}} on {{Empirical Methods}} in
  {{Natural Language Processing}} and the 9th {{International Joint
  Conference}} on {{Natural Language Processing}} ({{EMNLP-IJCNLP}})}, K.~Inui,
  J.~Jiang, V.~Ng, and X.~Wan, Eds.\hskip 1em plus 0.5em minus 0.4em\relax Hong
  Kong, China: Association for Computational Linguistics, Nov. 2019, pp.
  11--20.

\bibitem{corso2020interpretable}
A.~Corso and M.~J. Kochenderfer, ``Interpretable safety validation for
  autonomous vehicles,'' in \emph{2020 IEEE 23rd International Conference on
  Intelligent Transportation Systems (ITSC)}.\hskip 1em plus 0.5em minus
  0.4em\relax IEEE, 2020, pp. 1--6.

\bibitem{decastro2020interpretable}
J.~DeCastro, K.~Leung, N.~Ar{\'e}chiga, and M.~Pavone, ``Interpretable policies
  from formally-specified temporal properties,'' in \emph{2020 IEEE 23rd
  International Conference on Intelligent Transportation Systems (ITSC)}.\hskip
  1em plus 0.5em minus 0.4em\relax IEEE, 2020, pp. 1--7.

\bibitem{kang2022vision}
M.~Kang, W.~Lee, K.~Hwang, and Y.~Yoon, ``Vision transformer for detecting
  critical situations and extracting functional scenario for automated vehicle
  safety assessment,'' \emph{Sustainability}, vol.~14, no.~15, p. 9680, 2022.

\bibitem{MenzelPegasusScenarios2019}
T.~Menzel, G.~Bagschik, L.~Isensee, A.~Schomburg, and M.~Maurer, ``From
  functional to logical scenarios: Detailing a keyword-based scenario
  description for execution in a simulation environment,'' in \emph{2019 IEEE
  Intelligent Vehicles Symposium (IV)}, 2019, pp. 2383--2390.

\bibitem{LIDLHybridFramwework2020}
Y.~Li, H.~Wang, L.~M. Dang, T.~N. Nguyen, D.~Han, A.~Lee, I.~Jang, and H.~Moon,
  ``A deep learning-based hybrid framework for object detection and recognition
  in autonomous driving,'' \emph{IEEE Access}, vol.~8, pp. 194\,228--194\,239,
  2020.

\bibitem{petsiuk2018rise}
V.~Petsiuk, A.~Das, and K.~Saenko, ``Rise: Randomized input sampling for
  explanation of black-box models,'' \emph{arXiv preprint arXiv:1806.07421},
  2018.

\bibitem{shao2023safety}
H.~Shao, L.~Wang, R.~Chen, H.~Li, and Y.~Liu, ``Safety-enhanced autonomous
  driving using interpretable sensor fusion transformer,'' in \emph{Conference
  on Robot Learning}.\hskip 1em plus 0.5em minus 0.4em\relax PMLR, 2023, pp.
  726--737.

\bibitem{brajovic2023model}
D.~Brajovic, N.~Renner, V.~P. Goebels, P.~Wagner, B.~Fresz, M.~Biller,
  M.~Klaeb, J.~Kutz, J.~Neuhuettler, and M.~F. Huber, ``Model reporting for
  certifiable ai: A proposal from merging eu regulation into ai development,''
  \emph{arXiv preprint arXiv:2307.11525}, 2023.

\bibitem{waschleReviewAISafety2022}
M.~W{\"a}schle, F.~Thaler, A.~Berres, F.~P{\"o}lzlbauer, and A.~Albers, ``A
  review on {{AI Safety}} in highly automated driving,'' \emph{Frontiers in
  Artificial Intelligence}, vol.~5, 2022.

\bibitem{wangDataBanzhafRobust2023}
J.~Wang and R.~Jia, ``Data {{Banzhaf}}: {{A Robust Data Valuation Framework}}
  for {{Machine Learning}},'' in \emph{Proceedings of {{Machine Learning
  Research}}}, vol. 206, 2023, pp. 6388--6421.

\bibitem{wang2023application}
C.~Wang, F.~Guo, R.~Yu, L.~Wang, and Y.~Zhang, ``The application of driver
  models in the safety assessment of autonomous vehicles: Perspectives,
  insights, prospects,'' \emph{IEEE Transactions on Intelligent Vehicles},
  2023.

\bibitem{vilone2020explainable}
G.~Vilone and L.~Longo, ``Explainable artificial intelligence: a systematic
  review,'' \emph{arXiv preprint arXiv:2006.00093}, 2020.

\bibitem{marcu2024lingoqavideoquestionanswering}
\BIBentryALTinterwordspacing
A.-M. Marcu, L.~Chen, J.~Hünermann, A.~Karnsund, B.~Hanotte, P.~Chidananda,
  S.~Nair, V.~Badrinarayanan, A.~Kendall, J.~Shotton, E.~Arani, and
  O.~Sinavski, ``Lingoqa: Video question answering for autonomous driving,''
  2024. [Online]. Available: \url{https://arxiv.org/abs/2312.14115}
\BIBentrySTDinterwordspacing

\bibitem{elhage2021mathematical}
N.~Elhage, N.~Nanda, C.~Olsson, T.~Henighan, N.~Joseph, B.~Mann, A.~Askell,
  Y.~Bai, A.~Chen, T.~Conerly \emph{et~al.}, ``A mathematical framework for
  transformer circuits,'' \emph{Transformer Circuits Thread}, vol.~1, 2021.

\end{thebibliography}

\begin{IEEEbiography}[{\includegraphics[width=1in,height=1.25in,clip,keepaspectratio]{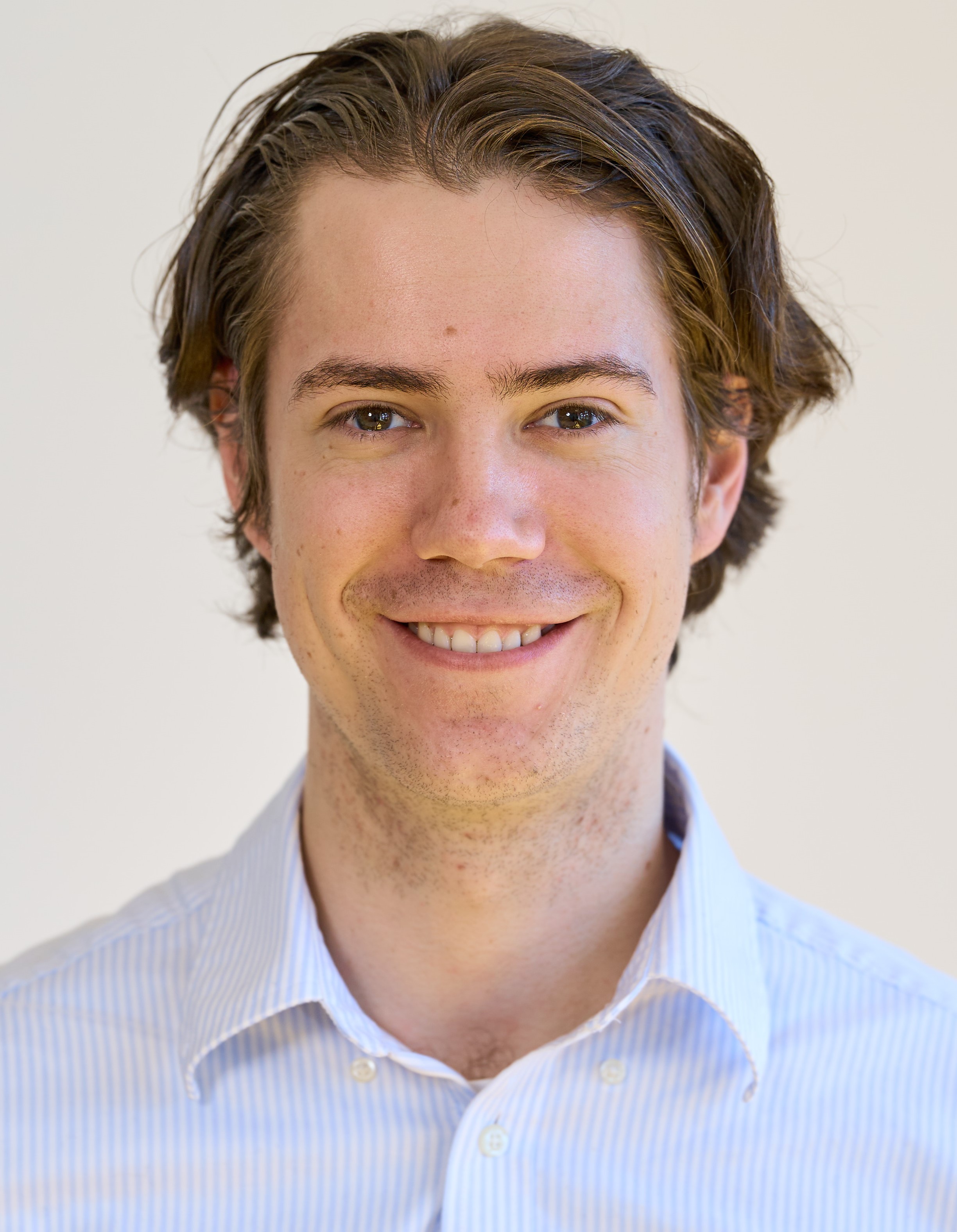}}]{Anton Kuznietsov}
received the B.Sc. and M.Sc. degrees in computational engineering from the Technical University of Darmstadt, Darmstadt, Germany, in 2020 and 2022, respectively. 
Since 2023, he has been working as a research associate with the Institute of Automotive Engineering at the Technical University of Darmstadt. His research interests include explainable AI and AI safety in autonomous driving. 
\end{IEEEbiography}
\begin{IEEEbiography}[{\includegraphics[width=1in,height=1.25in,clip,keepaspectratio]{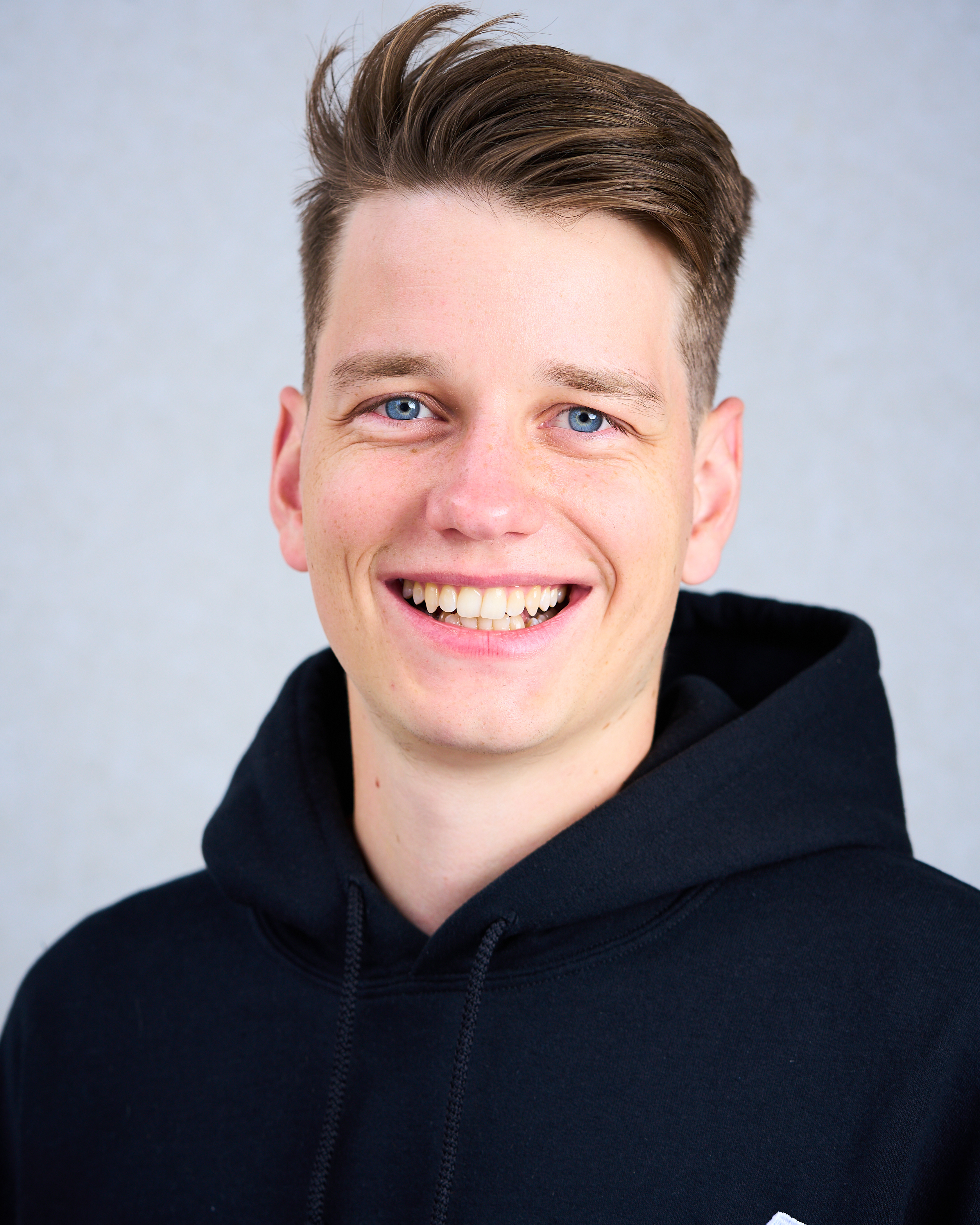}}]{Balint Gyevnar}
graduated with First Class Honours in Master of Informatics from the University of Edinburgh in 2021. 
He is currently a 4$^{th}$ year PhD student at the University of Edinburgh in the Autonomous Agents Research Group under the supervision of Stefano Albrecht, Shay Cohen, and Christopher Lucas.
He is interested in researching human-centred explainable methods for multi-agent systems, focusing on causality, conversational agents, and AI safety for human-aligned trust calibration.
\end{IEEEbiography}
\begin{IEEEbiography}[{\includegraphics[width=1in,height=1.25in,clip,keepaspectratio]{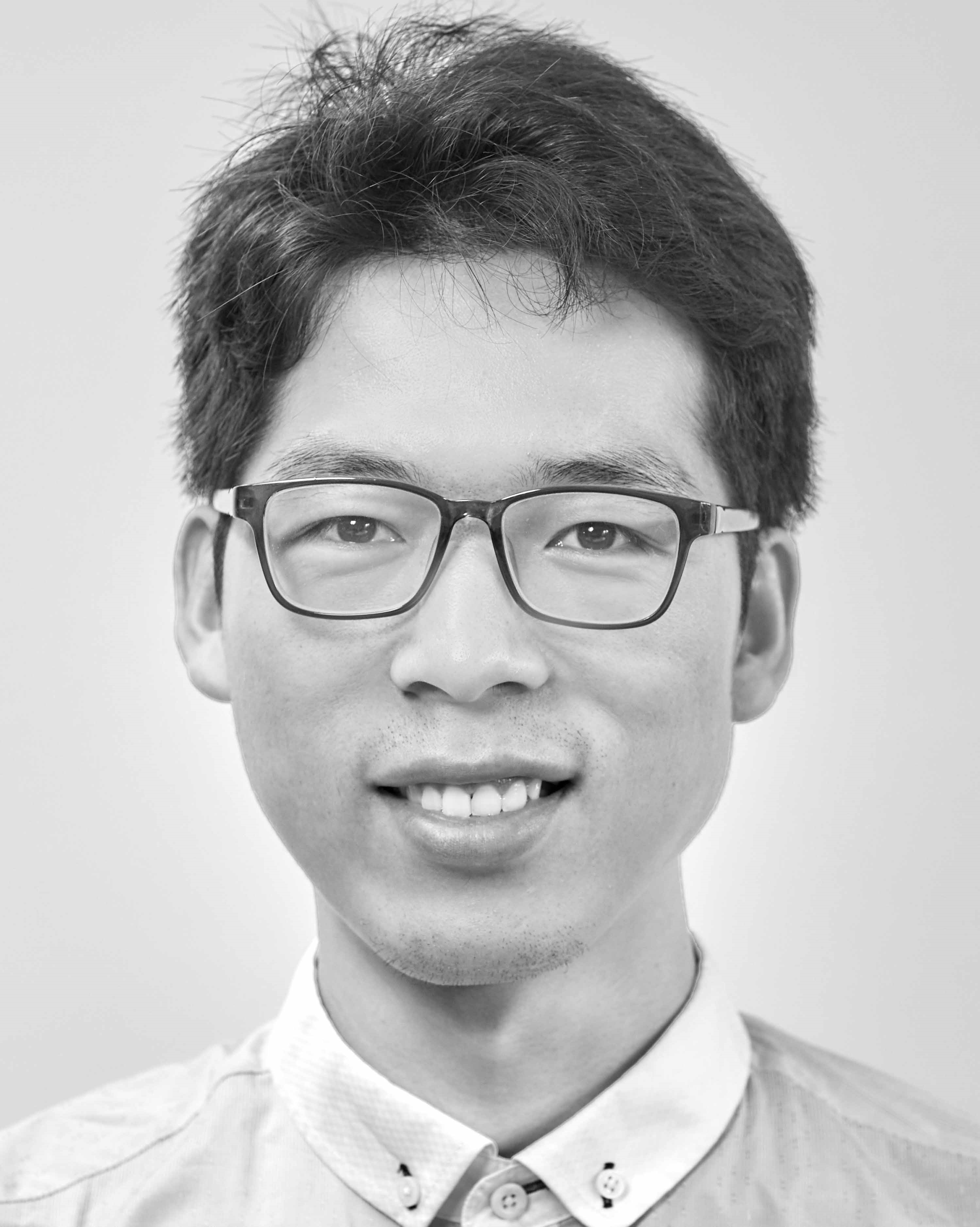}}]{Cheng Wang}
received the B.Sc. degree from the School of Automotive Engineering at Wuhan University of Technology, Wuhan, China, in 2014 and the M.Sc. degree from the School of Automotive studies at Tongji University, Shanghai, China, in 2017 and the Ph.D. degree from the Institute of Automotive Engineering at the Technical University of Darmstadt, Darmstadt, Germany, in 2021. From 2022 to 2024, he worked as a research associate at the University of Edinburgh. He is now an assistant professor at the Heriot-Watt University, Edinburgh. His research interest includes explainable AI and safety verification and validation of autonomous vehicles.
\end{IEEEbiography}
\begin{IEEEbiography}[{\includegraphics[width=1in,height=1.25in,clip,keepaspectratio]{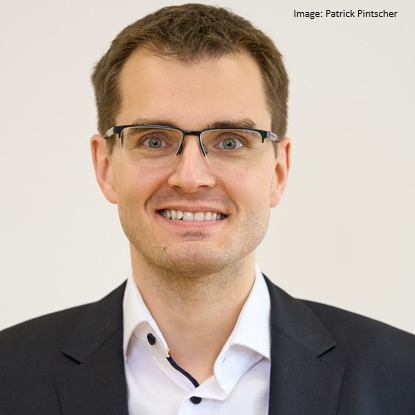}}]{Steven Peters}
was born in 1987, received his PhD (Dr.-Ing.) in 2013, at Karlsruhe Institute of
Technology, Karlsruhe, Baden-Württemberg, Germany. From 2016 to 2022 he worked as Manager of Artificial Intelligence Research at Mercedes-Benz AG
in Germany. He is a Full Professor at the Technical
University of Darmstadt, Darmstadt, Germany and
heads the Institute of Automotive Engineering in the
Department of Mechanical Engineering since 2022.
\end{IEEEbiography}
\begin{IEEEbiography}[{\includegraphics[width=1in,height=1.25in,clip,keepaspectratio]{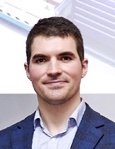}}]{Stefano V. Albrecht} is an Associate Professor of Artificial Intelligence in the School of Informatics, University of Edinburgh. He leads the Autonomous Agents Research Group (https://agents.inf.ed.ac.uk/) which develops machine learning algorithms for autonomous systems, including applications in autonomous driving. Previously, he was a postdoctoral researcher at the University of Texas at Austin. He obtained his PhD and MSc degrees in AI from the University of Edinburgh, and BSc degree in computer science from the Technical University of Darmstadt.
\end{IEEEbiography}

\vfill

\end{document}